\RequirePackage{fix-cm} 
\documentclass[smallcondensed]{svjour3}     
\smartqed  
\usepackage{microtype}
\usepackage{graphicx}
\usepackage{subfigure}
\usepackage{booktabs} 
\usepackage{hyperref}

\usepackage{color}
\usepackage{amsmath,amssymb,amstext}
\usepackage{multirow}
\usepackage{xcolor}
\usepackage{bbm}
\usepackage{bm}
\usepackage{tikz}
\usetikzlibrary{bayesnet, shapes, arrows, positioning, calc, patterns, shadows, external}

\usepackage{wrapfig}

\definecolor{red}{rgb}{.9,0.1,0.1}
\definecolor{blue}{rgb}{.2,0.5,0.7}

\newcommand\<{\langle}
\renewcommand\>{\rangle}
\newcommand{\isep}{\mathrel{{.}\,{.}}\nobreak}

\newcommand{\balpha}{{\bm \alpha}}

\newcommand{\B}{{\bm B}}

\newcommand{\bI}{\mathbf{I}}

\newcommand{\cL}{\mathcal{L}}
\newcommand{\cl}{\ell}

\newcommand{\bmu}{{\bm \mu}}

\DeclareMathOperator*{\N}{\mathbbm{N}}

\newcommand{\cQ}{\mathcal{Q}}

\DeclareMathOperator*{\R}{\mathbbm{R}}
\newcommand{\Rp}{{\R}_{\geq0}}

\renewcommand{\S}{\mathbb{S}}

\newcommand{\W}{{\bm W}}

\newcommand{\X}{{\bm X}}
\newcommand{\cX}{\mathcal{X}}
\newcommand{\x}{{\bm x}}
\newcommand{\tx}{\bar{\x}}
\newcommand{\tX}{\bar{\X}}

\newcommand{\bz}{{\bm z}}
\newcommand{\tz}{\tilde{\bz}}

\newcommand{\Dec}{\mathcal{G}}
\newcommand{\Enc}{\mathcal{F}}
\newcommand{\Softmax}{\operatorname{softmax}}

\begin{document}
    
    \title{Joint Optimization of an Autoencoder for Clustering and Embedding}

    \subtitle{}
    
    
    \author{Ahcene~Boubekki\dag \and 
    Michael~Kampffmeyer\dag  \and 
    Ulf~Brefeld\ddag \and 
    Robert~Jenssen\dag
    }
    
    
    \institute{
        {} \at \dag Machine Learning Group, Department of Physics and Technology \\ 
        UiT The Arctic University of Norway \\ 
        Hansine Hansens veg 18, 9019 Troms\o{}, Norway \\
        \email{ ahcene.boubekki ; michael.c.kampffmeyer ; robert.jenssen @uit.no}           
        \and
        {} \at \ddag  Institute of Information Systems\\
        21335 Lüneburg, 
        Universitätsallee 1, Germany\\
        \email{brefeld@leuphana.de}
    }

    \date{Received: date / Accepted: date}

    \maketitle
    
    
    \begin{abstract}

        
        Deep embedded clustering has become a dominating approach to unsupervised categorization of objects with deep neural networks.
The optimization of the most popular methods alternates between the training of a deep autoencoder and a $k$-means clustering of the autoencoder's embedding. The diachronic setting, however, prevents the former to benefit from valuable information acquired by the latter.
In this paper, we present an alternative where the autoencoder and the clustering are learned simultaneously.
This is achieved by providing novel theoretical insight, where we show that the objective function of a certain class of Gaussian mixture models (GMM's) can naturally be rephrased as the loss function of a one-hidden layer autoencoder thus inheriting the built-in clustering capabilities of the GMM. 
That simple neural network, referred to as the clustering module, can be integrated into a deep autoencoder resulting in a deep clustering model able to jointly learn a clustering and an embedding.
Experiments confirm the equivalence between the clustering module and Gaussian mixture models.
Further evaluations affirm the empirical relevance of our deep architecture as it outperforms related baselines on several data sets.

        \keywords{clustering \and deep autoencoders \and {embedding} \and $k$-means \and Gaussian mixture models}
    \end{abstract}

    \section{Introduction}
    
    Clustering is one of the oldest and most difficult problems in machine learning and a great deal of different clustering techniques have been proposed in the literature. Perhaps the most prominent clustering approach is the $k$-means algorithm~\cite{lloyd1982least}. Its simplicity renders the algorithm particularly attractive to practitioners beyond the field of data mining and machine learning~\cite{punj1983cluster,gasch2002exploring,hennig2013find,frandsen2015automatic}. A variety of related algorithms and extensions have been studied~\cite{dunn1973fuzzy,krishna1999genetic,dhillon2004kernel}.
    The model behind the {standard k-means} algorithm is an isotropic Gaussian mixture model~\cite{kulis2012revisiting,lucke2019k} (GMM).
    As such, it can only produce linear partitions of the input space. 
    
    A {dominating} recent line of research aims to alleviate this issue by learning an embedding of the data with a deep autoencoder (DAE) and to have {variants of} Gaussian mixture models operate on the embedded instances \cite{xie2016unsupervised,yang2016towards,guo2017improved,fard2020deep,miklautz2020deep}. {This can be referred to as \emph{deep embedded clustering}.} {An important reason for the emergence of this modern line of research in clustering is that it combines elements (the deep autoencoder and the GMM clustering) that are theoretically well understood when considered separately}. {In practice, since the embedding map is nonlinear, the result is that the otherwise linear clustering translates to a nonlinear solution in input space.} This approach resembles the rationale behind kernel methods~\cite{kernel-trick}. The difference lies in either computing a  kernel matrix  before learning the clustering versus learning the feature map as the function approximated by the encoder. Another advantage of an autoencoder (AE) based approach is {provided by} the decoder of the network that allows to map the centroids back into (an approximated version of the) input space. Depending on the kernel function/matrix, this is not always possible with traditional approaches.
    
    {However, even with advantages of deep embedded clustering as described above, the fact that the embedding procedure and the clustering procedure are decoupled means that the former cannot benefit from the latter, and vice versa. In the most naive approach, the deep autoencoder is first trained, and then the features output by the autoencoder are clustered. In \cite{xie2016unsupervised}, an improvement is made by alternatively shrinking the clusters in the embedded space around centroids for then to update the latter using $k$-means. The shrinking may however lead to an adaptation of the clustering to the embedding instead of the desired clustering-induced embedding. In general, these approaches cannot handle a random initialization which usually leads to noisy embeddings and, in turn, meaningless clusterings. 
    }  
    
    {Despite the obvious need,  {few} deep embedded clustering procedures have been proposed featuring an inherent capability of bringing out clustering structure in a joint and simultaneous procedure. This is clearly due to the lack of a theoretical foundation that could support such an objective.}    
    {In this paper, we provide novel theoretical insight that enables us to simultaneously optimize both the embedding and the clustering by relying on a neural network interpretation of Gaussian mixture models. Our solution derives from the fact that during the EM, the objective function of a certain class of GMM can be rephrased as the loss function of an autoencoder, as we show as a key result. A network trained with that loss is thus guaranteed to learn a linear clustering. 
    Our contributions are threefold. (1) We state and prove a theorem connecting Gaussian mixture models and autoencoders. (2) We define the clustering module (CM) as a one hidden layer AE  with $\Softmax$ activation trained with the loss resulting from the previous point. (3) We integrate the CM into a deep autoencoder to form a novel class of deep clustering architectures, denoted AE-CM.
    Building on the theoretical insight, the experiments confirm the empirical equivalence of the clustering module with GMM.
    As for the AE-CM, the model outperforms on several datasets baselines extending Gaussian mixture models to nonlinear clustering with deep autoencoders.}
    
    The {remainder}  of the paper is organized as follows. 
    Section~\ref{sec:related-work}  reviews related works and baselines and Section~\ref{sec:loss} provides the theory underpinning our contributions. The clustering module and the AE-CM are introduced in  
    Section~\ref{sec:cm}. We report empirical evaluations in Section~\ref{sec:exp}. 
    Section \ref{sec:conclusion} concludes the paper.

    \section{Related Work}\label{sec:related-work}

    
    
    
    
    Recently, several deep learning models have been proposed to group data in an unsupervised manner~\cite{tian2014learning,springenberg2015unsupervised,yang2016joint,kampffmeyer2017deep,ghasedi2017deep,haeusser2018associative,mcconville2019n2d,mukherjee2019clustergan,uugur2020variational,van2020scan}.
    Since our main contribution unravels the connection between Gaussian mixture models and autoencoders, models based on associative~\cite{haeusser2018associative,yang2019deep}, spectral~\cite{tian2014learning,yang2019deep,bianchi2020spectral} or subspace~\cite{zhang2019neural,miklautz2020deep} clusterings are outside the scope of this paper. 
    Furthermore, unlike \cite{yang2016joint,chang2017deep,kampffmeyer2017deep,ghasedi2017deep,guo2017deep,caron2018deep,ji2019invariant,van2020scan}, our model falls within the category of general purpose deep embedded clustering models which builds upon GMM~\cite{xie2016unsupervised,guo2017improved,yang2016towards,fard2020deep}. 
    {We mention that recent works related to specific topics such as perturbation robustness and non-redundant subspace clustering also utilize ideas that are to some degree related to the general concept of deep embedded clustering~\cite{yang2020adversarial,miklautz2020deep}}.
    Given the high generality of our method, we detail models that are also built without advanced mechanisms, but can still support them.
    
    One of the first approaches in {this latter} category {of dominating modern deep clustering algorithms} is DEC~\cite{xie2016unsupervised}, which consists of an embedding network coupled with an ad-hoc matrix representing the centroids. During training, the network is trained to shrink the data around their closest centroids in a way similar to t-SNE~\cite{maaten2008visualizing}. However, instead of matching the distributions in the input and feature spaces, the target distribution is a mixture of Student-t models inferred from a $k$-means clustering of the embedding itself. The latter is updated every few iterations to keep track of evolution of the embedding and the centroids are stored in the ad-hoc matrix.
    In order to avoid sub-optimal solutions when starting from random initialization, the embedding network is pre-trained. 
    Nevertheless, the difference between the two training phases can cause stability issues, such as collapsing centroids.
    IDEC~\cite{guo2017improved} extends DEC and alleviates this issue {to some degree} by using the reconstruction loss of the DAE also during the training phase. 
    Although IDEC is more robust than DEC, both are highly dependent on the initial embedding as the clustering phase quickly shrinks the data around the centroids. 
    The Deep Clustering Network (DCN)~\cite{yang2016towards} {further strengthens the deep embedding clustering framework and} is based on an architecture comparable to DEC and IDEC but includes hard clustering. 
    The optimization scheme consists of three steps: 1) Optimize the DAE to reconstruct and bring the data-points closer to their closest centroids (similar to $k$-means); 2) update assignments to the closest clusters; 3) update the centroids iteratively using the assignments. Although alternating optimization helps prevent instabilities during the optimization, DCN still requires initialization schemes in order to reach good solutions.
    Our proposed model instead, while being theoretically-driven, resolves the need for alternative optimization schemes the previous methods rely on. 
    Since we learn a variant of GMM using a neural network, the optimizations of both the embedding and the clustering profit from each other. In practice, our model is thus less reliant on pre-training schemes.
    
    To some degree our approach is similar the very recent method known as DKM~\cite{fard2020deep}. 
    The model shares the same architecture as DEC, namely, a deep autoencoder and an ad-hoc matrix representing the centroids. However, here, the model optimizes the network and the centroids simultaneously.
    To achieve that the loss function includes a term similar to the loss of fuzzy c-means~\cite{bezdek2013pattern}, i.e, $k$-means with soft-assignments.
    The cluster coefficients are computed from the distances to the centroids and normalized using a parameterized $\Softmax$. In order to convert soft assignments into hard ones, the parameter follows an annealing strategy.
    In our model, the assignment probability function is indirectly related to the distance to the closest cluster representative, as used in DKM.
    Our the loss function aims to minimize the distance between the data points and their reconstruction as convex combinations of the centroids. This means that the learning of the centroid and of the assignment probabilities regularize each other, mitigating the need of an annealing strategy. 

    {There are also other recent and promising approaches to clustering that are leveraging deep neural networks. However, these operate in a very different manner compared to deep embedded clustering, which our novel theoretical insight sheds new light on and which enables our new proposed autoencoder optimization for joint clustering and embedding.} 
    {We nevertheless briefly review some of these approaches and also compare to representatives of such methods in our experiments.} 
    In particular, a great deal of works have focus on the generative aspect of Gaussian mixture models and studied variations based on deep generative models such as variational autoencoders (VAE)~\cite{kingma2013auto} and generative adversarial network (GAN)~\cite{goodfellow2014generative}.
    
    Adversarial approaches for clustering have been proposed with examples being CatGAN~\cite{springenberg2015unsupervised} and adversarial autoencoders (AAE)~\cite{makhzani2015adversarial}.  
    The former optimizes the mutual information and predicts a categorical distribution of classes in the data while maximizing at the same time the robustness against an adversarial generative model.
    Instead, AAE uses two adversarial networks to impose a Gaussian and a categorical distribution in the latent space.
    The recent ClusterGAN~\cite{mukherjee2019clustergan} makes an original use of an autoencoder as the input is not a data point but a sample drawn from the product of a multinomial and a Gaussian distributions. The decoded sample is then fed to the discriminator and the encoder.
    This method is all robust to random network initialization.

    Variational approaches that enable clustering include methods like Gaussian Mixture VAE (GMVAE)~\cite{dilokthanakul2016deep} and Variational Deep Embedding (VaDE)~\cite{jiang2016variational}. Although the former explicitly refers to GMMs, both methods are similar. 
    The difference is that the parameters of the Gaussian distributions in embedded space depend only on the cluster index in VaDE whereas GMVAE involves a second random variable.
    Both models require a pre-trained autoencoder.
    Variational information bottleneck with Gaussian mixture model (VIB-GMM)~\cite{uugur2020variational} also assumes a mixture of Gaussian distributions on the embedding, however the model is trained accordingly to the deep variational information bottleneck (VIB) framework.
    The latter can be seen as a information theory-driven generalization of GAN and VAE~\cite{tishby2000information,alemi2016deep}.
    VIB-GMM reveals to be a robust alternative to previous approaches as it is able to produce meaningful clusterings from a randomly initialized autoencoder.


    .
    
    \section{Theoretical Groundwork}\label{sec:loss}
    
    Throughout the paper the set of positive integers is denoted by $\N^*$.
    The set of the $d$-dimensional stochastic vectors is written as $\S^d = \{ \x \in \Rp^d \: : \:  \sum_{i=1}^d x_i = 1\}.$
    When the context allows, the ranges of the indices are abbreviated using the upper-bound, e.g., a vector $\x \in \R^d$ decomposes as $\x=\< x_j \>_{1\leq j \leq d}= \< x_j \>_d$.
    The zero vector of $\R^d$ is written as ${\bm 0}_d$. 
    The notation extends to any number.
    The identity matrix of $\R^{d \times d}$ is denoted by $\bI_d$.
    
    \subsection{From GMMs to Autoencoders}
    
    We aim to fit an isotropic Gaussian mixture model with $K \in \N^*$ components and a Dirichlet prior on the average cluster responsibilities on a dataset $\cX=\{\x_i\}_N \subset \R^d$. The \emph{expected value of the complete-data log-likelihood} function of the model, also called the $\cQ$-function~\cite{Bishop2006} is:
    \begin{align}\label{eq:Q}
    \begin{split}
    \cQ(\bm\Gamma,\bm\Phi,\bmu)  = \sum_{i=1}^N \sum_{k=1}^K \gamma_{ik} \big( \log \phi_k - ||\bm x_i - \bmu_k||^2 \big)  + \sum_{k=1}^K(\alpha_k-1) \log \tilde{\bm \gamma}_k.
    \end{split}
    \end{align}
    {In this expression,}
    $z_i \in \N$ is the cluster assignment of the data point $\x_i$, 
    $\gamma_{ik}:= p(z_i=k | \x_i)$ is the posterior probability of $z_i=k$, also called the responsibility of cluster $k$ on a data-point $\x_i$, 
    $\phi_{k}:= p(z_i=k) $ is the prior probability or mixture weight of cluster $k$ and $ \bmu_k \in \R^{ d}$ is the centroid of cluster $k$. 
    The average responsibilities of cluster $k$ is $\tilde{ \gamma}_k = \frac{1}{N} \sum_{i=1}^N \gamma_{ik}$ and $\tilde{\bm \gamma} \in \S^K$.
    The concentration hyperparameter of the Dirichlet prior is $\< \alpha_k \>_K \in \R^K \setminus \{ {\bm 0_K} \}$.
    The co-variance matrices do not appear {in this expression} as they are all constant and equal to $\frac{1}{2} \bI_d$ due to the isotropic assumption.
    We summarize the parameters into matrices $\bm \Gamma \in \R^{N \times K}$, $\bm\Phi \in \S^{K}$ and $\bmu \in \R^{K \times d}$.
    Note that the cluster responsibility vector of $\x_i$ is stochastic, i.e., ${\bm \gamma}_i = \< \gamma_{ik} \>_K \in \S^K$.
    
    Although, the isotropic co-variances allow for great simplifications, they restrict the model to spherical Gaussian distributions. We later alleviate this constraint by introducing a deep autoencoder.
    The Dirichlet prior involves an extra parameter but also smoothens the optimization.
    Note that these two assumptions make the model a relaxation of the one underlying $k$-means~\cite{kulis2012revisiting,lucke2019k}.

    \begin{theorem} \label{th1}
    The maximization of the $\cQ$-function in Eq.~\eqref{eq:Q} with respect to $\bm \Phi$ yields a term that can be interpreted as the reconstruction loss of an autoencoder. 
    \end{theorem}
    
    \begin{proof}
    During the EM-algorithm, the maximization   with respect to $\bm\Phi$ updates the latter as the average cluster responsibility vector:
    \begin{align*}
    \phi_k = \frac{1}{N} \sum_{i=1} \gamma_{ik} = \tilde{ \gamma}_k.
    \end{align*}
    Using this result, the $\cQ$-function can be rephrased as
    \begin{align}\label{eq:Q}
    \cQ(\bm\Gamma,\bmu)  = \sum_{i=1}^N \sum_{k=1}^K \gamma_{ik} \log \tilde{\bm \gamma}_k - \sum_{i=1}^N \sum_{k=1}^K \gamma_{ik} ||\bm x_i - \bmu_k||^2   + \sum_{k=1}^K(\alpha_k-1) \log \tilde{\bm \gamma}_k.
    \end{align}
    The first term corresponds to the entropy of $\<\tilde{ \gamma}_k\>_K$ which, given the Dirichlet prior, is constant and can thus be omitted. 
    We expand now $\gamma_{ik}\| \bm x_i - \bmu_k \|^2$  by adding and subtracting the norm of  $\tx_i = \sum_{k=1}^K \gamma_{ik} \bmu_k$: For any given $i \in [1 \isep N]$,
    \begin{align*}
    \begin{split}
    \sum_{k=1}^K   \gamma_{ik}||\x_i  -  \bmu_k||^2 &=  \sum_{k=1}^K \gamma_{ik}||\x_i||^2   - 2  \x_i^\top \Big(\sum_{k=1}^K \gamma_{ik}\bmu_k\Big) + \sum_{k=1}^K \gamma_{ik}||\bmu_k||^2 \\
    & \hspace*{3cm}    +  ||\tx_i||^2  -  ||\tx_i||^2\\
    & = ||\x_i||^2 - 2 \x_i^\top \tx_i + ||\tx_i||^2 \\ 
    & \hspace*{3cm} +  \sum_{k=1}^K \gamma_{ik}||\bmu_k||^2  -  \sum_{k=1}^K \sum_{l=1}^K \gamma_{ik}\gamma_{il} \bmu_k^T \bmu_k \\
    &=  ||\x_i  \!-\!  \tx_i||^2  \!+\!   \sum_{k=1}^K \gamma_{ik}(1 \!-\! \gamma_{ik} )||\bmu_k||^2 \!-\!  \sum_{k=1}^K \sum_{\substack{l=1 \\ l\neq k}}^K \gamma_{ik}\gamma_{il} \bmu_k^T \bmu_l.
    \end{split}
    \end{align*}
    Note that this simplification  grounds on the isotropic assumption. It is not obvious how to reach a similar result without it. The $\cQ$-function becomes:
    \begin{align}\label{eq:Q1}
    \begin{split}
    \cQ(\bm\Gamma,\bmu) &= -    \underbrace{ \sum_{i=1}^N ||\x_i  -  \tx_i||^2}_{=:E_1}
    - \underbrace{ \sum_{i=1}^N \sum_{k=1}^K  \gamma_{ik}( 1 -  \gamma_{ik} )||\bmu_k||^2}_{=:E_2} \\
    & + \underbrace{ \sum_{i=1}^N \sum_{k=1}^K \sum_{\substack{l=1 \\ l\neq k}}^K \gamma_{ik}\gamma_{il} \bmu_k^T \bmu_l }_{=:E_3}
    - \underbrace{\sum_{k=1}^K (1-\alpha_k ) \log\tilde{ \gamma}_k}_{=:E_4}.
    \end{split}
    \end{align}
    The variable $\tx_i$ is actually a function of $\x$ that factorizes into two functions $\Enc$ and $\Dec$.
    The former computes $\bm\gamma_{i}$ which, as a posterior probability, is a function of $\x_i$. The latter is the dot-product with $\bmu$.
    \begin{align}\label{eq:rec}
    \begin{split}
    \Enc: &\begin{array}{ccl} \R^d & \rightarrow & \S^K \\ \x & \mapsto & \bm \Enc( \x; {\bm \eta} )= \< p(z=k| \x) \>_K = \bm \gamma\end{array} \\
    \Dec: &\begin{array}{ccl} \S^K & \rightarrow & \R^d \\ \bm\gamma & \mapsto & \bm \Dec( \bm\gamma; {\bm \mu} )= \sum_{k=1}^K \gamma_{k} \bmu_k = \tx  \end{array} \\
    \end{split}
    \end{align}
    The first term of {Eq.~\eqref{eq:Q1}} can thus be interpreted as the reconstruction term characteristic of a loss of an autoencoder, {consisting of the} encoder and decoder functions, which are $\Enc$ and $\Dec$, respectively.
    \qed
    \end{proof}    
    
    \subsection{Analysis of the terms}\label{sec:loss-terms}
    The four terms of the expansion of the $\cQ$-function as Eq.~\eqref{eq:Q1} gives new insights into the training of GMMs. 
    
    \paragraph{$E_1$: Reconstruction} 
    The term $E_1$ suggests an autoencoder structure to optimize $\cQ$.
    The decoder, $\Dec$, is a linear function. However, without loss of generality, it can be treated in practice as an affine function, i.e. a single layer network with bias.
    Regarding the encoder $\Enc$, the architecture can be anything as long as it has a stochastic output.

    \paragraph{$E_2$: Sparsity and regularization}
    
    The second term $E_2$ is related to the Gini impurity index~\cite{cart84} applied to $\bm \gamma_{i}$:
    \begin{align*}
    \begin{split}
    \sum_{k=1}^K  \gamma_{ik} ( 1 - \gamma_{ik} )\|\bmu_k\|^2  \leq  \sum_{k=1}^K  \gamma_{ik} ( 1 - \gamma_{ik} ) \|\bm\mu\|^2_F =  {\bf Gini}(\bm \gamma_{i}) \|\bm\mu\|^2_F,  
    \end{split}
    \end{align*}
    where $\|.\|_F$ is the Frobenius norm.
    The Gini index is standard in decision tree theory to select branching features and is an equivalent of the entropy.
    It is nonnegative and null if, and only if, $\bm \gamma_i$ is a one-hot vector.
    Hence, minimizing this term favors sparse $\bm \gamma_i$  resulting in clearer assignments. 
    
    The terms $\|\bmu_k\|^2$ play a role similar to an $\cl_2$-regularization: they prevent the centroids from diverging away from the data-points. 
    However, they may also favor the trivial solution where all the centroids are merged into zero.
    
    \paragraph{$E_3$: Sparsity and cluster merging}
    To study the behavior of $E_3$ during the optimization, let us consider a simple example with one observation and two clusters, i.e, $\bm \Gamma \equiv \bm\gamma_{1}=(\gamma, 1-\gamma)$.
    If the observation is unambiguously assigned to one cluster, $\bm\gamma_{1}$ is a one-hot vector and $E_3$ is null. If it is not the case, the difference between $E_2$ and $E_3$ factorizes as follows:
    \begin{align}
    E_2-E_3 = \gamma(1-\gamma) \big( \|\bmu_1\|^2 + \|\bmu_2\|^2 - \bmu_1^T \bmu_2 \big) = \gamma(1-\gamma) \| \bmu_1 - \bmu_2\|^2.
    \end{align}
    The optimization will thus either push  $\bm\gamma_{1}$ toward a more sparse vector, or merge the two centroids. 
    Appendix~\ref{sec:cm-exp-e3} presents an analysis of the role of this term.

    \paragraph{$E_4$: Balancing}
    The Dirichlet prior steers the distribution of the cluster assignments. 
    If none of the $\alpha_k$ is null, the prior will push the optimization to use all the clusters, moderating thus the penchant of $E_2$ for the trivial clustering~\cite{yang2016towards,guo2017improved}. 
    
    Note that if ${\bm \alpha} = \left(1 + \frac{1}{K}\right) {\bm 1}_K$, $E_4$ is, up to a constant, equal to the Kullback-Leibler (KL) divergence between a multinomial distribution with parameter $\tilde{\bm \gamma}$ and the uniform multinomial distribution:
    \begin{align*}
    \begin{split}
    \sum_{k=1}^K (1-\alpha_k ) \log \tilde{\bm \gamma}_k = \sum_{k=1}^K (1 - (1 + \frac{1}{K})) \log \tilde{\bm \gamma}_k = D_{KL} \left(\frac{1}{K} {\bm 1}_K \Big\| \tilde{\bm \gamma} \right).
    \end{split}
    \end{align*}

    \section{Clustering and Embedding with Autoencoders}\label{sec:cm}
    
    Theorem~\ref{th1} says that an autoencoder could be involved during the EM optimization of an isotropic GMM.
    We go one step further and by-pass the EM to directly optimize the $\cQ$-function in Eq.~\eqref{eq:Q1}) using an autoencoder.

    \subsection{The Clustering Module}
    
    We define the Clustering Module (CM) as the one-hidden layer autoencoder with encoding and decoding functions $\Enc$ and $\Dec$ such as:
    \begin{align}\label{eq:cm-fun}
    \begin{split}
     \Enc(\X) & =  \Softmax(\X \W_{\text{enc} } + \B_{\text{enc}}) = {\bm \Gamma} \\
     \Dec({\bm \Gamma}) & = {\bm \Gamma} \W_{\text{dec}}  + \B_{\text{dec}}=\tX,
    \end{split}
    \end{align}
    where $\X \in \R^{N \times d} \sim \cX $,  code representation/cluster responsibilities $\bm \Gamma  = \< \gamma_{ik} \>_{N \times K} \in \R^{N \times K}$ s.t. $\bm \gamma_{i} \in \S^K$, and the reconstruction $\tX \in \R^{N \times d}$.
    The weight and bias parameters of the encoder are $\W_{\text{enc}} \in \R^{d \times K}$ and $\B_{\text{enc}} \in \R^{K}$, respectively, and analogously for the decoder $\W_{\text{dec}} \in \R^{K \times d}$ and $\B_{\text{dec}} \in \R^{d}$.
    The $\Softmax$ enforces the row-stochasticity of the code, ie., $\bm\Gamma$.
    The associated loss function is the negative of Eq.~\eqref{eq:Q1}:
    \begin{align}\label{eq:loss-cm}
    \begin{split}
    \cL_{\text{CM}}(\cX;\Theta) & =  { \sum_{i=1}^N ||\x_i  +  \tx_i||^2}
    + { \sum_{i=1}^N \sum_{k=1}^K  \gamma_{ik}( 1 -  \gamma_{ik} )||\bmu_k||^2} \\
    & - { \sum_{i=1}^N \sum_{k=1}^K \sum_{\substack{l=1 \\ l\neq k}}^K \gamma_{ik}\gamma_{il} \bmu_k^T \bmu_l }
    + {\sum_{k=1}^K (1-\alpha_k ) \log\tilde{ \gamma}_k}.
    \end{split}
    \end{align}
    with $\Theta=\Big(\W_{\text{enc}},\B_{\text{enc}},\W_{\text{dec}},\B_{\text{dec}}\Big)$.
    The centroids of the underlying GMM correspond to the images of $\Dec$ of the canonical basis of $\R^K$.

    \paragraph{Initialization}\label{sec:cm-init}
    
    The CM can be initialized using $k$-means or any initialization scheme thereof such as \emph{$k$-means++}~\cite{arthur2007k}. 
    In such a case, the column-vectors of $\W_{\text{dec}}$ are set equal to the desired centroids.
    The pseudo-inverse of this matrix becomes the encoder's weights, $\W_{\text{enc}}$, and both bias vectors are set to null.
    
    \paragraph{Averaging epoch}\label{sec:cm-avg}
    \begin{wrapfigure}{r}{.5\textwidth}
        \vspace{-2em}
        \includegraphics[width=\linewidth]{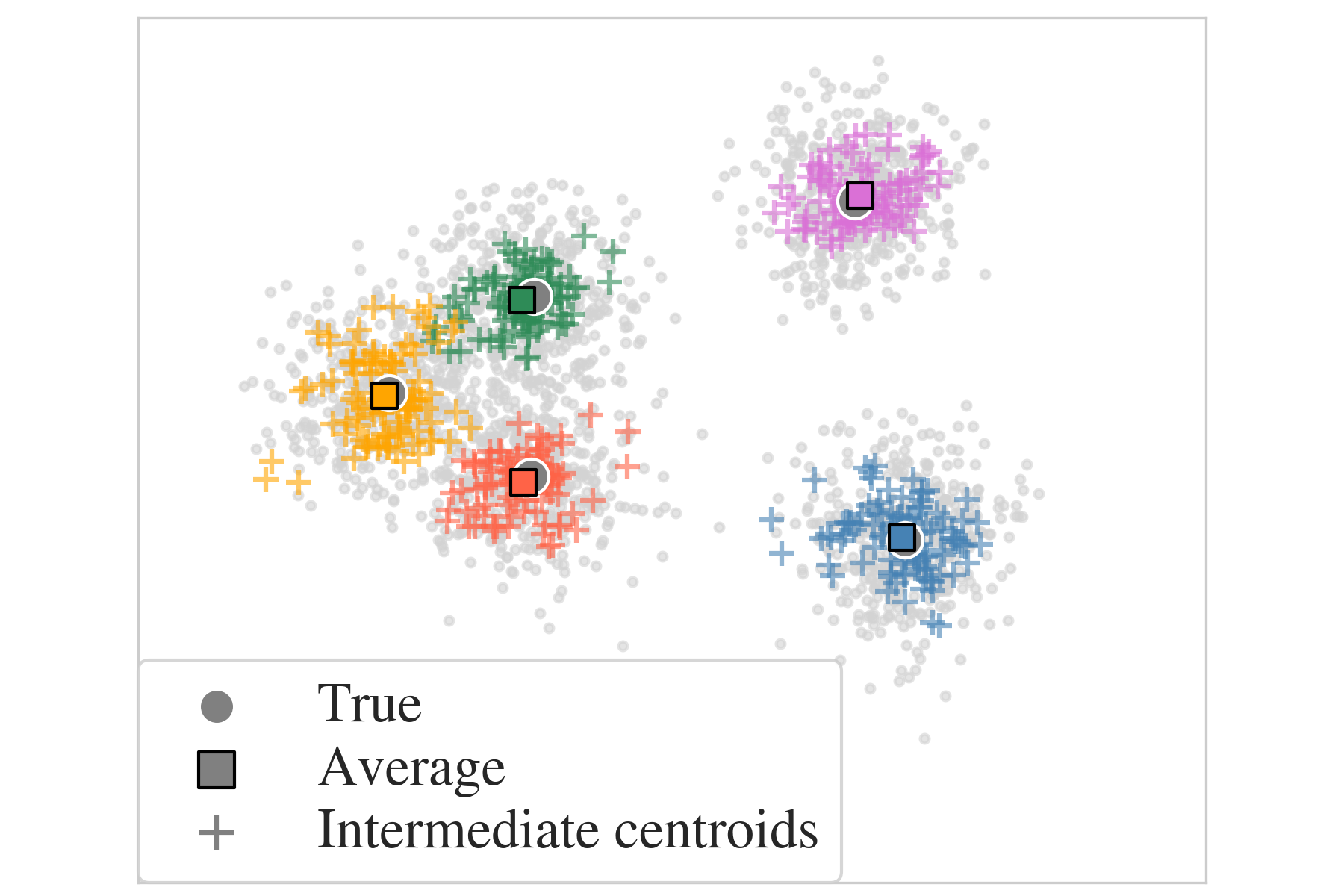}
        \caption{The intermediate centroids of the last epoch are spread, whereas their averages almost match the true centroids.}
        \label{fig:cm-avg}
    \end{wrapfigure}
    
    In practice, {the CM will be optimized by mini-batch learning with stochastic gradient descent. In such a procedure,} the optimizer updates the positions of the centroids given the current batch.
    A small batch-size relative to the size of $\cX$ may cause dispersion of the intermediate centroids. 
    Hence, choosing the final centroids based on the last iteration may be sub-optimal.
    
    We illustrate this phenomenon in Figure~\ref{fig:cm-avg}.
    The data consists of $N = 2,000$ points in $\R^2$ drawn from a mixture of five bi-variate Gaussians ($K = 5$) (gray dots). 
    The data is standardized before processing.
    A CM is trained in mini-batches of size $20$ over $50$ epochs using stochastic gradient descent. 
    The concentration is set to $\bm\alpha = {\bm 5}_K$.
    The dispersion of the centroids after each iteration of the last epoch (crosses) is significant.
    On the other hand, their average positions (squares) provide a good approximation of the true centers (circles).
    Therefore, we include one extra epoch to any implementation of the CM to compute the average position of the individual centroids over the last iterations. 
    
    \subsection{Embedding with feature maps}\label{sec:cnet}

    The theory behind GMMs limits the CM to a linear decoder, thus {enabling merely} a linear partition of the input space.
    {In addition,} the isotropy assumption, specific to the CM, bars clusters to spread differently.
    We {alleviate} both \emph{limitations} using a similar approach to that of kernel methods~\cite{kernel-trick}:
    we {non-linearly} project the input into a feature space where it will be clustered.
    However, we do not learn the kernel matrices, {providing an implicit feature map.} {Instead, we learn explicitly} the feature maps using a deep autoencoder (DAE).
    
    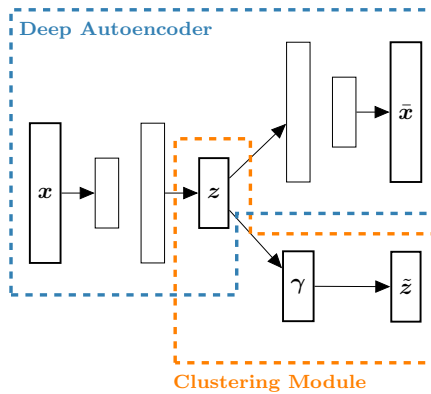
\begin{wrapfigure}[18]{r}{0.5\linewidth}
        \vspace{-2em}
        \begin{minipage}{.9\linewidth}
            \centering
        \scalebox{.9}{
            \begin{tikzpicture}
            \node (x)  at (0,0) [rectangle,draw,thick,minimum width=1em,minimum height=6em]  {$\x$};
            \node (e1) at (0,0) [rectangle,draw, right=of x ,xshift=-1.5em, minimum width=1em,minimum height=3em]  {};
            \node (e2) at (0,0) [rectangle,draw, right=of e1,xshift=-2.em, minimum width=1em,minimum height=6em]  {};
            
            \node (z)  at (0,0) [rectangle,draw,thick,right=of e2,xshift=-1.5em, minimum width=1em,minimum height=3em]  {$\bz$};
            
            \node (d2) at (0,0) [rectangle,draw, above right=of z,xshift=-.5em,yshift=-4em, minimum width=1em,minimum height=6em]  {};
            \node (d1) at (0,0) [rectangle,draw, right=of d2,xshift=-2.em, minimum width=1em,minimum height=3em]  {};
            \node (tx)  at (0,0) [rectangle,draw,thick,right=of d1,xshift=-1.5em,minimum width=1em,minimum height=6em]  {$\tx$};
            
            \node (tz) at (0,0) [rectangle,draw, thick,below=of tx,xshift=0em, minimum width=1em,minimum height=3em]  {$\tz$};
            \node (g) at (0,0) [rectangle,draw, thick,below=of d2,xshift=0em,yshift=0em, minimum width=1em,minimum height=3em]  {$\bm \gamma$};
            
            \edge {x}  {e1};
            \edge {e2}  {z};
            
            \edge {z}  {d2};
            \edge {z}  {g};
            
            \edge {d1}  {tx};
            
            \edge {g}  {tz};
            
            \draw [dashed, orange, line width=0.5mm] (1.9,-2.5) -- (1.9,.8) -- (3.,.8) -- (3.,-.6) -- (5.7,-.6) -- (5.7,-2.5) -- (1.9,-2.5);
            \node[right, orange] at (1.75,-2.8) {\small \bf Clustering Module};
            
            \draw [dashed, blue, line width=0.5mm] (-.5,-1.5) -- (2.8,-1.5) -- (2.8,-.3) -- (5.7,-.3) -- (5.7,2.7) -- (-.5,2.7) -- (-.5,-1.5);
            \node[right, blue] at (-.5,2.4) {\small \textbf{Deep Autoencoder}};
            
            \end{tikzpicture}
        }
        {\caption[Schematic representation of AE-CM.]
            {Schematic representation of the  AE-CM. Combining a clustering module and a deep autoencoder allows to jointly learn a clustering and an embedding.}
            
            \label{fig:arch}}
        \end{minipage}
    \end{wrapfigure}

    The idea is to optimize the CM and the DAE simultaneously, in order to let the latter find distortions of the input space along the way that guides the CM toward a better {optimum}. 
    Using a deep autoencoder architecture prevents the optimization to produce degenerate feature maps~\cite{guo2017improved}.
    It also preserves the generative nature of the model: points in the input space can be generated from a combination of centroids in the feature space.
    We refer to this model as the AE-CM. 
    
    The model consists of a clustering module nested into a deep autoencoder
    The architecture is illustrated in Figure~\ref{fig:arch}.
    The first part of the DAE encodes an input $\x\in \R^d$ into a vector $\bz \in \R^p$.
    Note, CM now works on code representation $\bz$ and not directly on the input $\x$.
    The code $\bz$ is fed to the CM and to the decoder of the DAE, yielding two outputs: $\tz$, the CM's reconstruction of $\bz$, and $\tx$, the DAE's reconstruction of $\x$. 
    
    \subsubsection{Adapting the loss function to a deep architecture}
    
    Empirical evaluation showed that current gradient descent optimizers (e.g., Adam \cite{kingma2014adam}) often return sub-optimal solutions when the reconstruction of the deep autoencoder is simply added to the loss of the CM.
    To help the optimization to find better optima, we add the assumption that the centroids are orthonormal:
    \begin{align}\label{eq:orth}
    \forall k,l , \: \bmu_k^T \bmu_l = \delta_{kl} = \left\lbrace  \begin{array}{cl} 1 & \text{if } k=l, \\ 0 & \text{otherwise}.\end{array} \right.
    \end{align}
    Although the previous formula involves only $\bmu$ which is learned by the nested clustering module, it affects the surrounding DAE.
    Indeed, the constraint encourages it to produce an embedding where the centroids can simultaneously be orthonormal and minimize CM's loss.
    As a consequence, $E_2$ simplifies and $E_3$ becomes null:
    \begin{align}\label{eq:orth-csq}
    \begin{split}
    E_2 & = \sum_{i=1}^N \sum_{k=1}^K  \gamma_{ik}( 1 -  \gamma_{ik} )||\bmu_k||^2 = \sum_{i=1}^N \sum_{k=1}^K  \gamma_{ik}( 1 -  \gamma_{ik} ) \\
    E_3 & = \sum_{i=1}^N \sum_{k=1}^K \sum_{\substack{l=1 \\ l\neq k}}^K \gamma_{ik}\gamma_{il} \bmu_k^T \bmu_l = 0
    \end{split}
    \end{align}
    {Note that the constraint Eq.~\eqref{eq:orth} is satisfied for the "ideal" clustering, in which case clusters will be mapped to the corners of a simplex in the embedding space, as discussed recently in \cite{kampffmeyer2017deep}. In this perspective, our inclusion of this constraint helps guide the clustering towards the ideal clustering, and at the same time simplifies the loss function.}
    
    We employ Lagrange multipliers to integrate the orthonormality constraint.
    That way the final loss can be stated as follows:
    \begin{align}\label{eq:loss-cnet}
    \begin{split}
    \cL_{\operatorname{AE-CM}}(\cX;\Theta)& = \beta \sum_{i=1}^N  ||\x_i - \tx_i||^2 \hspace{9em} \text{(Reconstruction DAE)}  \\ 
    & + \sum_{i=1}^N ||\bz_i  -  \tz_i||^2 \hspace{11em} \text{(Reconstruction CM)}\\
    & + \sum_{i=1}^N \sum_{k=1}^K \gamma_{ik} ( 1 - \gamma_{ik} ) \hspace{13.5em} \text{(Sparsity)} \\
    & + \sum_{k=1}^K (1-\alpha_k)\log( \tilde{\bm \gamma}_k ) \hspace{10.5em} \text{(Dirichlet Prior)} \\ 
    & + \lambda || \bmu^T \bmu - \bI_K ||_1,  \hspace{11.5em} \text{(Orthonormality)}
    \end{split}
    \end{align}
    where $\lambda > 0$ is the Lagrange multiplier and $\beta > 0$ weights the DAE's reconstruction loss. We choose the $\ell_1$-norm to enforce orthonormality, however other norms can be used. 
    
    Note that if the dimension of the embedding $p$ is larger than the number of centroids $K$, the embedding can always be transformed to satisfy the orthonormality of the centroids. 
    On the other hand, if $K > p$, the assumption becomes restrictive also in term of possible clusterings. 
    Nevertheless, its importance can be reduced with a small $\lambda$.
    This assumption also helps to avoid the centroids to collapse as their norm is required to be $1$.
    An analysis of the Lagrange multiplier is provided in Appendix~\ref{sec:cnet-exp-hp}.
    
    
    The loss function of the AE-CM {thus} depends on four hyper-parameters: the weight $\beta>0$, the concentration $\balpha \in \S^K$, the Lagrange multiplier $\lambda>0$, and the size of the batches $B \in \N^*$. 

    \subsubsection{Implementation details}
    \label{sec:cnet-impl}

    
    Since the AE-CM builds upon the CM, any implementation also contains an averaging epoch.
    In case of pre-training, both sub-networks need to be initialized. 
    We favor a straightforward end-to-end training of the DAE (without drop-out or noise) over a few epochs.
    The clustering module is then initialized using $k$-means++ on the embedded dataset.
    Finally, the CM is optimized alone using $\cL_{\text{CM}}$ for a few epochs.
    

    \section{Experiments}\label{sec:exp}

    In this section, we evaluate the clustering module and the AE-CM on several data sets covering different types of data.
    To highlight the generality of our method, we rely only fully connected architecture, ie. we do not use convolution layers even for image data sets.
    That said, we focus on general purpose baselines.
    The experiments were conducted on an Intel(R) Xeon(R) CPU E5-2698 v4 @ 2.20GHz with 32Gb of RAM supported with a NVIDIA(R) Tesla V100 SXM2 32GB GPU.
    
    \subsection{Experimental Setting}\label{sec:cm-exp-set}
    
    \subsubsection{Datasets} 
    We {leverage} eight common benchmark data sets in the deep clustering literature plus one synthetic data set: 
    \begin{itemize}
    \item[]  {\bf MNIST}~\cite{MNIST} contains $70, 000$ handwritten images of the digits 0 to 9. The images are grayscale with the digits centered in the $28\times 28$ images. The pixel values are normalized before processing.\\
    \item[]  {\bf fMNIST}~\cite{FMNIST} contains $70, 000$ images of fashion products organized in $10$ classes. The images are grayscale with the product centered in the $28\times 28$ images. The pixel values are normalized before processing.\\
    \item[]  {\bf USPS}\footnote{\url{https://github.com/XifengGuo/IDEC/files/1613386/usps.zip}} contains $9, 298$ images of digits 0 to 9. The images are grayscale with size $28\times 28$ pixels. The pixel values are normalized before processing.\\
    \item[]  {\bf CIFAR10}~\cite{CIFAR10} contains $60, 000$ color images of 10 classes of subjects (dogs, cats, airplanes...). Images are of size $32 \times 32$. The pixel values are normalized before processing.\\
    \item[]  {\bf Reuters10k}\footnote{\url{https://github.com/XifengGuo/IDEC/tree/master/data/reuters}}, here abbreviated {\bf R10K}, consists of $800, 000$ news {articles}. The dataset is pre-processed as in \cite{guo2017improved} to return a subset of $10, 000$ random samples embedded into a $2, 000$-dimensional space (tf-idf transformation) and distributed over $4$ (highly) {imbalanced}  categories.\\
    \item[]  {\bf 20News}\footnote{\url{http://people.csail.mit.edu/jrennie/20Newsgroups}} contains $18, 846$ messages from newsgroups on 20 topics. Features consists of the tf-idf transformation of the $2,000$ most frequent words.\\
    \item[]  {\bf 10x73k}~\cite{10X73K} consists of $73,233$ RNA-transcript belonging to 8 different cell types \cite{jang2017categorical}. The features consists of the $\log$ of the gene expression variance of the 720 genes with the largest variance. The dataset is relatively sparse with $40\%$ of entries null.\\
    \item[]  {\bf Pendigit}~\cite{Pendigit} consists of $10, 992$ sequences of coordinates on a tablet captured as writers write digits, thus 10 classes. The dataset is normalized before processing.\\
    \item[]  {\bf 5 Gaussians} consists of $N = 2,000$ points in $\R^2$ drawn from a mixture of five bi-variate Gaussians ($K = 5$). The dataset is depicted in Figure~\ref{fig:cm-avg}.
    \end{itemize}

    
    \subsubsection{Evaluation metrics}
    The clustering performance of each model is evaluated using three frequently-used metrics: the Adjusted Rand Index (ARI)~\cite{ARI}, the Normalized Mutual Information (NMI)~\cite{NMI}, and the clustering accuracy (ACC)~\cite{ACC}. These metrics range between $0$ and $1$ where the latter indicates perfect clustering.
    For legibility, values are always multiplied by $100$.
    For each table, scores not statistically different ($t$-test $p<0.05$) from the best score of the column are marked in boldface. 
    The $*$ indicates the model with the best run.
    A failure ($-$) corresponds to an ARI close to $0$.
    
    \subsubsection{Baselines}
    
    We include two baselines for the CM: $k$-means (KM), a GMM with full co-variance and an isotropic GMM (iGMM) with uniform mixture weights.
    The latter differs from the model the clustering module derives but the Dirichlet prior on the responsibilities yields non tractable updates and a Dirichlet prior on the mixture weights harms the performance.
    
    We compare the AE-CM to four baselines reviewed in Section~\ref{sec:related-work}: 
    DEC~\cite{xie2016unsupervised}, its extension IDEC~\cite{guo2017improved}, DCN~\cite{yang2016towards} and DKM~\cite{fard2020deep}. 
    We include as the naive approach (AE+KM) consisting of a trained DAE followed by $k$-means on the embedding.
    We also add ClusterGAN~\cite{mukherjee2019clustergan} and VIB-GMM~\cite{uugur2020variational} as alternatives based on variational autoencoders~\cite{kingma2013auto} and generative adversarial networks~\cite{goodfellow2014generative}, {respectively}. 
    
    Random and pre-trained initialization are indicated with \emph{${}^r$} and \emph{${}^p$}, respectively. If omitted, the initialization is random.
    Every experiment is repeated $20$ times. 
    
    \subsubsection{Implementation}
    
    Both CM and AE-CM are implemented using TensorFlow 2.1~\cite{abadi2016tensorflow}\footnote{Code available at: \\ \url{https://github.com/Ahcene-B/clustering-Module}}. 
    We also re-implemented DEC, IDEC and DCN.
    All deep models but ClusterGAN and VIB-GMM use the same fully connected autoencoder $d$-500-500-2000-$p$-2000-500-500-$d$ and leaky relu activations, where $d$ and $p$ are the input and feature space dimensions, respectively. 
    For ClusterGAN and VIB-GMM, we used the architecture provided in the original code.
    As well, the DAE reconstruction loss is the mean square error, regardless of the dataset and of the model, except for VIB-GMM on images which requires a cross-entropy loss (it under performs, otherwise).
    CM and its baselines are trained for up to 150 epochs, deep models for 1000 epochs.
    The hyper-parameters (batch-size, $p$, concentration, etc.) are listed in Tables~\ref{apx-tab:cm-exp-setting} and \ref{apx-tab:cnet-exp-setting}.

\subsection{CM: Evaluation}\label{sec:cm-exp-set}

    Recall that the loss of the clustering module is a lower bound of the objective function of its underlying isotropic GMM which approximates $k$-means.
    Moreover, the optimization of the CM is based on gradient descent instead of EM. 
    We compare these three models {as a sanity check,}  and show that, despite the differences, they report similar clustering performance.
    We also present an ablation study of the loss and a model selection scheme for the CM.
    An analysis of the hyper-parameters and of the Dirichlet prior are reported in Appendix~\ref{sec:cm-exp-hp} and \ref{sec:cm-exp-prior}, respectively.


    \subsubsection{CM: Clustering performance} \label{sec:cm-exp-perf}
    
    In this experiment, we compare clustering performances and initialization schemes.
    Random and $k$-means++ {initializations}  are indicated with the {superscripts} \emph{${}^r$} and \emph{${}^p$}, respectively.
    Each experiment is repeated $20$ times. We report averages {in}  Table~\ref{tab:cm-exp-results}.
    For each dataset, scores not statistically significant different from the highest ($p<.05$) {score} are marked in boldface.
    The $*$ indicate the model with the highest best score among its 20 runs.
    An extended table including standard deviations and best run can be found in Table~\ref{apx-tab:cm-exp-results-full} of Appendix~\ref{sec:cm-exp-results-full}.
    Average runtime for MNIST are reported on Table~\ref{tab:cm-exp-runtime}.
    
    \begin{table*}[t!]
        \caption[Comparison of clustering performance in terms of mean ARI.]
        {The clustering performance ($\times 100$) of different models on the selected datasets.  
        }\label{tab:cm-exp-results}
        \scriptsize
        \renewcommand{\arraystretch}{1.2}
        \centering
        
        \noindent
        \begin{tabular*}{.99\linewidth}{@{\extracolsep{\fill}}p{0.13\textwidth}p{0.025\textwidth}p{0.025\textwidth}p{0.04\textwidth}p{0.025\textwidth}p{0.025\textwidth}p{0.04\textwidth}p{0.025\textwidth}p{0.025\textwidth}p{0.04\textwidth}p{0.025\textwidth}p{0.025\textwidth}c}
            \toprule
            \multirow{2}{*}{Model} & \multicolumn{3}{c}{MNIST} & \multicolumn{3}{c}{fMNIST} & \multicolumn{3}{c}{USPS} & \multicolumn{3}{c}{CIFAR10} \\
            & ARI & NMI & ACC & ARI & NMI & ACC & ARI & NMI & ACC & ARI & NMI & ACC \\
            \midrule
            
KM${}^r$ & $ 37.8$ & $\mathbf{ 49.9}^*$ &$ 54.5^*$ &$ 36.6$ &$ 51.6$ &$ 53.2$ &$ 52.6$ & $\mathbf{ 61.8}$ &$ 63.0$ &$ 4.2$ &$ 8.1$ &$ 20.8$   \\
KM${}^p$ & $ 36.9$ & $\mathbf{ 49.2}$ &$ 54.0$ &$ 35.2$ &$ 51.0$ &$ 50.8$ &$ 50.2$ &$ 60.8$ &$ 61.1$ &$ 4.2$ &$ 8.1$ &$ 20.7$   \\ \midrule

GMM${}^r$ & $ 23.2$ &$ 37.8$ &$ 40.3$ &$ 34.3$ &$ 49.3$ &$ 51.8$ &$ 35.1$ &$ 52.5$ &$ 48.2$ & $\mathbf{ 5.0}$ &$ 9.1$ & $\mathbf{ 22.5}^*$   \\
GMM${}^p$ & $ 24.8$ &$ 37.5$ &$ 42.3$ &$ 34.3$ &$ 49.3$ &$ 52.4$ &$ 33.0$ &$ 52.0$ &$ 45.1$ & $\mathbf{ 5.1}^*$ & $\mathbf{ 9.3}$ & $\mathbf{ 22.4}$   \\ \midrule

iGMM${}^r$ & $ 31.3$ &$ 42.7$ &$ 48.5$ &$ 35.7$ &$ 50.7$ &$ 51.4$ &$ 44.2$ &$ 55.3$ &$ 56.2$ &$ 4.1$ &$ 7.9$ &$ 21.1$   \\
iGMM${}^p$ & $ 31.1$ &$ 42.7$ &$ 47.5$ &$ 35.4$ &$ 50.8$ &$ 51.6$ &$ 44.4$ &$ 56.1$ &$ 55.6$ &$ 4.1$ &$ 7.8$ &$ 21.1$   \\ \midrule

CM${}^r$ &  $\mathbf{ 39.7}^*$ & $\mathbf{ 50.0}$ & $\mathbf{ 56.8}$ & $\mathbf{ 42.3}$ & $\mathbf{ 54.0}$ & $\mathbf{ 62.0}$ & $\mathbf{ 54.3}$ & $\mathbf{ 62.8}$ & $\mathbf{ 67.0}^*$ &$ 4.8$ &$ 8.9$ & $\mathbf{ 22.0}$   \\
CM${}^p$ &  $\mathbf{ 39.1}$ & $\mathbf{ 49.5}$ & $\mathbf{ 55.9}$ & $\mathbf{ 41.4}^*$ & $\mathbf{ 53.4}^*$ & $\mathbf{ 60.7}^*$ & $\mathbf{ 53.4}^*$ & $\mathbf{ 62.8}^*$ &$ 63.7$ &$ 4.9$ & $\mathbf{ 9.3}^*$ & $\mathbf{ 22.3}$   \\
            
            \bottomrule
            \multirow{2}{*}{Model} & \multicolumn{3}{c}{R10K} & \multicolumn{3}{c}{20News} & \multicolumn{3}{c}{10x73k} & \multicolumn{3}{c}{Pendigit} \\
            & ARI & NMI & ACC & ARI & NMI & ACC & ARI & NMI & ACC & ARI & NMI & ACC \\
            \midrule
            
KM${}^r$ &  $\mathbf{ 33.8}$ &$ 38.1$ & $\mathbf{ 61.2}$ &$ 14.8$ &$ 32.3$ &$ 31.0$ &$ 36.5$ &$ 55.4$ &$ 55.0$ & $\mathbf{ 56.5}^*$ &$ 67.8$ & $\mathbf{ 71.1}$   \\
KM${}^p$ & $ 29.5$ &$ 36.0$ &$ 58.3$ &$ 14.8$ &$ 33.5$ &$ 32.0$ &$ 36.7$ &$ 55.5$ &$ 55.3$ & $\mathbf{ 58.1}$ &$ 68.9$ & $\mathbf{ 72.7}$   \\ \midrule

GMM${}^r$ & $ 13.5$ &$ 18.4$ &$ 46.8$ &$ 13.2$ &$ 34.6$ &$ 31.6$ &$ 32.2$ &$ 50.4$ &$ 52.4$ &$ 51.3$ &$ 68.4^*$ &$ 65.7^*$   \\
GMM${}^p$ & $ 11.8$ &$ 14.8$ &$ 47.5$ &$ 10.7$ &$ 31.2$ &$ 27.2$ &$ 32.0$ &$ 50.8$ &$ 51.5$ &$ 54.3$ & $\mathbf{ 69.7}$ & $\mathbf{ 71.1}$   \\ \midrule

iGMM${}^r$ &  $\mathbf{ 39.8}$ & $\mathbf{ 44.2}$ & $\mathbf{ 66.9}$ & $\mathbf{ 18.4}$ & $\mathbf{ 41.8}^*$ & $\mathbf{ 37.4}^*$ &$ 34.2$ &$ 53.7$ &$ 53.9$ & $\mathbf{ 58.1}$ & $\mathbf{ 69.0}$ & $\mathbf{ 72.7}$   \\
iGMM${}^p$ & $ 27.3$ &$ 32.3$ &$ 60.4$ &$ 13.4$ &$ 36.8$ &$ 31.4$ &$ 33.1$ &$ 53.1$ &$ 52.1$ & $\mathbf{ 56.9}$ & $\mathbf{ 68.9}$ & $\mathbf{ 72.0}$   \\ \midrule

CM${}^r$ &  $\mathbf{ 38.5}$ & $\mathbf{ 41.0}$ & $\mathbf{ 64.9}$ &$ 9.7$ &$ 21.2$ &$ 18.3$ & $\mathbf{ 54.8}$ & $\mathbf{ 63.8}$ & $\mathbf{ 69.5}$ & $\mathbf{ 57.3}$ &$ 67.0$ & $\mathbf{ 72.0}$   \\
CM${}^p$ & $ 32.6^*$ &$ 39.2^*$ &$ 60.2^*$ &$ 16.3^*$ &$ 28.8$ &$ 30.8$ & $\mathbf{ 55.4}^*$ & $\mathbf{ 64.0}^*$ & $\mathbf{ 71.0}^*$ & $\mathbf{ 57.3}$ &$ 66.9$ & $\mathbf{ 72.3}$   \\
            
            \bottomrule
        \end{tabular*}
    \end{table*}
    
    \begin{table*}[t!]
        \caption
        {Average runtime of each model to cluster MNIST in 150 epochs.}\label{tab:cm-exp-runtime}
        \scriptsize
        \renewcommand{\arraystretch}{1.2}
        \centering
        
        \noindent
        \begin{tabular*}{.5\linewidth}{@{\extracolsep{\fill}}ccccc}
            \toprule
            Model & KM & GMM & iGMM & CM \\ \midrule
            Runtime & 1.7s & 71.5s & 15.9s & 2m44s \\
            \bottomrule
        \end{tabular*}
    \end{table*}
    
    
    As expected, the clustering module performs similarly to iGMM and $k$-means on every dataset with respect to almost every metrics and for any initialization scheme.
    Often the $k$-means++ initialization does not improve the results.
    In the case of the clustering module the difference is never significant except on the 20News dataset.

    \subsubsection{CM: Runtimes}
    
    We compare here the runtimes of the different method on MNIST. For a fair comparison, we do not use any early-topping criterion and all the methods are run for exactly 150 epochs. We report on Table~\ref{tab:cm-exp-runtime} average over 10 runs.
    
    \begin{table*}[t!]
        \caption
        {Average runtime of each model to cluster MNIST in 150 epochs.}\label{tab:cm-exp-runtime}
        \scriptsize
        \renewcommand{\arraystretch}{1.2}
        \centering
        
        \noindent
        \begin{tabular*}{.5\linewidth}{@{\extracolsep{\fill}}ccccc}
            \toprule
            Model & KM${}^r$ & GMM${}^r$ & iGMM${}^r$ & CM${}^r$ \\ \midrule
            Runtime & 1.7s & 2m4s & 56s & 2m44s \\
            \bottomrule
        \end{tabular*}
    \end{table*}

    It appears clearly that EM-based models are much faster. Interestingly GMM is slower than iGMM, although they share the same implementation. This difference is certainly due to the extra computations needed to update the covariance matrices.

    \subsubsection{CM: Ablation study of the loss}\label{sec:cm-abl}
    
    The loss function of the {CM} arises as a whole from the $\cQ$ function of the underlying GMM.
    Nevertheless, {for additional insight,} we {perform} here an ablation study of its terms.
    We train a CM with different combinations of the terms of its original loss (Eq.~\eqref{eq:loss-cm}).
    To highlight the influence of each term, we focus on the \emph{5 Gaussians} dataset (depicted in Figure~\ref{fig:cm-avg}).
    Table~\ref{tab:cm-abl} reports the clustering performance in terms of ARI.
    
    \begin{table*}[!h]
        \vspace{-1.5em}
        \caption{ Clustering performance (ARI) of the CM on the \emph{5 Gaussians} dataset trained with various combination of the terms of its original loss (first line).
        }
        \label{tab:cm-abl}
        \centering
        \renewcommand{\arraystretch}{1.}
        \noindent
        \begin{tabular*}{\linewidth}{@{\extracolsep{\fill}}p{0.1\textwidth}c|cccc|ccc|c}
            \toprule
            $E_0$ & \checkmark & & \checkmark & \checkmark & \checkmark & \checkmark & \checkmark & \checkmark & \checkmark \\
            $E_1$ & \checkmark  & \checkmark &  & \checkmark  & \checkmark  & \checkmark & & & \\
            $E_2$ & \checkmark  & \checkmark  & \checkmark &  & \checkmark &  & \checkmark & & \\
            $E_3$ & \checkmark  & \checkmark  & \checkmark  & \checkmark & & &  & \checkmark \\ \midrule
            ARI & $\mathbf{83.3}$ & $38.2$ & $45.0$ & $66.8$ & $71.4$ & $50.5$ & $46.2$ & $67.6$ & $59.0$ \\
        \bottomrule
        \end{tabular*}
    \end{table*}

    The gap in terms of ARI between the CM trained with the complete loss (first line) and any other combination of its terms confirms the coherence of the loss.
    The model trained without the reconstruction term $E_0$ reports the worse score.
    The ablation of the single terms indicate that the reconstruction term $E_0$ is the most important followed by the sparsity term $E_1$.
    Although the removal of only $E_3$ has the least impact, it is the only term that combined with $E_0$ reports better performance than a loss function made of solely the reconstruction term $E_0$. 
    
    \begin{figure*}[!h]
        \centering
            \includegraphics[width=1\linewidth]{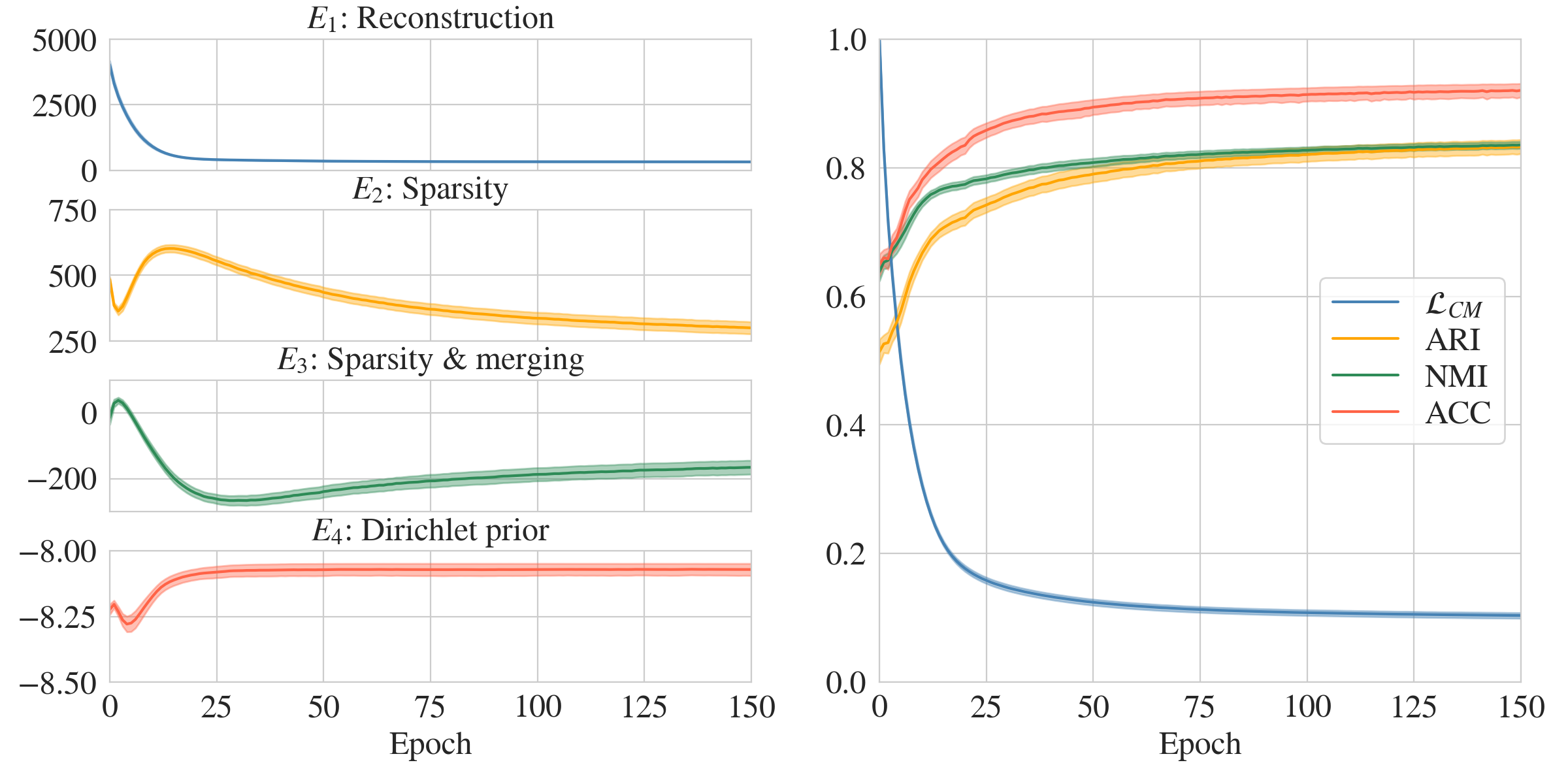}
        \caption{Evolution of the total loss, each of its term and of the clustering metrics during the training of the CM on the 5 Gaussians dataset.}
        \label{fig:cm-abl}
    \end{figure*}

    Figure~\ref{fig:cm-abl} illustrates the behavior of each term of the loss of the CM ($\cL_{CM}$) during the training. 
    Given the variations of each curves, it seems that during the first 25 epochs the optimization focuses on minimizing the reconstruction term even if it implies an increase of the other ones. 
    However, the complete loss curve does not flatten out until $E_3$ reaches its minimum.
    From there, a second phase begins where the total loss and the clustering metrics slowly grow in opposite directions.
    Interestingly, as the metrics increase and the clustering improves, $E_3$ also increases, which is contrary to the expected behavior. 
    On the other hand, $E_2$ which is of the same magnitude as $E_3$ and is also influenced by the sparsity of the assignments, continuously decreases without reaching a minimum.
    Overall, these curves suggest that both $E_2$ and $E_3$ could be used to either stop early the training or selecting the best run.

    \subsubsection{CM: Model selection}\label{sec:cm-exp-loss-perf}
    

    In an unsupervised scenario, true labels are not available. 
    It is thus necessary to have an internal measure to avoid selecting a sub-optimal solution.
    
    There are two natural choices: select either the clustering associated with the lowest loss or the less ambiguous clustering.
    In the first case, the sparsity of the clusters responsibilities $\bm \gamma_i$ might be eclipsed by other aspects optimized by the $\cL_{\text{CM}}$, such as the reconstruction term.
    On the other hand, by selecting only given the sparsity, we may end up always choosing the most degenerate clustering.
    Leaning on the analysis of Figure~\ref{fig:cm-abl}, we propose to use $\cL_{\text{sp}}=E_2+E_3$ which sums both terms of the loss governing the sparsity of the $\bm \gamma_i$s but also involves the norm of the $\bmu_k$s.

    In Table~\ref{tab:cm-loss-corr-best}, we report the ARI of the runs with the lowest $\cL_{\text{sp}}$ for each dataset. For comparison, we also report the average and the largest ARI. Scores selected according to $\cL_{\text{sp}}$ that were higher than the average are marked in boldface.
    
    \begin{table*}[!h]
        \caption{ Adjusted Rand index of the run with lowest $\cL_{\text{sp}}$, the average run and the best run. Score larger than the average are marked in boldface.}
        \label{tab:cm-loss-corr-best}
        \centering
        \renewcommand{\arraystretch}{1.0}
        \noindent
        \begin{tabular*}{.98\linewidth}{@{\extracolsep{\fill}}p{0.1\textwidth}p{0.025\textwidth}p{0.025\textwidth}p{0.04\textwidth}p{0.025\textwidth}p{0.025\textwidth}p{0.04\textwidth}p{0.025\textwidth}p{0.025\textwidth}p{0.04\textwidth}p{0.025\textwidth}p{0.025\textwidth}c}
            \toprule
            & \multicolumn{3}{c}{MNIST} & \multicolumn{3}{c}{fMNIST} & \multicolumn{3}{c}{USPS} & \multicolumn{3}{c}{CIFAR10} \\
            Criterion & $\cL_{{\text{sp}}}$ & Avg. & Best & $\cL_{{\text{sp}}}$ & Avg. & Best & $\cL_{{\text{sp}}}$ & Avg. & Best & $\cL_{{\text{sp}}}$ & Avg. & Best \\
            \midrule
            CM${}^r$  & $\mathbf{42.7}$&$39.7$&$43.5$ & $\mathbf{44.2}$&$42.3$&$44.6$ & $\mathbf{54.8}$&$54.3$&$58.6$ & $\mathbf{4.8}$&$4.79$&$5.07$ \\
            CM${}^p$  & $\mathbf{43.3}$&$39.1$&$43.5$ & $\mathbf{44.0}$&$41.4$&$44.7$ & $\mathbf{55.8}$&$53.4$&$59.2$ & $\mathbf{4.94}$&$4.94$&$5.22$ \\
            \midrule
            & \multicolumn{3}{c}{R10K} & \multicolumn{3}{c}{20News} & \multicolumn{3}{c}{10x73k} & \multicolumn{3}{c}{Pendigit} \\
            Criterion & $\cL_{{\text{sp}}}$ & Avg. & Best & $\cL_{{\text{sp}}}$ & Avg. & Best & $\cL_{{\text{sp}}}$ & Avg. & Best & $\cL_{{\text{sp}}}$ & Avg. & Best \\
            \midrule
            CM${}^r$  & $\mathbf{56.0}$&$38.5$&$56.0$ & $\mathbf{9.92}$&$9.74$&$10.9$ & $54.3$&$54.8$&$57.4$ & $57.0$&$57.3$&$60.5$ \\
            CM${}^p$  & $\mathbf{62.9}$&$32.6$&$62.9$ & $13.1$&$16.3$&$22.0$ & $54.9$&$55.4$&$62.4$ & $56.4$&$57.3$&$60.1$ \\
            \bottomrule
        \end{tabular*}
    \end{table*}

    A model selection based on $\cL_{\text{sp}}$ finds the best runs only for R10K.
    Nevertheless, it selects runs with ARI greater than the average in more than half of the cases.
    In the other cases, the difference to the average score remains below $1$ point of ARI expect for CM${}^p$ on 20News.
    The average absolute difference with the best score is $1.60$ and $3.10$ for CM${}^r$ and CM${}^p$, respectively. Without 20News, on which CM${}^p$ performs the worst, that average difference drops to $2.25$ for CM${}^p$.
    These are satisfying results that substantiates our heuristic that $\cL_{\text{sp}}=E_2+E_3$ can be used as an internal metric for the CM.

    \subsection{AE-CM: Evaluation}\label{sec:cnet-exp}
    
    In this section, we compare the clustering performance of our {novel} deep clustering model AE-CM against a set of baselines.
    We study the robustness of the model with respect to the number of clusters and a model selection scheme.
    We also evaluate the quality of the embeddings through the $k$-means clusterings thereof.
    Finally, we review the generative capabilities of our model.

    \subsubsection{AE-CM: Clustering performance}\label{sec:cnet-exp-perf}
    
    We compare now clustering performances and initialization schemes {for representative} deep clustering models.
    We reports average ARI, NMI and ACC over $20$ runs in Table~\ref{tab:cnet-exp-results}.
    An extended table including standard deviations and best run can be found in Table~\ref{apx-tab:cnet-exp-results-full} of Appendix~\ref{sec:cnet-exp-results-full}.
    
    \begin{table*}[t!]
        \caption
        {The clustering scores ($\times 100$) of {representative}  deep clustering models on the selected datasets. 
        }\label{tab:cnet-exp-results}
        \scriptsize
        \renewcommand{\arraystretch}{1.2}
        \centering
        
        \noindent
        \begin{tabular*}{.99\linewidth}{@{\extracolsep{\fill}}p{0.12\textwidth}p{0.025\textwidth}p{0.025\textwidth}p{0.04\textwidth}p{0.025\textwidth}p{0.025\textwidth}p{0.04\textwidth}p{0.025\textwidth}p{0.025\textwidth}p{0.04\textwidth}p{0.02\textwidth}p{0.018\textwidth}c}
            \toprule
            \multirow{2}{*}{Model} & \multicolumn{3}{c}{MNIST} & \multicolumn{3}{c}{fMNIST} & \multicolumn{3}{c}{USPS} & \multicolumn{3}{c}{CIFAR10} \\
            & ARI & NMI & ACC & ARI & NMI & ACC & ARI & NMI & ACC & ARI & NMI & ACC \\
            \midrule

AE+KM & $ 65.6$ &$ 71.5$ &$ 78.6$ &$ 39.0$ &$ 55.6$ &$ 53.0$ &$ 57.1$ &$ 64.6$ &$ 67.5$ &$ 3.2$ &$ 6.5$ &$ 18.9$   \\ \midrule

DCN${}^r$ & $ 10.1$ &$ 25.6$ &$ 25.4$ &$ 17.0$ &$ 33.5$ &$ 29.0$ &$ 17.9$ &$ 36.5$ &$ 37.9$ &$ 3.2$ &$ 5.9$ &$ 18.0$   \\
DCN${}^p$ & $ 75.6$ & $\mathbf{ 82.5}$ &$ 83.1$ &$ 38.6$ &$ 57.1$ &$ 53.1$ &$ 63.9$ &$ 73.1$ &$ 72.5$ &$ 0.1$ &$ 0.6$ &$ 10.7$   \\ \midrule

DEC${}^r$ & $ 11.1$ &$ 19.0$ &$ 28.9$ &$ 22.9$ &$ 38.1$ &$ 39.2$ &$ 36.3$ &$ 46.9$ &$ 46.8$ &$ 3.1$ &$ 5.7$ &$ 18.6$   \\
DEC${}^p$ & $ 73.8$ &$ 79.0$ &$ 83.1$ &$ 41.9$ &$ 58.6$ &$ 54.8$ & $\mathbf{ 70.0}$ & $\mathbf{ 78.1}$ & $\mathbf{ 76.3}$ &$ 3.1$ &$ 5.6$ &$ 18.2$   \\ \midrule

IDEC${}^r$ & $ 27.5$ &$ 39.0$ &$ 42.5$ &$ 35.2$ &$ 50.8$ &$ 48.1$ &$ 41.8$ &$ 53.2$ &$ 54.0$ &$ 2.2$ &$ 3.6$ &$ 14.0$   \\
IDEC${}^p$ & $ 74.9$ &$ 80.1$ &$ 83.4$ &$ 42.8$ &$ 59.8$ &$ 55.4$ & $\mathbf{ 70.0}$ & $\mathbf{ 78.0}$ & $\mathbf{ 76.1}$ &$ 4.2$ &$ 7.4$ &$ 20.2$   \\ \midrule

DKM${}^r$ & $ 72.5$ &$ 77.3$ &$ 81.2$ &$ 41.8$ &$ 56.4$ &$ 54.6$ &$ 58.3$ &$ 67.0$ &$ 68.6$ & $\mathbf{ 5.8}^*$ & $\mathbf{ 9.9}^*$ &$ 21.3$   \\
DKM${}^p$ & $ 74.0$ &$ 78.3$ &$ 82.7$ &$ 36.2$ &$ 52.0$ &$ 47.0$ &$ 60.4$ &$ 71.8$ &$ 68.9$ & $\mathbf{ 5.9}$ & $\mathbf{ 10.0}$ &$ 19.7$   \\ \midrule

ClusterGAN${}^r$ & $ 63.6$ &$ 71.8$ &$ 76.8$ & $\mathbf{ 46.5}$ & $\mathbf{ 60.7}$ & $\mathbf{ 59.0}$ &$ 57.4$ &$ 67.9$ &$ 70.0$ &$ 3.2$ &$ 7.6$ &$ 20.4$   \\

VIB-GMM${}^r$ & $ 73.3^*$ &$ 78.3^*$ &$ 81.5^*$ &$ 43.7^*$ &$ 58.4^*$ & $\mathbf{ 59.3}^*$ &$ 59.9$ &$ 67.7$ &$ 68.4$ & $\mathbf{ 6.0}$ & $\mathbf{ 10.1}$ & $\mathbf{ 24.0}^*$   \\ \midrule

AE-CM${}^r$ &  $\mathbf{ 77.9}$ &$ 80.9$ & $\mathbf{ 86.1}$ &$ 43.7$ &$ 55.6$ & $\mathbf{ 59.2}$ &$ 55.1$ &$ 63.4$ &$ 65.8$ &$ 4.1$ &$ 7.5$ &$ 20.4$   \\
AE-CM${}^p$ &  $\mathbf{ 79.4}$ & $\mathbf{ 82.4}$ & $\mathbf{ 86.5}$ &$ 43.1$ &$ 56.3$ & $\mathbf{ 58.5}$ & $\mathbf{ 69.7}^*$ &$ 76.7^*$ & $\mathbf{ 76.8}^*$ &$ 4.1$ &$ 7.6$ &$ 20.2$   \\

            
            \bottomrule
            \multirow{2}{*}{Model} & \multicolumn{3}{c}{R10K} & \multicolumn{3}{c}{20News} & \multicolumn{3}{c}{10x73k} & \multicolumn{3}{c}{Pendigit} \\
            & ARI & NMI & ACC & ARI & NMI & ACC & ARI & NMI & ACC & ARI & NMI & ACC \\
            \midrule

AE+KM & $ 61.0^*$ &$ 56.8$ &$ 74.5^*$ &$ 11.3$ &$ 27.4$ &$ 24.8$ &$ 54.3$ &$ 72.5$ &$ 64.4$ &$ 55.2$ &$ 68.2$ &$ 70.2$   \\ \midrule

DCN${}^r$ & $ 18.0$ &$ 19.3$ &$ 49.5$ &$ 0.0$ &$ 0.2$ &$ 5.6$ &$ 5.3$ &$ 17.2$ &$ 23.9$ &$ 0.1$ &$ 0.8$ &$ 10.8$   \\
DCN${}^p$ &  $\mathbf{ 65.4}$ & $\mathbf{ 61.1}$ & $\mathbf{ 76.6}$ &$ 11.7$ &$ 33.5$ &$ 25.3$ &$ 9.6$ &$ 13.8$ &$ 25.2$ &$ 56.8$ &$ 72.0$ &$ 70.8$   \\ \midrule

DEC${}^r$ & $ 12.2$ &$ 13.2$ &$ 43.8$ &$ 3.2$ &$ 7.8$ &$ 10.0$ &$ 31.4$ &$ 43.5$ &$ 46.3$ &$ 36.8$ &$ 52.3$ &$ 49.3$   \\
DEC${}^p$ & $ 56.8$ &$ 56.0$ &$ 72.8$ &$ 5.5$ &$ 11.3$ &$ 11.8$ &$ 53.5$ &$ 67.1$ &$ 62.1$ &$ 59.6$ &$ 72.8$ &$ 72.2$   \\ \midrule

IDEC${}^r$ & $ 8.6$ &$ 9.5$ &$ 44.1$ &$ 0.0$ &$ 0.1$ &$ 5.5$ &$ 33.7$ &$ 46.5$ &$ 44.4$ &$ 43.3$ &$ 61.2$ &$ 53.9$   \\
IDEC${}^p$ & $ 59.7$ &$ 56.3$ &$ 73.9$ &$ 5.9$ &$ 12.6$ &$ 12.0$ &$ 60.1$ &$ 75.9$ &$ 66.5$ &$ 57.9$ &$ 71.6$ &$ 71.0$   \\ \midrule

DKM${}^r$ & $ 51.3$ &$ 49.5$ &$ 72.3$ &$ 4.7$ &$ 14.1$ &$ 10.9$ &$ 65.5$ &$ 71.3$ &$ 77.0$ &$ 52.4$ &$ 65.6$ &$ 66.9$   \\
DKM${}^p$ & $ 57.7$ &$ 55.5$ & $\mathbf{ 76.5}$ &$ 20.9$ &$ 39.2$ &$ 34.3$ &$ 38.1$ &$ 55.4$ &$ 51.6$ &$ 15.4$ &$ 27.4$ &$ 25.1$   \\ \midrule

ClusterGAN${}^r$ & $ 33.7$ &$ 35.5$ &$ 61.4$ &$ 18.6$ &$ 34.1$ &$ 34.1$ &$ 39.5$ &$ 52.1$ &$ 55.5$ & $\mathbf{ 62.9}$ &$ 74.2$ & $\mathbf{ 75.6}$   \\

VIB-GMM${}^r$ & $ 27.8$ &$ 28.7$ &$ 56.6$ &$ 0.0$ &$ 0.0$ &$ 0.0$ &$ 51.5$ &$ 60.7$ &$ 60.0$ & $\mathbf{ 64.5}$ & $\mathbf{ 75.7}$ & $\mathbf{ 74.2}$   \\ \midrule

AE-CM${}^r$ & $ 42.9$ &$ 45.6$ &$ 67.7$ & $\mathbf{ 31.5}^*$ & $\mathbf{ 45.3}^*$ & $\mathbf{ 43.3}^*$ &$ 73.1$ &$ 79.0$ &$ 80.4$ & $\mathbf{ 64.6}^*$ & $\mathbf{ 75.0}^*$ & $\mathbf{ 75.7}^*$   \\
AE-CM${}^p$ & $ 64.1$ &$ 60.0^*$ & $\mathbf{ 76.3}$ &$ 16.8$ &$ 29.0$ &$ 32.5$ & $\mathbf{ 82.3}^*$ & $\mathbf{ 83.7}^*$ & $\mathbf{ 89.4}^*$ & $\mathbf{ 60.1}$ & $\mathbf{ 70.4}$ & $\mathbf{ 73.0}$   \\

            \bottomrule

        \end{tabular*}
    \end{table*}

    In their original papers, the DEC, IDEC and DCN are pre-trained (${}^p$).
    We report here slightly lower scores that we {ascribe} to our implementation and slightly different architectures.
    Nonetheless, the take-home message here is the consistency of their poor results when randomly initialized. {This} reflects an inability to produce cluster from scratch, regardless of implementation.
    Note that, in that case, even $k$-means outperforms all of them on all the datasets (see Table~\ref{tab:cm-exp-results}). 
    On the other hand, DKM does return competitive score for both initialization scheme.
    Such results yield the question of how much clustering {that is actually} performed by DEC, IDEC and DCN and how much {that} is due to the pre-training phase.
    
    In practice, DKM  has proven sensitive to the choice of its $\lambda$ hyper-parameter and to the duration of the optimization.
    For example, we could not find a value able to cluster Pendigit.
    We conjecture that expanding the clustering term of DKM's loss, as we did between Eq.~\eqref{eq:Q} and \eqref{eq:Q1}, would improve the robustness of the model.
    
    
    On six of the datasets, at least one of the variants of AE-CM reports the highest average or highest best run.
    Especially, AE-CM${}^r$ produces competitive clusterings on all datasets except CIFAR10 despite its random initialization. 
    This setback is expected as clustering models are known to fail to cluster color images from the raw pixels~\cite{jiang2016variational,hu2017learning}.

    Also {our} AE-CM with random initialization always surpasses AE+KM except on R10K and CIFAR10, and even outperforms all the competitors by at least $20$ ARI points on 20News.
    On the down side, AE-CM${}^r$ is associated with large standard deviations which implies a less predictable optimization (see Table~\ref{apx-tab:cnet-exp-results-full} of Appendix~\ref{sec:cnet-exp-results-full}).
    Therefore, we investigate an internal measure to select the best run. 
    
    \subsubsection{AE-CM: Runtimes}
    
    We compare here the runtimes of the different method on MNIST. For a fair comparison, all the methods use the same batch size of 256 instances. We report on Table~\ref{tab:cnet-exp-runtime} average over 10 runs. We do not used any early-stopping criterion. 
    
    \begin{table*}[h!]
        \caption
        {Average runtime of each model to cluster MNIST in 150 epochs.}\label{tab:cnet-exp-runtime}
        \scriptsize
        \renewcommand{\arraystretch}{1.2}
        \centering
        
        \noindent
        \begin{tabular*}{.99\linewidth}{@{\extracolsep{\fill}}ccccccccc}
            \toprule
            Model & DEC${}^r$ & IDEC${}^r$ & DCN${}^r$ & DKM${}^r$ & ClusterGAN${}^r$ & VIB-GMM${}^r$ & AE-CM${}^r$ \\ \midrule
            Runtime & 25m13s & 27m51s & 39m18s & 38m29s & 3h17m30s & 29m26 & 19m28 \\
            \bottomrule
        \end{tabular*}
    \end{table*}
    
    Note that our implementation of DEC, IDEC and DCN are based on that of AE-CM. Hence these are the most comparable. The advantage goes to the model joint optimization, AE-CM, which is more that 5 minutes faster. 
    ClusterGAN is the slowest method. It also has the most complex architecture.

    \subsubsection{AE-CM: Robustness to the number of clusters}\label{sec:cnet-k}
    
\begin{figure*}[!h]
        \centering
            \includegraphics[width=1\linewidth]{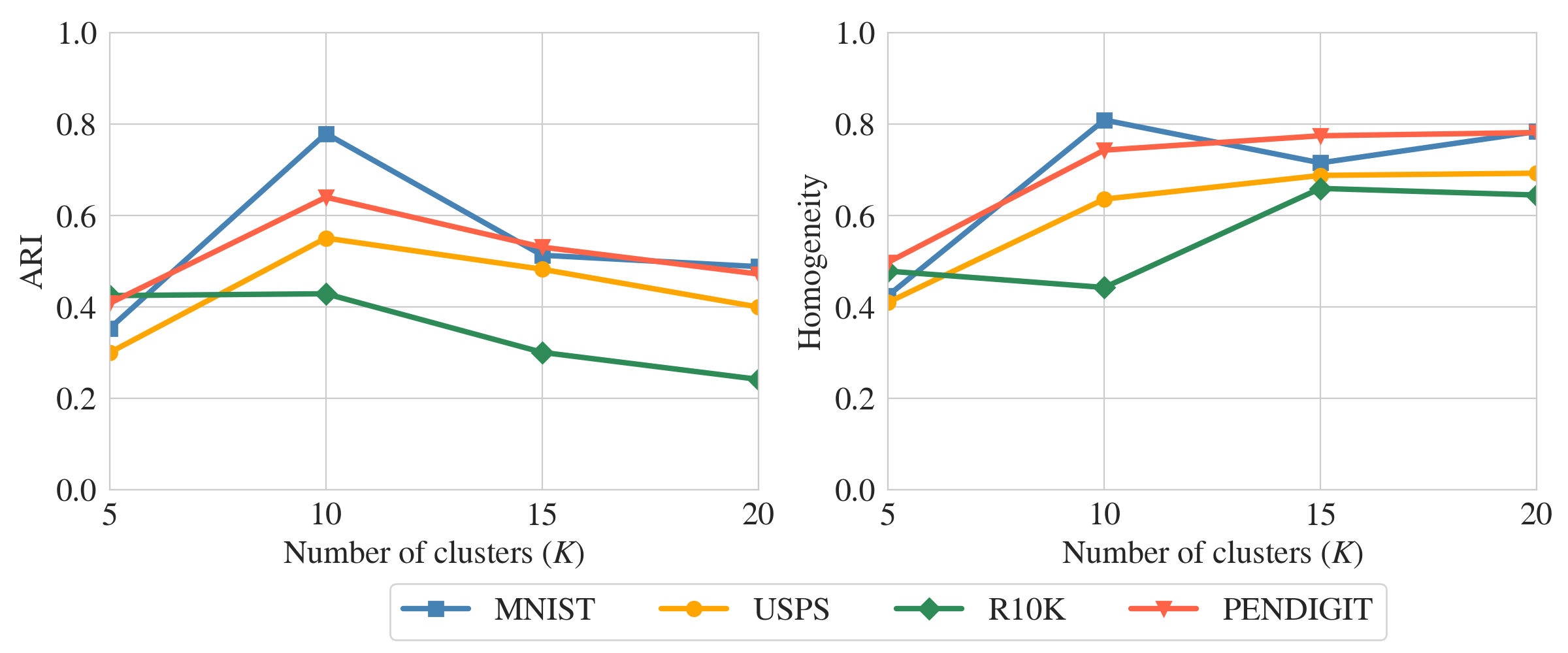}
        \caption{Adjusted Rand index and homogeneity score vs number of clusters. The correct number of clusters being $K=10$.}
        \label{fig:cnet-k}
    \end{figure*}

    In the previous experiments, we provided the true number of clusters to all algorithms for all datasets. In this experiment, we investigate the behavior of the AE-CM when it is set with a different number of clusters on four datasets: MNIST, USPS, R10K and Pendigit.
    Figure~\ref{fig:cnet-k} shows the evolution of the ARI (left) and the homogeneity~\cite{rosenberg2007v} score (right). The latter measures the purity in terms of true labels of each cluster.
    The number of cluster varies from 5 to 20.
    The correct value for all datasets is 10.
    
    The ARI curves (left plot) reach their maximum at 10 and then decrease.
    This behavior is expected since this metric (as well as NMI and ACC) penalizes the number of clusters.
    On the other hand, the homogeneity curves (right plot) increase with $K$ and stabilize for $K$ larger than 10. 
    The convergence of these curves indicates that the clustering performance of the AE-CM do not degrade if $K$ is set larger than the ground truth. 
    Such a results suggests that, when $K$ is larger than the ground truth, the AE-CM finds solutions that are partitions of those found with smaller $K$. 
    Such a phenomenon is illustrated in Appendix~\ref{sec:cm-exp-e3}.

    \subsubsection{AE-CM: Model selection}\label{sec:cnet-exp-loss-perf}

    Similarly to the CM, we discuss here a model selection heuristic for the AE-CM.
    The rationale behind the use of a DAE is to have an encoding facilitating the objective of the clustering module.
    Hence, we propose to use the heuristic of the CM (Section~\ref{sec:cm-exp-loss-perf}).
    In Table~\ref{tab:cnet-loss-corr-best}, we report the ARI of the runs with the lowest $\cL_{\text{sp}}$ for each dataset. For comparison, we also report the average and best ARI. Selected scores greater than the average are marked in boldface.
    
    \begin{table*}[!h]
        \caption{Adjusted Rand index of the run with lowest $\cL_{\text{sp}}$, the average run and the best run. Scores larger than the average are marked in boldface.}
        \label{tab:cnet-loss-corr-best}
        \centering
        \renewcommand{\arraystretch}{1.2}
        \noindent
        \begin{tabular*}{.99\linewidth}{@{\extracolsep{\fill}}p{0.14\textwidth}p{0.025\textwidth}p{0.025\textwidth}p{0.04\textwidth}p{0.025\textwidth}p{0.025\textwidth}p{0.04\textwidth}p{0.025\textwidth}p{0.025\textwidth}p{0.04\textwidth}p{0.025\textwidth}p{0.025\textwidth}c}
            \toprule
            & \multicolumn{3}{c}{MNIST} & \multicolumn{3}{c}{fMNIST} & \multicolumn{3}{c}{USPS} & \multicolumn{3}{c}{CIFAR10} \\
            Criterion & $\cL_{{\text{sp}}}$ & Avg. & Best & $\cL_{{\text{sp}}}$ & Avg. & Best & $\cL_{{\text{sp}}}$ & Avg. & Best & $\cL_{{\text{sp}}}$ & Avg. & Best \\
            \midrule
            AE-CM${}^r$  & $\mathbf{88.6}$&$77.9$&$88.6$ & $\mathbf{47.8}$&$43.7$&$48.9$ & $\mathbf{55.4}$&$55.1$&$60.6$ & $\mathbf{5.29}$&$4.15$&$5.29$ \\
            
            AE-CM${}^p$  & $\mathbf{80.3}$&$79.4$&$80.3$ & $37.3$&$43.1$&$48.4$ & $61.0$&$69.7$&$80.3$ & $2.97$&$4.13$&$5.56$ \\
            \midrule
            & \multicolumn{3}{c}{R10K} & \multicolumn{3}{c}{20News} & \multicolumn{3}{c}{10x73k} & \multicolumn{3}{c}{Pendigit} \\
            Criterion & $\cL_{{\text{sp}}}$ & Avg. & Best & $\cL_{{\text{sp}}}$ & Avg. & Best & $\cL_{{\text{sp}}}$ & Avg. & Best & $\cL_{{\text{sp}}}$ & Avg. & Best \\
            \midrule
            AE-CM${}^r$  & $36.7$&$42.9$&$62.7$ & $\mathbf{36.7}$&$31.5$&$38.7$ & $72.8$&$73.1$&$85.6$ & $\mathbf{64.6}$&$64.0$&$69.5$ \\
            
            AE-CM${}^p$  & $\mathbf{64.8}$&$64.1$&$66.7$ & $14.6$&$16.8$&$21.2$ & $79.9$&$82.3$&$86.9$ & $\mathbf{66.3}$&$65.5$&$70.5$ \\
            \bottomrule
        \end{tabular*}
    \end{table*}

    Again, scores associated to the lowest $\cL_{\text{sp}}$ are better than the average more than half of the time.
    The criterion detects the best runs of AE-CM${}^r$ on MNIST and CIFAR10 and of AE-CM${}^p$ on MNIST.
    The average absolute difference to the highest score is $6.5$ and $6.6$ ARI points for AE-CM${}^r$ and AE-CM${}^p$, respectively. 
    In summary, although $\cL_{\text{sp}}$ as a criterion does not necessarily select the best run, it filters out the worst runs.

    \subsubsection{AE-CM: Embeddings for $k$-means}\label{sec:cnet-exp-emb}

    All baselines, including our model, are non-linear extensions of $k$-means and aim to improve the AE+KM.
    We audit the methods by running $k$-means on the embeddings produced by the $20$ runs with random initialization on the MNIST dataset computed for Table~\ref{tab:cnet-exp-results} and report the average ARI, NMI and ACC in Table~\ref{tab:cnet-emb}. 
    
    \begin{table*}[!h]
        \caption
        {Average clustering performance of $k$-means on different embeddings. 
        }\label{tab:cnet-emb}
        \centering
        \scriptsize
        \renewcommand{\arraystretch}{1.2}
        \begin{tabular*}{.99\linewidth}{@{\extracolsep{\fill}}cccccccc} \toprule
            & KM & AE+KM & DCN+KM & DEC+KM & IDEC+KM & DKM+KM & AE-CM+KM \\ \midrule
            ARI & $37.8$ & $65.6$ & $64.5$ & $11.1$ & $27.5$ & $\mathbf{74.5}$ & $\mathbf{75.2}$ \\
            NMI & $49.9$ & $71.5$ & $71.1$ & $19.0$ & $38.9$ & $\mathbf{78.8}$ & $\mathbf{78.5}$ \\
            ACC & $54.5$ & $78.6$ & $76.2$ & $28.9$ & $42.5$ & $\mathbf{83.0}$ & $\mathbf{84.8}$ \\
            \bottomrule
            
        \end{tabular*}
    \end{table*}
    
    First, $KM$ reports the worse results.
    This means that applying $k$-means on a feature space learned by an autoencoder does improve the quality of the clustering.
    Next, the results clearly show the superiority of methods utilizing a joint optimization, i.e., DKM${}^r$+KM and our AE-CM${}^r$+KM.
    Interestingly, the scores of DCN${}^r$+KM are here better than those of DCN${}^r$.
    This discrepancy is certainly due the moving average used to update the centroids.
    
    \begin{figure}[!b]
        \centering
        \includegraphics[width=.99\textwidth]{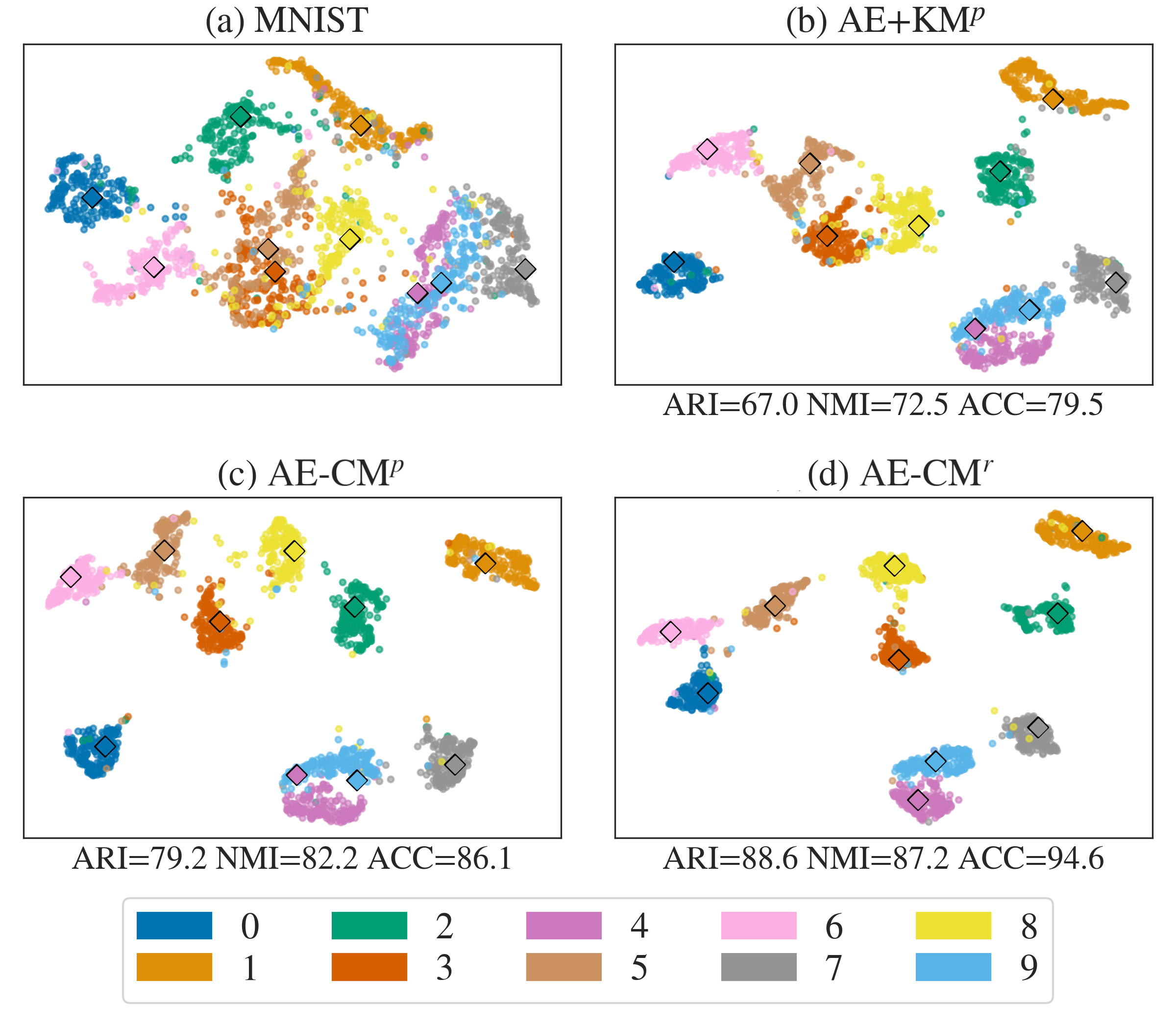}
        
        \caption
        [UMAP representation of a subset of MNIST and embeddings thereof learned by AE, IDEC and AE-CM.]
        {UMAP representation of a subset of MNIST and embeddings thereof learned by AE, IDEC and AE-CM. The squares indicate the centroids.  \looseness=-1 }
        \label{fig:cnet-umap}
    \end{figure}
    
    We continue the analysis of the embeddings with UMAP~\cite{UMAP} projections. Figure~\ref{fig:cnet-umap} depicts the projections of different embeddings of the same $2,000$ data-points.
    Figure~\ref{fig:cnet-umap}a represents the UMAP projection from the input space. 
    For (\ref{fig:cnet-umap}b), we used the best run of the AE+KM.
    For consistency, we used that embedding for the AE-CM${}^p$. 
    Hence, Figure~\ref{fig:cnet-umap} (c) does not show the best run the AE-CM${}^p$.
    Finally (\ref{fig:cnet-umap}d) is based on the best run of the AE-CM${}^r$.

    The projection of MNIST from the input space (\ref{fig:cnet-umap}a) has two pairs of classes entangled: $(3,5)$ and $(4,9)$.
    The end-to-end training of the DAE (\ref{fig:cnet-umap}b) successfully isolates each class except for cluster $4$ (dark pink) and cluster $9$ (light blue) which stay grouped together, although separable. 
    The AE-CM${}^p$ (\ref{fig:cnet-umap}c) further contracts the cluster around the centroids found by the AE+KM, but fails to separate $4$ and $9$.
    Remark that even the best run of the AE-CM${}^p$ does not to correctly split the data points.
    The centroids for $4$ and $9$ in Figures~\ref{fig:cnet-umap}b and c are in comparable positions: they align along the gap separating the true clusters. 
    This suggests that the optimization of the AE-CM${}^p$ did not move them much.
    This remark applies to the pre-trained baselines, as well.
    Lastly, the AE-CM${}^r$ successfully produces homogeneous groups (\ref{fig:cnet-umap}f). 
    Remark that the original entanglements of the pairs $(3,5)$ and $(4,9)$ are suggested by the trails between the corresponding clusters.
    
    The previous observations summarize into two insights on the behavior of the AE-CM.
    If the AE-CM starts with an embedding associated to a low reconstruction loss for the DAE, the optimization contracts the clusters which yields higher ARI scores.
    However, it is unable to move the centroids to reach another local optimum. 
    Although the AE-CM${}^r$ separates $(4,9)$, it also produces clusters more spread than those of the AE-CM${}^p$.
    The improved performances of the latter over AE+KM indicates that the AE-CM${}^r$ would benefit from tighter groups. 
    
    \subsubsection{AE-CM: Sample generation and interpolation}\label{sec:cnet-exp-gen}
    
\begin{figure}[!h]
        \centering
        \includegraphics[width=.9\linewidth]{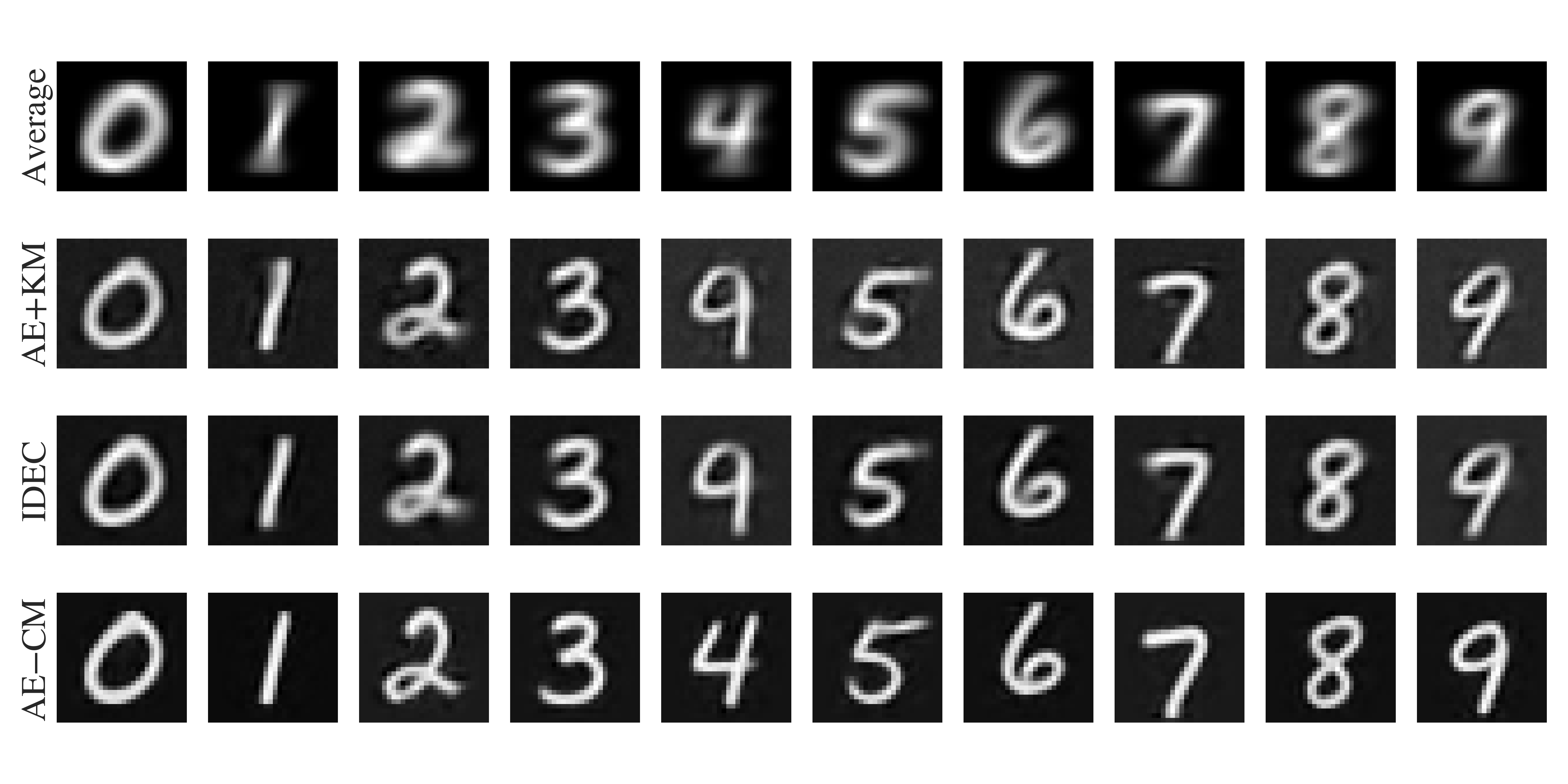}
        \caption
        {Centroids mapped back to image space for AE+KM, IDEC${}^p$, DKM${}^r$, and AE-CM${}^r$. The first row displays the average image of each class.}
        \label{fig:cnet-gen-mu}
    \end{figure}

    Thanks to the reversible feature maps obtained by the DAE, both the AE-CM and its baselines (except DEC) are generative models.
    Figure~\ref{fig:cnet-gen-mu} shows the decoding of the centroids of the best run of  AE+KM (ARI=$67.7$), IDEC${}^p$ (ARI=$77.2$), DKM${}^r$  (ARI=$83.6$) and AE-CM${}^r$  (ARI=$88.6$). 
    AE+KM's and IDEC${}^p$'s centroids for the $4$ and $9$ both look like $9$’s. 
    With an ARI and an ACC larger than $80$ and $90$, respectively, DKM${}^r$ and AE-CM${}^r$ both clustered the data correctly and found correct centroids for each class. 
    Both models produce clear images for each class, which align reasonably well with the washed-out average image of the respective classes (first row). 
    
    \begin{figure}[!h]
        \centering
        \includegraphics[width=\linewidth]{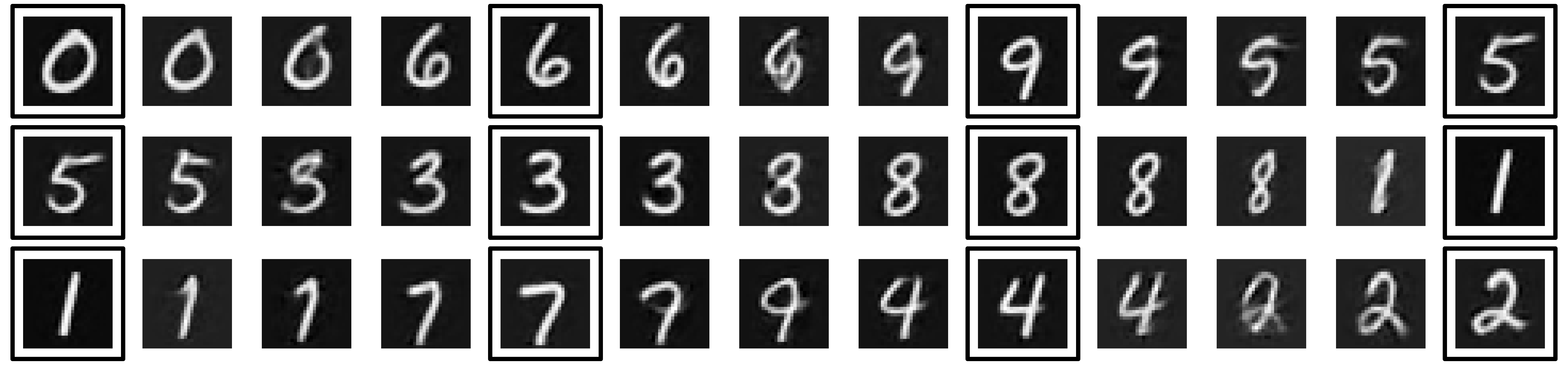}
        \caption
        {Linear interpolations between different centroids (plots with border) produced with the AE-CM.\looseness=-1}
        \label{fig:cnet-gen-inter}
    \end{figure} 
    

    Being a generative model, the AE-CM can also be used to interpolate between classes.
    Figure~\ref{fig:cnet-gen-inter} shows a path made of nine interpolations between the ten centroids of the AE-CM${}^r$.
    We observe smooth transitions between all the pairs, which indicate that the model learned a smooth manifold from noise (random initialization).

    \section{Conclusion}\label{sec:conclusion}
    
    We presented a novel clustering {algorithm} that is jointly optimized with the embedding of an autoencoder to allow for nonlinear and interpretable clusterings. 
    We first as a key result showed that the objective function of an isotropic GMM can be turned into a loss function for autoencoders. The clustering module (CM), defined as the smallest resulting network, was shown to perform similarly to its underlying GMM in extensive empirical evaluations. 
    
    Importantly, we showed how the clustering module can be straightforwardly incorporated into deep autoencoders to allow for nonlinear clusterings. The resulting clustering network, the AE-CM, empirically outperformed existing centroid-based deep clustering architectures and performed on par with representative contemporary state-of-the-art deep clustering strategies.
    Nevertheless, the AE-CM, and to a lesser extent of the clustering module itself, presented a greater volatility when trained from a randmly initialized network. 
    We expect that we could improve on that point by involving an annealing strategy on the parameter, similarly to what is done in DKM and VIB-GMM.
    
    A future line of work consist of extending the panel of deep architectures into which the clustering module can be nested. In order to improve performance on image data sets, especially, it is necessary to involve convolution. 
    However, standard image-specific architectures are not structured as autoencoder. This raises the question of the robustness of our model with respect to the symmetry of the DAE, especially for applications where the computation of class representative is not a must.
    
    From a theoretical point of view, we believe that the derivations that led to the neural interpretation of Gaussian mixture models could benefit other mixture models such as the von Mises-Fisher mixture models~\cite{hasnat2017mises} or hidden Markov models (HMM). The case of Gaussian-HMM seems especially promising as it allows to bridge with Recurrent networks~\cite{salaun2019comparing}.

    \bibliographystyle{spmpsci}      
    \bibliography{biblio}
    
    \appendix
    
    \section{Additional experiments}
    
    \subsection{UCI datasets}
    
    In this section, we report clustering performance of iGMM, DEC, IDEC, DCN and DKM as well as our models CM and AE-CM on a selection of UCI datasets.
    We consider here only random initialization. Deep clustering models share the same architecture: $d-2\times k-d$, where $d$ is the dimension of the input and $k$ the number of clusters.
    The $k$-means clusterings of DEC and IDEC are updated every 5 epochs and $\gamma=0.1$.
    The $\lambda$ hyperparameter of  DCN and DKM are set to $1$ and $0.001$, respectively.
    For CM and AE-CM the three hyper-parameters $(\alpha,\beta,\lambda)$ were leaned using Bayesian optimization.
    We report average ARI, NMI and ACC over 10 runs.

    \begin{table*}[h!]
        \caption[Comparison of clustering performance in terms of mean ARI.]
        {The clustering performance ($\times 100$) of different models on the selected datasets.  
        }\label{tab:cm-exp-results}
        \scriptsize
        \renewcommand{\arraystretch}{1.2}
        \centering
        
        \noindent
        \begin{tabular*}{.99\linewidth}{@{\extracolsep{\fill}}p{0.13\textwidth}p{0.025\textwidth}p{0.025\textwidth}p{0.04\textwidth}p{0.025\textwidth}p{0.025\textwidth}p{0.04\textwidth}p{0.025\textwidth}p{0.025\textwidth}c}
            \toprule
            \multirow{2}{*}{Model} & \multicolumn{3}{c}{Breast} & \multicolumn{3}{c}{Ecoli} & \multicolumn{3}{c}{Glass} \\
            & ARI & NMI & ACC & ARI & NMI & ACC & ARI & NMI & ACC \\
            \midrule

iGMM${}^r$ & $ 40.0 $ & $ 51.4 $ & $ 48.3 $ & $ 50.6 $ & $ 64.2 $ & $ 64 $ & $ 21.8 $ & $ 41.7 $ & $ 44.9 $   \\

DEC${}^r$  &  $ 77.8 $ &  $ 66.7 $ &  $ 86.7 $ &  $ 48.8 $ &  $ 49.5 $ &  $ 63.6 $ &  $ 22.8 $ &  $ 31.9 $ &  $ 47.5 $ \\
IDEC${}^r$ &  $ 50.7 $ &  $ 44.6 $ &  $ 68.6 $ &  $ 35.2 $ &  $ 41.4 $ &  $ 55.2 $ &  $ 17.9 $ &  $ 30.4 $ &  $ 42.3 $ \\
DCN${}^r$  &  $ 72.7 $ &  $ 57.2 $ &  $ 79.7 $ &  $ 45.7 $ &  $ 49.9 $ &  $ 63.4 $ &  $ 20.2 $ &  $ 32.9 $ &  $ 43.1 $ \\
DKM${}^r$  &  $ 83.1 $ &  $ 73.2 $ &  $ 95.6 $ &  $ 51.2 $ &  $ 61.5 $ &  $ 64.3 $ &  $ 23.2 $ &  $ 37.9 $ &  $ 49.3 $ \\

CM${}^r$ &  $ 35.9 $ &  $ 48.4 $ &  $ 47.5 $ &  $ 66.7 $ &  $ 65.6 $ &  $ 76.0 $ &  $ 27.1 $ &  $ 43.2 $ &  $ 49.1 $ \\
AE-CM${}^r$ &  $ 72.9 $ &  $ 58.0 $ &  $ 72.8 $ &  $ 72.4 $ &  $ 68.1 $ &  $ 76.7 $ &  $ 37.8 $ &  $ 47.3 $ &  $ 59.8 $ \\
            
            \bottomrule
            \multirow{2}{*}{Model} & \multicolumn{3}{c}{Iris} & \multicolumn{3}{c}{Wine} & \multicolumn{3}{c}{Yeast}  \\
            & ARI & NMI & ACC & ARI & NMI & ACC & ARI & NMI & ACC \\
            \midrule

iGMM${}^r$ & $ 62.0 $ & $ 65.9 $ & $ 83.3 $ & $ 76.9 $ & $ 81.3 $ & $ 83.7 $ & $ 15.7 $ & $ 28.8 $ & $ 38.8 $ \\

DEC${}^r$  &  $ 37.5 $ &  $ 44.3 $ &  $ 61.6 $ &  $ 23.8 $ &  $ 28.7 $ &  $ 55.9 $ &  $ 9.1 $ &  $ 16.9 $ &  $ 31.6 $ \\
IDEC${}^r$ &  $ 26.6 $ &  $ 33.4 $ &  $ 60.0 $ &  $ 24.8 $ &  $ 28.3 $ &  $ 57.8 $ &  $ 7.6 $ &  $ 15.9 $ &  $ 31.1 $ \\
DCN${}^r$  &  $ 45.3 $ &  $ 53.2 $ &  $ 67.7 $ &  $ 41.3 $ &  $ 46.7 $ &  $ 65.7 $ &  $ 6.8 $ &  $ 15.4 $ &  $ 37.4 $ \\
DKM${}^r$  &  $ 58.4 $ &  $ 65.9 $ &  $ 77.2 $ &  $ 72.4 $ &  $ 70.6 $ &  $ 90.2 $ &  $ 17.5 $ &  $ 29.6 $ &  $ 42.4 $  \\

CM${}^r$ &  $ 59.1 $ &  $ 63.4 $ &  $ 81.5 $ &  $ 82.0 $ &  $ 82.2 $ &  $ 89.6 $ &  $ 15.5 $ &  $ 28.5 $ &  $ 38.0 $ \\
AE-CM${}^r$ &  $ 78.9 $ &  $ 85.0 $ &  $ 94.4 $ &  $ 86.0 $ &  $ 85.4 $ &  $ 93.0 $ &  $ 17.7 $ &  $ 27.1 $ &  $ 46.6 $ \\
            \bottomrule
        \end{tabular*}
    \end{table*}

    \subsection{Scikit-learn benchmark}
    
\begin{figure*}[!h]
        \centering
            \includegraphics[width=1\linewidth]{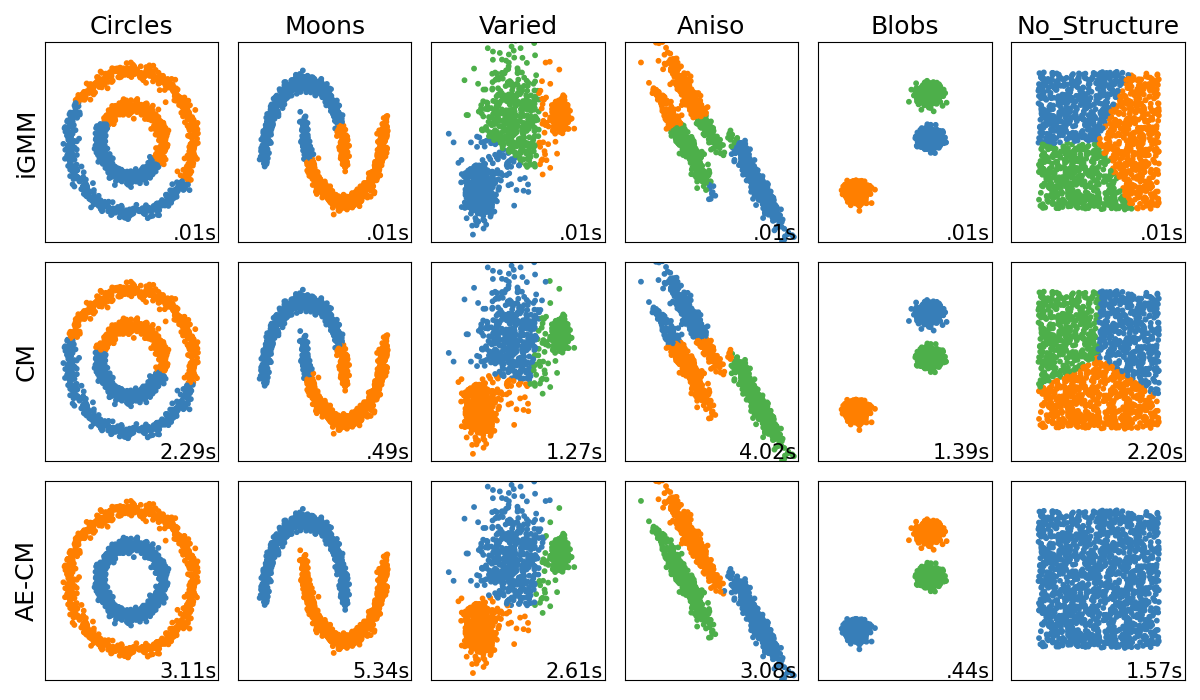}
        \caption{Clustering of six toy datasets by iGMM, CM and AE-CM.}
        \label{apx-fig:sklearn}
    \end{figure*}

    Figure~\ref{apx-fig:sklearn} reports clusterings of the scikit-learn toy datasets\footnote{\texttt{https://scikit-learn.org/stable/auto\_examples/cluster/plot\_cluster\_comparison.html}}.
    We compare here, iGMM, CM and AE-CM.
    The CM is optimized using SGD while the AE-CM relies on Adam.
    Training are stopped if the difference between the clusterings of two successive iterations are difference by less that $0.1\%$. 
    The models are run 10 times for up to 100 epochs with a fixed batch size of 20 instances.
    The average run time is shown in the lower right corner of each plot (Intel(R) Xeon(R) CPU E5-2698 v4 @ 2.20GHz).
    The parameters for each dataset are given in Table~\ref{apx-tab:sklearn}.
    
    \begin{table*}[!h]
    \caption{Hyper-parameters used for clustering the toy datasets.}
    \label{apx-tab:sklearn}
    \small
    \renewcommand{\arraystretch}{1.2}
    \centering
    
    \noindent
    \begin{tabular*}{.98\linewidth}{@{\extracolsep{\fill}}ccccc}
        \toprule
        Dataset & $\alpha$  & $\beta$   & $\lambda$ & Encoder Archi. \\ \midrule
        Moons   & 11        & 100       & 0.001     & 2-20-20-20-1 \\
        Circles & 11        & 0.001     & 0.001     & 2-quad-3 \\
        Varied  & 11        & 100       & 0.001     & 2-3 \\
        Aniso.  & 11        & 1         & 0.001     & 2-3 \\
        Blobs   & 11        & 1         & 0.001     & 2-3 \\
   No Structure & 0.1       & 1         & 0.001     & 2-3 \\
        \bottomrule
    \end{tabular*}
    \end{table*}
    
    The standard way to split the two circles and the two moons datasets is to use a polynomial kernel of degree 2 and at least 3 respectively. 
    
    To highlight the relationship between embedding and feature maps, we choose for these datasets specific architectures of the deep autoencoder of AE-CM.
    For the two moons dataset, we want the AE-CM to learn an embedding function approximating a polynomial of degree at least. Therefore, we use 3 layers with 20 units followed by a layers with a single unit. The decoder is the mirror of the encoder.
    For the two circles dataset, the encoder consists of a quadratic layer, i.e. 
    $$(x_1,x_2) \mapsto (1,x_1,x_2,x_1^2,x_1x_2,x_2^2,x_2x_1),$$ 
    followed by a dense layer with a dimension 3 output. That way, the function associated to the encoder is a feature for a polynomial kernel of degree 2. The training just has to find the proper weights of the quadratic function.
    As for the decoder, it consists of a single layer. 
    The AE-CM successfully clustered the two circles and two moons dataset, suggesting that it indeed learned embedding functions associated to polynomial kernels of degree 2 and at least 3, respectively.
    
    Finally, the last dataset consisting of a square filled with a single class, we chose an $\alpha$ lower than 1 for both CM and AE-CM. Such a setting, informs the model that the three classes will be very imbalanced. both models reacted differently. Indeed, only AE-CM was able to perfectly assign all the points to a single cluster.

    \section{Supplementary materials related to the clustering module}
    
    \subsection{CM: Hyper-parameters}\label{sec:cm-exp-hp}
    
\begin{figure*}[!h]
        \centering
        \includegraphics[width=.72\linewidth]{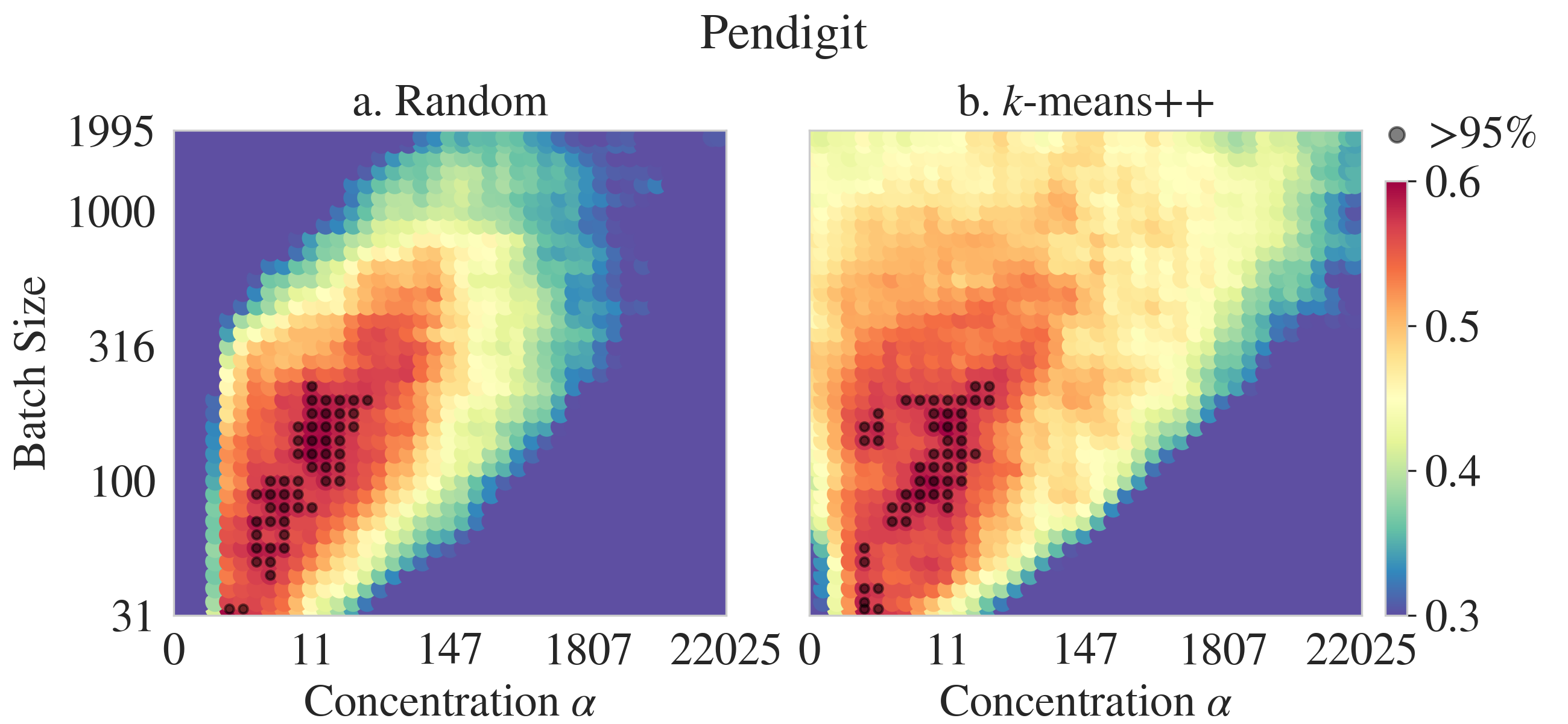}
        \caption{Clustering performance (ARI) for different combinations of batch-size, concentration and initialization scheme.     Black dots indicate average ARI greater than the ones reported in Table~\ref{tab:cm-exp-results}.}
        \label{fig:cm-hp}
    \end{figure*}

    The clustering module depends on two hyper-parameters: the size of the mini-batches and the concentration of the Dirichlet prior. 
    To visualize their influence on the clustering performance, we trained a CM on Pendigit with various sizes of batch and concentrations. 
    Figure~\ref{fig:cm-hp} shows the variation of the final ARI score for both a random (left) and a $k$-means++ initialization (right). 
    Both axes' scales are logarithmic: exponential base for the x-axis and base $10$ for the y-axis.
    Each combination is run once. 
    The ARI of each dot is the average of the nine neighboring combinations.
    Black dots indicate an average ARI greater than the ones reported in Table~\ref{tab:cm-exp-results}.

    There is a lower bound on $\alpha$ under which the optimization of a randomly initialized model underperforms or fails. 
    The $k$-means++ initialization removes this border and spreads out the well-performing area.
    The distribution in both settings means that the hyper-parameters can be tuned by fitting a bi-variate Gaussian distribution.
    
    \subsection{CM: asymmetric prior}\label{sec:cm-exp-prior}
    
    So far we considered only symmetric Dirichlet priors ($\bm \alpha = \alpha {\bm 1}_K, \: \alpha \in \R^+$) regardless of the imbalance between the labels. 
    Here, we repeat the previous experiment using the true labels distribution as the prior, i.e. $\bm \alpha = \alpha {\bm f}$ where ${\bm f} \<f_k\>_K \in \S^K $ is the frequency of each label. 
    In terms of implementation, $E_4$ is computed by sorting both $\balpha$ and $\bm \gamma_i$.
    We evaluate results on the  10x73k and Pendigit datasets, which have unbalance and balanced classes, respectively.
    We consider here only random initializations.
    Again, black dots indicate an average ARI greater than the ones reported in Table~\ref{tab:cm-exp-results}.
    
    \begin{figure*}[!h]
        \includegraphics[width=\linewidth]{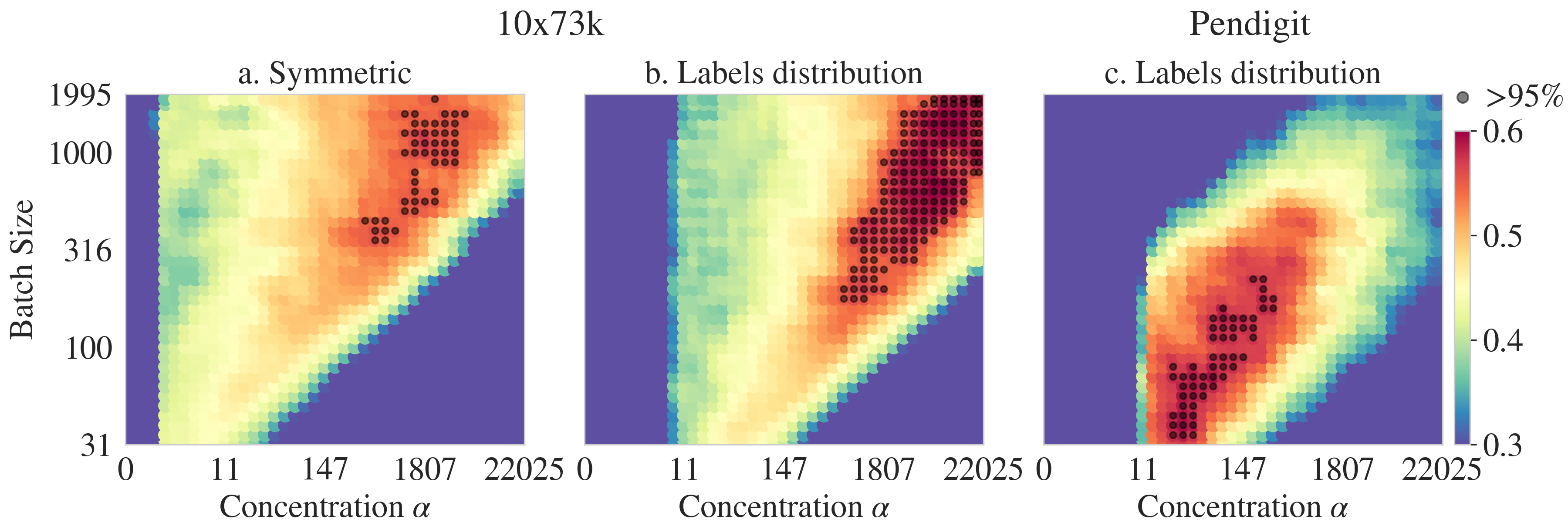}
        \caption{Clustering performance (ARI) for different combinations of batch-size, concentration and prior distribution.     Black dots indicate average ARI greater than the ones reported in Table~\ref{tab:cm-exp-results}.}
        \label{fig:cm-prior}
    \end{figure*}

    Figure~\ref{fig:cm-prior}b contains more black dots and a larger red area compared to \ref{fig:cm-prior}a. 
    The changes are greater than between Figure~\ref{fig:cm-prior}c and \ref{fig:cm-hp}b.
    This discrepancy between the datasets illustrates that unbalanced ones benefit more from a custom prior. 
    However, a higher concentration is needed to enforce the distribution: the lower bound on $\alpha$ is higher in Figure~\ref{fig:cm-prior}b and c than in \ref{fig:cm-prior}a and \ref{fig:cm-hp}a, respectively. 
    Using the true class distribution, especially if the data is unbalanced, does ease the hyper-parameter selection. 
    Nevertheless, such an information is not always known a priori.

    \subsection{CM: Merging clusters with $E_3$}\label{sec:cm-exp-e3}
    
\begin{figure*}[!b]
        \vspace{-2em}
        \includegraphics[width=\linewidth]{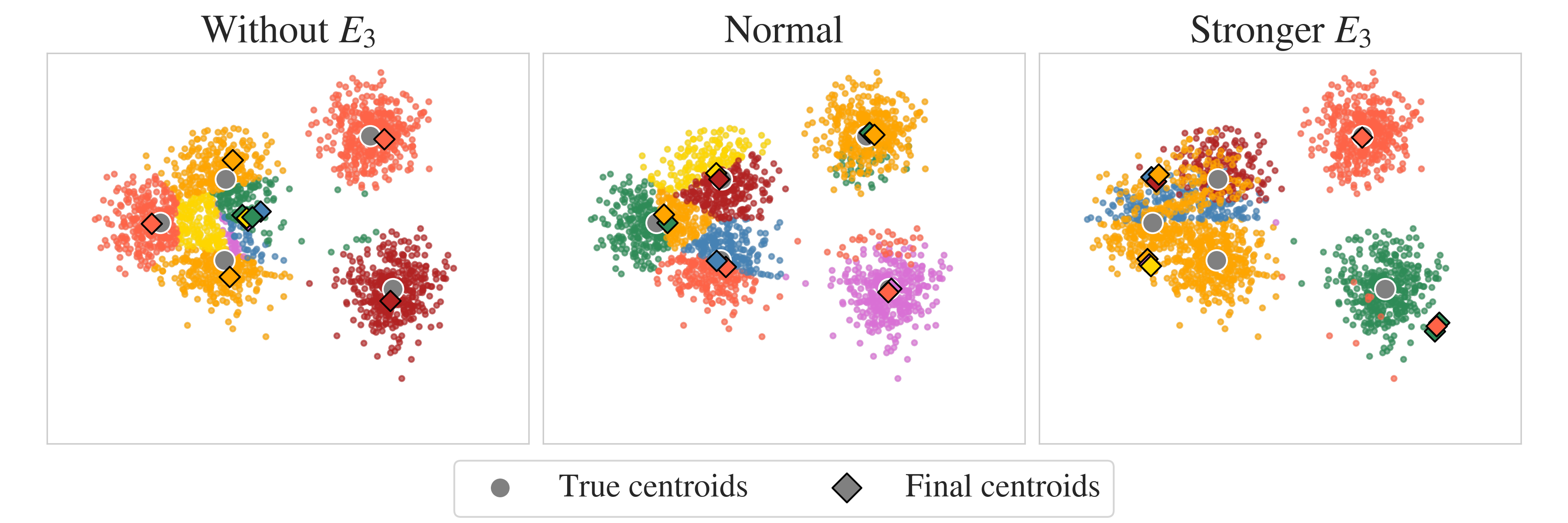}
        \caption{Final positions of the centroids depending on the importance of $E_3$ in the loss of the clustering module.}
        \label{fig:cm-e3}
    \end{figure*}

    We claimed in Section~\ref{sec:loss-terms} that $E_3$ favors the merging of clusters.
    To illustrate this phenomenon, we train CM${}^r$ on the 5 Gaussians dataset with twice the number of true clusters (i.e., $K=10$). 
    We compare three variants of CM's loss function: without $E_3$, with $E_3$ and with $E_3$ multiplied by $1.5$. 
    The final centroids and clustering are depicted in Figure~\ref{fig:cm-e3}. For legibility, overlapping centroids are slightly shifted using a Gaussian noise.
    
    A vanilla CM (Figure~\ref{fig:cm-e3}b) correctly positions five pairs of centroids on top of the true cluster centroids.
    Without $E_3$ (Fig.~\ref{fig:cm-e3}a), the model fails to merge the clusters properly. 
    While five centroids are close to each of the true cluster, the five remaining are gathered around $0$.
    Conversely, if $E_3$ is weighted stronger (Fig.~\ref{fig:cm-e3}c), the model becomes so prone to merge clusters that it partitions the left cloud using only two groups of centroids. 

    
    \begin{table*}[!t]
        \caption[Comparison of clustering performance in terms of mean ARI.]
        {The clustering results of the methods on the experimental datasets. 
        }\label{apx-tab:cm-exp-results-full}
        \tiny
        \renewcommand{\arraystretch}{1.4}
        \centering
        
        \noindent
        \begin{tabular*}{.98\linewidth}{@{\extracolsep{\fill}}p{0.08\textwidth}p{0.02\textwidth}p{0.025\textwidth}p{0.025\textwidth}p{0.035\textwidth}p{0.025\textwidth}p{0.025\textwidth}p{0.035\textwidth}p{0.025\textwidth}p{0.025\textwidth}p{0.035\textwidth}p{0.025\textwidth}p{0.025\textwidth}p{0.025\textwidth}}
            \toprule
            \multirow{2}{*}{Model} & & \multicolumn{3}{c}{MNIST} & \multicolumn{3}{c}{fMNIST} & \multicolumn{3}{c}{USPS} & \multicolumn{3}{c}{CIFAR10} \\
            & & ARI & NMI & ACC & ARI & NMI & ACC & ARI & NMI & ACC & ARI & NMI & ACC \\ \midrule
            
KM${}^r$ & avg std max & $ 37.8$  $ \pm 2.7  $ $ 43.4  $ & $\mathbf{ 49.9} $  $ \pm 1.9  $ $ 54.1  $ & $ 54.5$  $ \pm 4.3  $ $ 62.4  $ & $ 36.6$  $ \pm 1.7  $ $ 38.5  $ & $ 51.6$  $ \pm 1.2  $ $ 53.1  $ & $ 53.2$  $ \pm 3.4  $ $ 59.4  $ & $ 52.6$  $ \pm 2.3  $ $ 56.2  $ & $\mathbf{ 61.8} $  $ \pm 1.4  $ $ 63.9  $ & $ 63.0$  $ \pm 3.2  $ $ 68.3  $ & $ 4.2$  $ \pm 0.1  $ $ 4.3  $ & $ 8.1$  $ \pm 0.2  $ $ 8.3  $ & $ 20.8$  $ \pm 0.4  $ $ 21.7  $   \\
KM${}^p$ & avg std max & $ 36.9$  $ \pm 1.7  $ $ 40.8  $ & $\mathbf{ 49.2} $  $ \pm 1.3  $ $ 52.1  $ & $ 54.0$  $ \pm 3.0  $ $ 58.8  $ & $ 35.2$  $ \pm 1.6  $ $ 38.5  $ & $ 51.0$  $ \pm 1.0  $ $ 53.3  $ & $ 50.8$  $ \pm 3.4  $ $ 57.9  $ & $ 50.2$  $ \pm 4.9  $ $ 54.5  $ & $ 60.8$  $ \pm 2.5  $ $ 63.7  $ & $ 61.1$  $ \pm 5.2  $ $ 67.2  $ & $ 4.2$  $ \pm 0.1  $ $ 4.4  $ & $ 8.1$  $ \pm 0.2  $ $ 8.5  $ & $ 20.7$  $ \pm 0.3  $ $ 21.5  $   \\ \midrule

GMM${}^r$ & avg std max & $ 12.0$  $ \pm 1.6  $ $ 15.8  $ & $ 20.9$  $ \pm 2.1  $ $ 25.4  $ & $ 29.9$  $ \pm 1.7  $ $ 34.2  $ & $ 26.8$  $ \pm 2.9  $ $ 31.2  $ & $ 44.2$  $ \pm 2.3  $ $ 48.4  $ & $ 39.5$  $ \pm 2.9  $ $ 45.1  $ & $ 9.5$  $ \pm 4.0  $ $ 19.0  $ & $ 17.6$  $ \pm 3.9  $ $ 26.7  $ & $ 27.8$  $ \pm 3.1  $ $ 34.5  $ & $ 0.7$  $ \pm 0.0  $ $ 0.8  $ & $ 0.9$  $ \pm 0.0  $ $ 1.0  $ & $ 12.1$  $ \pm 0.1  $ $ 12.2  $   \\
GMM${}^p$ & avg std max & $ 22.7$  $ \pm 1.7  $ $ 25.2  $ & $ 35.6$  $ \pm 1.2  $ $ 37.7  $ & $ 42.6$  $ \pm 1.9  $ $ 45.3  $ & $ 34.6$  $ \pm 2.2  $ $ 38.3  $ & $ 52.6$  $ \pm 1.5  $ $ 54.9  $ & $ 47.2$  $ \pm 2.5  $ $ 49.8  $ & $ 35.1$  $ \pm 4.4  $ $ 39.8  $ & $ 50.9$  $ \pm 2.7  $ $ 53.7  $ & $ 52.0$  $ \pm 3.7  $ $ 56.6  $ & $ 3.4$  $ \pm 1.0  $ $ 4.5  $ & $ 6.9$  $ \pm 1.0  $ $ 8.1  $ & $ 20.2$  $ \pm 0.8  $ $ 21.1  $   \\ \midrule

iGMM${}^r$ & avg std max & $ 31.3$  $ \pm 1.2  $ $ 32.4  $ & $ 42.7$  $ \pm 0.9  $ $ 43.5  $ & $ 48.5$  $ \pm 1.6  $ $ 50.3  $ & $ 35.7$  $ \pm 1.5  $ $ 37.6  $ & $ 50.7$  $ \pm 0.8  $ $ 51.9  $ & $ 51.4$  $ \pm 2.7  $ $ 55.0  $ & $ 44.2$  $ \pm 2.6  $ $ 47.5  $ & $ 55.3$  $ \pm 2.2  $ $ 58.0  $ & $ 56.2$  $ \pm 2.8  $ $ 59.7  $ & $ 4.1$  $ \pm 0.1  $ $ 4.2  $ & $ 7.9$  $ \pm 0.2  $ $ 8.1  $ & $ 21.1$  $ \pm 0.3  $ $ 21.6  $   \\
iGMM${}^p$ & avg std max & $ 31.1$  $ \pm 1.3  $ $ 32.4  $ & $ 42.7$  $ \pm 0.7  $ $ 43.5  $ & $ 47.5$  $ \pm 2.6  $ $ 50.3  $ & $ 35.4$  $ \pm 1.8  $ $ 37.7  $ & $ 50.8$  $ \pm 1.1  $ $ 52.1  $ & $ 51.6$  $ \pm 2.7  $ $ 54.5  $ & $ 44.4$  $ \pm 1.8  $ $ 47.6  $ & $ 56.1$  $ \pm 1.3  $ $ 58.1  $ & $ 55.6$  $ \pm 2.5  $ $ 59.7  $ & $ 4.1$  $ \pm 0.1  $ $ 4.2  $ & $ 7.8$  $ \pm 0.2  $ $ 8.1  $ & $ 21.1$  $ \pm 0.3  $ $ 21.5  $   \\ \midrule

CM${}^r$ & avg std max & $\mathbf{ 39.7} $  $ \pm 1.9  $ $ 43.5  $ & $\mathbf{ 50.0} $  $ \pm 1.5  $ $ 53.0  $ & $\mathbf{ 56.8} $  $ \pm 2.1  $ $ 60.8  $ & $\mathbf{ 42.3} $  $ \pm 3.4  $ $ 44.6  $ & $\mathbf{ 54.0} $  $ \pm 2.3  $ $ 55.7  $ & $\mathbf{ 62.0} $  $ \pm 4.1  $ $ 64.4  $ & $\mathbf{ 54.3} $  $ \pm 2.1  $ $ 58.6  $ & $\mathbf{ 62.8} $  $ \pm 1.0  $ $ 65.1  $ & $\mathbf{ 67.0} $  $ \pm 2.5  $ $ 75.0  $ & $ 4.8$  $ \pm 0.2  $ $ 5.1  $ & $ 8.9$  $ \pm 0.3  $ $ 9.4  $ & $\mathbf{ 22.0} $  $ \pm 0.5  $ $ 23.2  $   \\
CM${}^p$ & avg std max & $\mathbf{ 39.1} $  $ \pm 1.4  $ $ 43.5  $ & $\mathbf{ 49.5} $  $ \pm 1.1  $ $ 52.9  $ & $\mathbf{ 55.9} $  $ \pm 1.3  $ $ 59.8  $ & $\mathbf{ 41.4} $  $ \pm 4.1  $ $ 44.7  $ & $\mathbf{ 53.4} $  $ \pm 2.8  $ $ 55.8  $ & $\mathbf{ 60.7} $  $ \pm 5.1  $ $ 64.5  $ & $\mathbf{ 53.4} $  $ \pm 4.1  $ $ 59.2  $ & $\mathbf{ 62.8} $  $ \pm 2.3  $ $ 66.9  $ & $ 63.7$  $ \pm 4.3  $ $ 69.6  $ & $\mathbf{ 4.9} $  $ \pm 0.1  $ $ 5.2  $ & $\mathbf{ 9.3} $  $ \pm 0.2  $ $ 9.7  $ & $\mathbf{ 22.3} $  $ \pm 0.5  $ $ 23.2  $   \\
            
            \bottomrule
            \multirow{2}{*}{Model} & & \multicolumn{3}{c}{R10K} & \multicolumn{3}{c}{20News} & \multicolumn{3}{c}{10x73k} & \multicolumn{3}{c}{Pendigit} \\
            & & ARI & NMI & ACC & ARI & NMI & ACC & ARI & NMI & ACC & ARI & NMI & ACC \\ \midrule

KM${}^r$ & avg std max & $\mathbf{ 33.8} $  $ \pm 14.3  $ $ 61.5  $ & $ 38.1$  $ \pm 10.2  $ $ 56.4  $ & $\mathbf{ 61.2} $  $ \pm 11.9  $ $ 80.3  $ & $ 14.8$  $ \pm 1.3  $ $ 17.4  $ & $ 32.3$  $ \pm 1.9  $ $ 36.1  $ & $ 31.0$  $ \pm 1.9  $ $ 34.5  $ & $ 36.5$  $ \pm 0.8  $ $ 37.3  $ & $ 55.4$  $ \pm 0.7  $ $ 56.1  $ & $ 55.0$  $ \pm 1.8  $ $ 56.3  $ & $\mathbf{ 56.5} $  $ \pm 3.7  $ $ 62.3  $ & $ 67.8$  $ \pm 1.9  $ $ 70.6  $ & $\mathbf{ 71.1} $  $ \pm 4.8  $ $ 76.8  $   \\
KM${}^p$ & avg std max & $ 29.5$  $ \pm 15.1  $ $ 61.5  $ & $ 36.0$  $ \pm 9.2  $ $ 56.5  $ & $ 58.3$  $ \pm 10.3  $ $ 80.3  $ & $ 14.8$  $ \pm 1.5  $ $ 17.6  $ & $ 33.5$  $ \pm 2.4  $ $ 37.9  $ & $ 32.0$  $ \pm 2.5  $ $ 36.4  $ & $ 36.7$  $ \pm 0.3  $ $ 37.3  $ & $ 55.5$  $ \pm 0.4  $ $ 56.0  $ & $ 55.3$  $ \pm 1.2  $ $ 56.3  $ & $\mathbf{ 58.1} $  $ \pm 3.1  $ $ 62.2  $ & $ 68.9$  $ \pm 1.2  $ $ 70.6  $ & $\mathbf{ 72.7} $  $ \pm 3.5  $ $ 76.7  $   \\ \midrule

GMM${}^r$ & avg std max & $ 0.1$  $ \pm 0.1  $ $ 0.3  $ & $ 0.1$  $ \pm 0.1  $ $ 0.3  $ & $ 26.4$  $ \pm 0.5  $ $ 27.1  $ & $ 0.1$  $ \pm 0.0  $ $ 0.2  $ & $ 0.7$  $ \pm 0.1  $ $ 0.8  $ & $ 7.0$  $ \pm 0.1  $ $ 7.3  $ & $ 26.3$  $ \pm 6.7  $ $ 34.2  $ & $ 49.0$  $ \pm 8.4  $ $ 60.5  $ & $ 42.5$  $ \pm 6.3  $ $ 51.4  $ & $ 54.4$  $ \pm 4.5  $ $ 63.2  $ & $ 68.1$  $ \pm 3.4  $ $ 74.9  $ & $ 66.6$  $ \pm 6.6  $ $ 79.0  $   \\
GMM${}^p$ & avg std max & $ 29.7$  $ \pm 15.0  $ $ 61.5  $ & $ 36.2$  $ \pm 9.2  $ $ 56.5  $ & $ 58.4$  $ \pm 10.2  $ $ 80.3  $ & $ 14.8$  $ \pm 1.5  $ $ 17.6  $ & $ 33.5$  $ \pm 2.4  $ $ 37.9  $ & $ 32.0$  $ \pm 2.5  $ $ 36.4  $ & $ 38.4$  $ \pm 2.3  $ $ 40.3  $ & $ 61.7$  $ \pm 1.5  $ $ 62.7  $ & $ 55.6$  $ \pm 3.3  $ $ 58.1  $ & $ 55.4$  $ \pm 3.1  $ $ 60.4  $ & $\mathbf{ 70.1} $  $ \pm 1.7  $ $ 73.8  $ & $\mathbf{ 71.5} $  $ \pm 3.0  $ $ 78.6  $   \\ \midrule

iGMM${}^r$ & avg std max & $\mathbf{ 39.8} $  $ \pm 5.2  $ $ 52.4  $ & $\mathbf{ 44.2} $  $ \pm 3.2  $ $ 51.0  $ & $\mathbf{ 66.9} $  $ \pm 3.8  $ $ 73.8  $ & $\mathbf{ 18.4} $  $ \pm 1.1  $ $ 20.5  $ & $\mathbf{ 41.8} $  $ \pm 1.8  $ $ 45.8  $ & $\mathbf{ 37.4} $  $ \pm 1.7  $ $ 41.2  $ & $ 34.2$  $ \pm 0.5  $ $ 34.8  $ & $ 53.7$  $ \pm 0.9  $ $ 55.5  $ & $ 53.9$  $ \pm 1.3  $ $ 54.8  $ & $\mathbf{ 58.1} $  $ \pm 3.8  $ $ 62.0  $ & $ 69.0$  $ \pm 1.6  $ $ 70.9  $ & $\mathbf{ 72.7} $  $ \pm 4.5  $ $ 77.3  $   \\
iGMM${}^p$ & avg std max & $ 27.3$  $ \pm 17.6  $ $ 45.5  $ & $ 32.3$  $ \pm 17.5  $ $ 49.2  $ & $ 60.4$  $ \pm 13.6  $ $ 75.6  $ & $ 13.4$  $ \pm 1.9  $ $ 16.8  $ & $ 36.8$  $ \pm 2.7  $ $ 41.8  $ & $ 31.4$  $ \pm 2.7  $ $ 38.2  $ & $ 33.1$  $ \pm 4.0  $ $ 35.2  $ & $ 53.1$  $ \pm 4.2  $ $ 57.9  $ & $ 52.1$  $ \pm 3.8  $ $ 54.7  $ & $\mathbf{ 56.9} $  $ \pm 4.3  $ $ 61.7  $ & $\mathbf{ 68.9} $  $ \pm 2.1  $ $ 70.8  $ & $\mathbf{ 72.0} $  $ \pm 4.8  $ $ 77.4  $   \\ \midrule

CM${}^r$ & avg std max & $\mathbf{ 38.5} $  $ \pm 10.6  $ $ 56.0  $ & $\mathbf{ 41.0} $  $ \pm 7.1  $ $ 55.2  $ & $\mathbf{ 64.9} $  $ \pm 8.1  $ $ 76.4  $ & $ 9.7$  $ \pm 0.7  $ $ 10.9  $ & $ 21.2$  $ \pm 0.7  $ $ 22.6  $ & $ 18.3$  $ \pm 0.8  $ $ 20.0  $ & $\mathbf{ 54.8} $  $ \pm 1.2  $ $ 57.4  $ & $\mathbf{ 63.8} $  $ \pm 0.6  $ $ 65.2  $ & $\mathbf{ 69.5} $  $ \pm 2.0  $ $ 74.3  $ & $\mathbf{ 57.3} $  $ \pm 1.7  $ $ 60.5  $ & $ 67.0$  $ \pm 1.2  $ $ 69.1  $ & $\mathbf{ 72.0} $  $ \pm 2.5  $ $ 76.4  $   \\
CM${}^p$ & avg std max & $ 32.6$  $ \pm 12.7  $ $ 62.9  $ & $ 39.2$  $ \pm 8.1  $ $ 57.8  $ & $ 60.2$  $ \pm 10.0  $ $ 80.9  $ & $ 16.3$  $ \pm 2.4  $ $ 22.0  $ & $ 28.8$  $ \pm 2.4  $ $ 34.1  $ & $ 30.8$  $ \pm 2.5  $ $ 36.1  $ & $\mathbf{ 55.4} $  $ \pm 3.0  $ $ 62.4  $ & $\mathbf{ 64.0} $  $ \pm 2.0  $ $ 67.8  $ & $\mathbf{ 71.0} $  $ \pm 3.8  $ $ 78.8  $ & $\mathbf{ 57.3} $  $ \pm 2.0  $ $ 60.1  $ & $ 66.9$  $ \pm 1.3  $ $ 68.6  $ & $\mathbf{ 72.3} $  $ \pm 2.3  $ $ 75.0  $   \\

            \bottomrule
        \end{tabular*}
    \end{table*}

    \subsection{CM: Clustering performance}\label{sec:cm-exp-results-full}
    Table~\ref{apx-tab:cm-exp-results-full} contains the full clustering results for CM, including standard deviation and the best run.
    
    \subsection{CM: Empirical setting}\label{apx-sec:cm-exp-set}
    
    For the experiments reported in Section~\ref{sec:cm-exp-set}, the clustering module is trained over $150$ epochs using the Adam optimizer (learning rate=$0.001$).
    The concentration $\alpha$ and batch-size $B$ used for each dataset are reported in Table~\ref{apx-tab:cm-exp-setting}.
    The hyper-parameters were optimized using Bayesian optimization over $2,000$ iterations.
    
    \begin{table*}[!h]
        \caption{Hyper-parameters used for the experiments in Section~\ref{sec:cm-exp-set}.}
        \label{apx-tab:cm-exp-setting}
        \small
        \renewcommand{\arraystretch}{1.2}
        \centering
        
        \noindent
        \begin{tabular*}{.98\linewidth}{@{\extracolsep{\fill}}cccccccccc}
            \toprule
            & {MNIST} & {fMNIST} & {USPS} & {CIFAR10} & {R10K} &{20News} & {10x73k} & {Pendigit} \\ \midrule
            $\alpha$ & $177$ & $80$ & $40$ & $164$ & $10$ & $11$ & $1000$ & $13$ \\
            $B$ & $111$ & $35$ & $150$ & $350$ & $400$ & $85$ & $500$ & $80$ \\
            \bottomrule
        \end{tabular*}
    \end{table*}
    
    \section{Supplementary materials for AE-CM}

    \subsection{AE-CM: Hyper-parameters}\label{sec:cnet-exp-hp}
    
    \begin{figure*}[!h]
        \centering
        \begin{tabular*}{.9\textwidth}{cc}
            (a) AE-CM${}^r$ & (b) AE-CM${}^p$ \\
            \includegraphics[width=.4\linewidth]{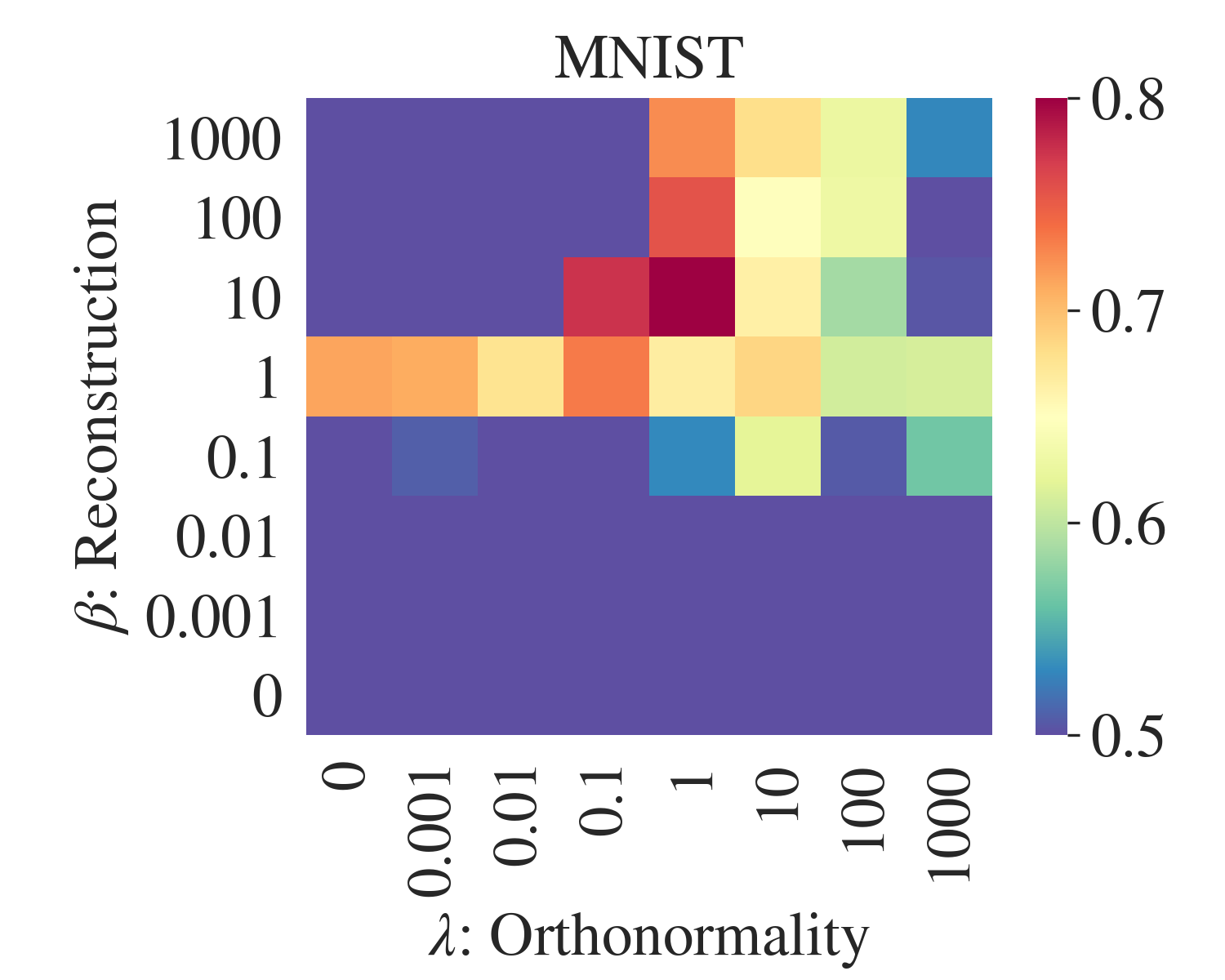} & \includegraphics[width=.4\textwidth]{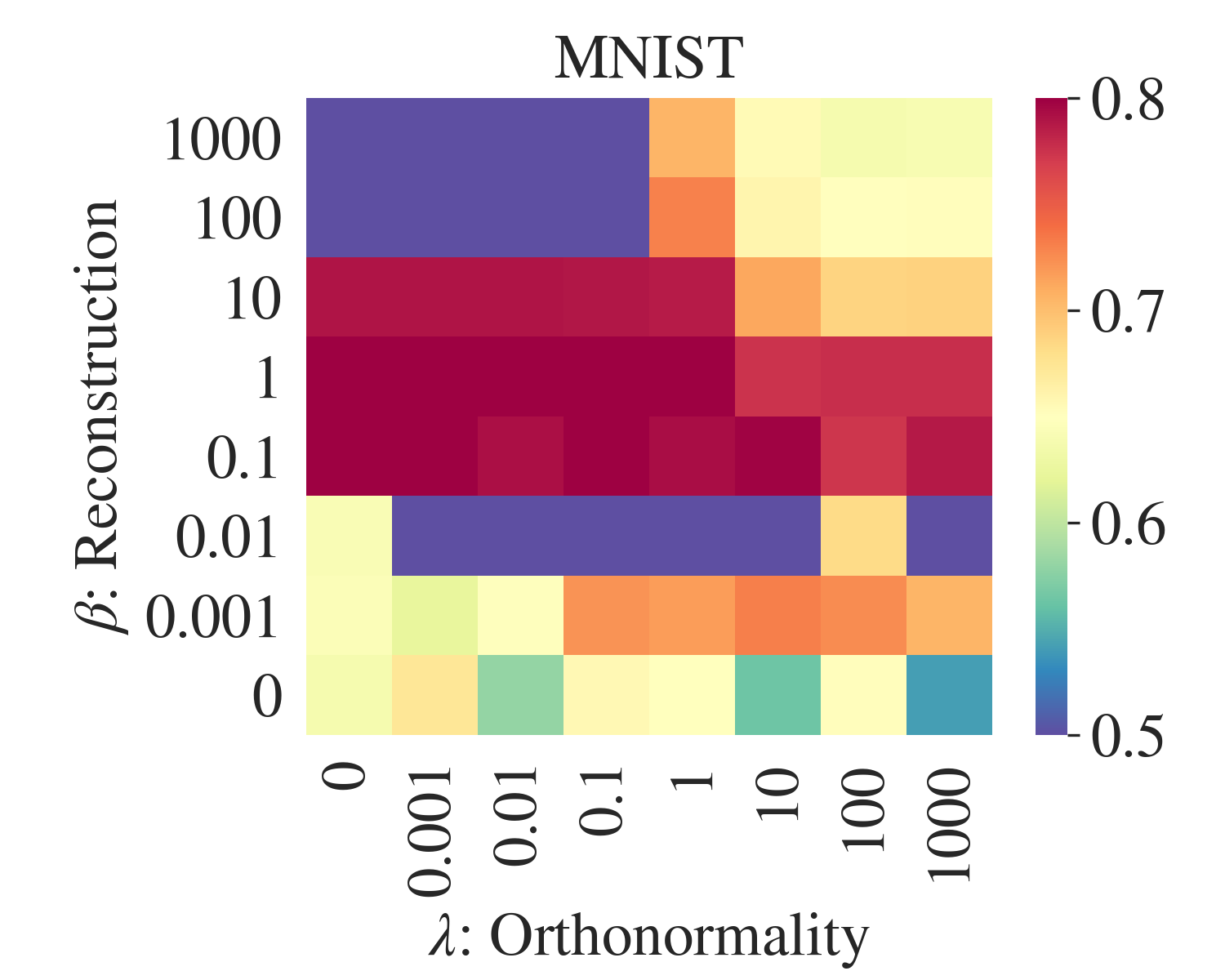} \\
        \end{tabular*}
        \caption{Clustering performance (ARI) of AE-CM on MNIST for different values of $\beta$ and $\lambda$ and initialization scheme.}
        \label{fig:cnet-hp}
    \end{figure*}
    
    Besides the architecture of the DAE, AE-CM has two hyper-parameters more than CM: $\beta$ weighting the reconstruction of the DAE and the Lagrange coefficient $\lambda$ enforcing the orthonormality of the centroids. 
    To visualize their influence on the clustering performance, a AE-CM is trained on MNIST with various values of $\beta$ and $\lambda$. Each combination is repeated five times. 
    Note that when $\beta=0$ the setting is equivalent to training only the encoder of the DAE (akin to DEC).
    Also, if $\lambda=0$ the orthogonality constraint is omitted. 
    Figure~\ref{fig:cnet-hp} represents the average ARI scores for each combination and both initialization schemes as a heat-map.

    With a random initialization (Figure~\ref{fig:cnet-hp}a), if both $\beta$ and $\lambda$ are not large enough, the clustering fails, excepted when $\beta=1$.
    In that case, the model performs well for every value of $\lambda$, even for $\lambda = 0$, i.e., without the orthonormality constraint.
    Conversely, the AE-CM${}^r$ always fails if trained without the reconstruction of the DAE, ($\beta=0$) . 
    As the order of magnitude of both parameters increases, the performances worsen.
    
    The distribution of AE-CM${}^p$ (Figure~\ref{fig:cnet-hp}b) presents similarities with the previous one.
    Overall the average performances are better for each combination. 
    When $\beta$ is very small, the ARI exceeds $0.5$.
    The performance also decrease as $\beta$ and $\lambda$ become larger.
    Most noticeable, the band around $\beta=1$ is still there, but it is thicker. 
    This is in line with the similar analysis on CM (Section~\ref{sec:cm-exp-hp}): Pre-trained models are less sensitive to hyperparameters.


    \subsection{AE-CM: Empirical setting}\label{sec:cnet-exp-set}
    
    For the experiments reported in Section~\ref{sec:cnet-exp}, AE-CM is trained over $150$ epochs using the Adam optimizer (learning rate=$0.001$).  
    Each layer of the DAE is activated with a $\rm {leaky}$-$\rm{ReLU}$ with a slope of $0.2$, except for the last one of the encoder and of the decoder. 
    AE-CM depends on four hyper-parameters: the weight $\beta>0$, the concentration $\balpha \in \S^K$, the Lagrange multiplier $\lambda>0$ and the size of the batches $B \in \N^*$.
    The four hyper-parameters plus the dimension of the feature space, $p$; were optimized using Bayesian optimization over $2,000$ iterations.
    The selected values are reported in Table~\ref{apx-tab:cnet-exp-setting}.

    \begin{table*}[!h]
        \caption{Hyper-parameters used to train AE-CM for the experiments in Section~\ref{sec:cnet-exp}.}
        \label{apx-tab:cnet-exp-setting}
        \small
        \renewcommand{\arraystretch}{1.2}
        \centering
        
        \noindent
        \begin{tabular*}{.98\linewidth}{@{\extracolsep{\fill}}cccccccccc}
            \toprule
                      & {MNIST} & {fMNIST}& {USPS}&{CIFAR10}& {R10K}  &{20News} & {10x73k}& {Pendigit} \\ \midrule
            $\alpha$  & $230$   & $13$    & $20$  & $64$    & $2$     & $10 $   & $7 $    & $13 $ \\
            $\beta$   & $5$     & $47$    & $0.5$ & $1$     & $1$     & $232$   & $15$    & $0.5$ \\
            $\lambda$ & $1$     & $1$     & $1$   & $1$     & $1$     & $1  $   & $1 $    & $1  $ \\
            $B$       & $500$   & $175$   & $256$ & $256$   & $256$   & $300$   & $7 $    & $100 $ \\
            $p$       & $10$    & $10$    & $10$  & $10$    & $100$   & $100$   &  $10$   & $10 $ \\
            \bottomrule
        \end{tabular*}
    \end{table*}
    
    The same architecture is used for the baselines, except for DKM where the activation are all $\rm{ReLU}$.
    DEC, IDEC and DCN update their clustering every $u$ iterations,
    IDEC and DCN rely on a hyper-parameter $\gamma$ and DKM on a $\lambda$.
    Regarding DKM, the annealing process of the $\Softmax$ parameter is updated every $5$ epochs.
    We report in Table~\ref{apx-tab:cnet-exp-setting} the values used for each dataset.
    
    \begin{table*}[!h]
        \caption{Hyper-parameters used to train the baselines for the experiments in Section~\ref{sec:cnet-exp}.}
        \label{apx-tab:dkm-exp-setting}
        \small
        \renewcommand{\arraystretch}{1.2}
        \centering
        
        \noindent
        \begin{tabular*}{.98\linewidth}{@{\extracolsep{\fill}}cccccccccc}
            \toprule
            & {MNIST} & {fMNIST} & {USPS} & {CIFAR10} & {R10K} &{20News} & {10x73k} & {Pendigit} \\ \midrule
            $u$ & $140$ & $140$ & $30$   & $140$ & $20$   & $20$  & $20$ & $20$ \\
            $\gamma$ & $0.1$ & $0.1$ & $0.1$ & $0.1$ & $0.1$ & $0.1$ & $0.1$ & $0.1$ \\
            
            $\lambda^r$ & $0.1$ & $0.01$ & $0.01$   & $0.01$ & $0.01$   & $10^{-4}$  & $10^{-4}$ & $10^{-4}$ \\
            $\lambda^p$ & $1.0$ & $0.01$ & $0.1$    & $0.1$  & $1.0$    & $0.01$    & $10^{-4}$ & $10^{-4}$ \\
            
            $B$      & $256$ & $256$& $256$  & $256$& $256$ & $256$ & $256$ & $256$ \\
            \bottomrule
        \end{tabular*}
    \end{table*}

    \subsection{AE-CM: Clustering performance}\label{sec:cnet-exp-results-full}
    Table~\ref{apx-tab:cnet-exp-results-full} contains the full clustering results for AE-CM, including standard deviation and the best run.
    
    \begin{table*}[!h]
        \caption[Comparison of clustering performance in terms of mean ARI.]
        { Clustering performance of the methods on the experimental datasets including average, standard deviation and best run. 
        }\label{apx-tab:cnet-exp-results-full}
        \tiny
        \renewcommand{\arraystretch}{1.3}
        \centering
        
        \noindent
        \begin{tabular*}{.98\linewidth}{@{\extracolsep{\fill}}p{0.08\textwidth}p{0.02\textwidth}p{0.025\textwidth}p{0.025\textwidth}p{0.035\textwidth}p{0.025\textwidth}p{0.025\textwidth}p{0.035\textwidth}p{0.025\textwidth}p{0.025\textwidth}p{0.035\textwidth}p{0.025\textwidth}p{0.025\textwidth}p{0.025\textwidth}}
            \toprule
            \multirow{2}{*}{Model} & & \multicolumn{3}{c}{MNIST} & \multicolumn{3}{c}{fMNIST} & \multicolumn{3}{c}{USPS} & \multicolumn{3}{c}{CIFAR10} \\
            & & ARI & NMI & ACC & ARI & NMI & ACC & ARI & NMI & ACC & ARI & NMI & ACC \\ \midrule
            
AE+KM & avg std max & $ 65.6$  $ \pm 1.0  $ $ 67.7  $ & $ 71.5$  $ \pm 0.7  $ $ 73.0  $ & $ 78.6$  $ \pm 0.6  $ $ 79.6  $ & $ 39.0$  $ \pm 0.9  $ $ 41.0  $ & $ 55.6$  $ \pm 0.5  $ $ 56.7  $ & $ 53.0$  $ \pm 2.3  $ $ 56.9  $ & $ 57.1$  $ \pm 1.0  $ $ 59.4  $ & $ 64.6$  $ \
pm 0.9  $ $ 66.6  $ & $ 67.5$  $ \pm 1.0  $ $ 69.5  $ & $ 3.2$  $ \pm 0.3  $ $ 3.7  $ & $ 6.5$  $ \pm 0.5  $ $ 7.2  $ & $ 18.9$  $ \pm 0.5  $ $ 20.2  $   \\ \midrule

DCN${}^r$ & avg std max & $ 10.1$  $ \pm 11.0  $ $ 30.4  $ & $ 25.6$  $ \pm 17.4  $ $ 51.6  $ & $ 25.4$  $ \pm 9.3  $ $ 45.1  $ & $ 17.0$  $ \pm 11.7  $ $ 40.7  $ & $ 33.5$  $ \pm 16.9  $ $ 57.5  $ & $ 29.0$  $ \pm 11.3  $ $ 52.4  $ & $ 17.9$  $ \pm 11.5  $ $ 44.1  $ & $ 
36.5$  $ \pm 11.1  $ $ 56.0  $ & $ 37.9$  $ \pm 7.5  $ $ 52.2  $ & $ 3.2$  $ \pm 2.2  $ $ 5.8  $ & $ 5.9$  $ \pm 3.5  $ $ 10.3  $ & $ 18.0$  $ \pm 4.6  $ $ 22.9  $   \\
DCN${}^p$ & avg std max & $ 75.6$  $ \pm 1.4  $ $ 77.6  $ & $\mathbf{ 82.5} $  $ \pm 1.2  $ $ 85.0  $ & $ 83.1$  $ \pm 0.9  $ $ 84.5  $ & $ 38.6$  $ \pm 2.0  $ $ 41.7  $ & $ 57.1$  $ \pm 0.8  $ $ 58.6  $ & $ 53.1$  $ \pm 1.8  $ $ 56.4  $ & $ 63.9$  $ \pm 1.2  $ $ 67.4  $ 
& $ 73.1$  $ \pm 1.0  $ $ 75.5  $ & $ 72.5$  $ \pm 0.7  $ $ 74.5  $ & $ 0.1$  $ \pm 0.4  $ $ 1.3  $ & $ 0.6$  $ \pm 1.2  $ $ 4.0  $ & $ 10.7$  $ \pm 1.5  $ $ 14.5  $   \\ \midrule

DEC${}^r$ & avg std max & $ 11.1$  $ \pm 2.6  $ $ 16.7  $ & $ 19.0$  $ \pm 3.2  $ $ 24.1  $ & $ 28.9$  $ \pm 3.6  $ $ 36.6  $ & $ 22.9$  $ \pm 3.9  $ $ 29.7  $ & $ 38.1$  $ \pm 4.1  $ $ 47.0  $ & $ 39.2$  $ \pm 4.2  $ $ 47.0  $ & $ 36.3$  $ \pm 4.6  $ $ 43.0  $ & $ 46.9$ 
 $ \pm 4.5  $ $ 55.0  $ & $ 46.8$  $ \pm 4.0  $ $ 53.7  $ & $ 3.1$  $ \pm 0.9  $ $ 4.7  $ & $ 5.7$  $ \pm 1.5  $ $ 8.6  $ & $ 18.6$  $ \pm 1.4  $ $ 21.7  $   \\
DEC${}^p$ & avg std max & $ 73.8$  $ \pm 0.7  $ $ 75.2  $ & $ 79.0$  $ \pm 0.4  $ $ 79.5  $ & $ 83.1$  $ \pm 0.5  $ $ 84.1  $ & $ 41.9$  $ \pm 2.0  $ $ 45.9  $ & $ 58.6$  $ \pm 1.9  $ $ 60.6  $ & $ 54.8$  $ \pm 2.2  $ $ 58.4  $ & $\mathbf{ 70.0} $  $ \pm 0.8  $ $ 71.3  $ 
& $\mathbf{ 78.1} $  $ \pm 0.6  $ $ 79.2  $ & $\mathbf{ 76.3} $  $ \pm 0.6  $ $ 77.3  $ & $ 3.1$  $ \pm 1.4  $ $ 4.5  $ & $ 5.6$  $ \pm 2.4  $ $ 7.8  $ & $ 18.2$  $ \pm 3.6  $ $ 21.8  $   \\ \midrule

IDEC${}^r$ & avg std max & $ 27.5$  $ \pm 5.0  $ $ 38.3  $ & $ 39.0$  $ \pm 5.2  $ $ 50.1  $ & $ 42.5$  $ \pm 4.8  $ $ 50.2  $ & $ 35.2$  $ \pm 5.1  $ $ 48.4  $ & $ 50.8$  $ \pm 5.6  $ $ 62.5  $ & $ 48.1$  $ \pm 5.9  $ $ 62.5  $ & $ 41.8$  $ \pm 8.0  $ $ 53.3  $ & $ 53.2$  $ \pm 7.0  $ $ 64.4  $ & $ 54.0$  $ \pm 5.8  $ $ 64.9  $ & $ 2.2$  $ \pm 2.3  $ $ 5.3  $ & $ 3.6$  $ \pm 3.6  $ $ 9.3  $ & $ 14.0$  $ \pm 3.9  $ $ 20.4  $   \\
IDEC${}^p$ & avg std max & $ 74.9$  $ \pm 1.0  $ $ 77.2  $ & $ 80.1$  $ \pm 0.6  $ $ 81.3  $ & $ 83.4$  $ \pm 0.6  $ $ 84.9  $ & $ 42.8$  $ \pm 1.8  $ $ 47.0  $ & $ 59.8$  $ \pm 0.7  $ $ 61.3  $ & $ 55.4$  $ \pm 2.2  $ $ 59.2  $ & $\mathbf{ 70.0} $  $ \pm 0.8  $ $ 71.5  $ & $\mathbf{ 78.0} $  $ \pm 0.7  $ $ 79.3  $ & $\mathbf{ 76.1} $  $ \pm 0.6  $ $ 77.0  $ & $ 4.2$  $ \pm 0.4  $ $ 4.9  $ & $ 7.4$  $ \pm 0.7  $ $ 8.5  $ & $ 20.2$  $ \pm 1.1  $ $ 22.0  $   \\ \midrule

DKM${}^r$ & avg std max & $ 72.5$  $ \pm 5.0  $ $ 83.6  $ & $ 77.3$  $ \pm 3.1  $ $ 83.5  $ & $ 81.2$  $ \pm 5.0  $ $ 92.3  $ & $ 41.8$  $ \pm 2.2  $ $ 46.0  $ & $ 56.4$  $ \pm 0.9  $ $ 58.6  $ & $ 54.6$  $ \pm 3.3  $ $ 59.4  $ & $ 58.3$  $ \pm 4.1  $ $ 65.4  $ & $ 67.0$  $ \pm 2.4  $ $ 70.9  $ & $ 68.6$  $ \pm 4.5  $ $ 77.1  $ & $\mathbf{ 5.8} $  $ \pm 0.7  $ $ 7.1  $ & $\mathbf{ 9.9} $  $ \pm 1.2  $ $ 12.6  $ & $ 21.3$  $ \pm 2.1  $ $ 24.3  $   \\
DKM${}^p$ & avg std max & $ 74.0$  $ \pm 2.8  $ $ 76.9  $ & $ 78.3$  $ \pm 1.7  $ $ 80.2  $ & $ 82.7$  $ \pm 3.0  $ $ 85.3  $ & $ 36.2$  $ \pm 3.0  $ $ 40.5  $ & $ 52.0$  $ \pm 2.3  $ $ 56.3  $ & $ 47.0$  $ \pm 3.7  $ $ 51.5  $ & $ 60.4$  $ \pm 3.3  $ $ 67.6  $ & $ 71.8$  $ \pm 1.9  $ $ 76.0  $ & $ 68.9$  $ \pm 3.7  $ $ 74.9  $ & $\mathbf{ 5.9} $  $ \pm 0.4  $ $ 6.2  $ & $\mathbf{ 10.0} $  $ \pm 0.6  $ $ 10.8  $ & $ 19.7$  $ \pm 1.2  $ $ 20.8  $   \\ \midrule

Cluster GAN${}^r$ & avg std max & $ 63.6$  $ \pm 8.0  $ $ 80.3  $ & $ 71.8$  $ \pm 5.1  $ $ 81.6  $ & $ 76.8$  $ \pm 6.5  $ $ 90.1  $ & $\mathbf{ 46.5} $  $ \pm 1.6  $ $ 48.9  $ & $\mathbf{ 60.7} $  $ \pm 1.3  $ $ 62.7  $ & $\mathbf{ 59.0} $  $ \pm 1.3  $ $ 61.7  $ & $ 57.4$  $ \pm 2.2  $ $ 61.4  $ & $ 67.9$  $ \pm 1.6  $ $ 71.2  $ & $ 70.0$  $ \pm 2.3  $ $ 73.9  $ & $ 3.2$  $ \pm 0.5  $ $ 4.4  $ & $ 7.6$  $ \pm 0.9  $ $ 10.0  $ & $ 20.4$  $ \pm 0.8  $ $ 21.7  $   \\

VIB-GMM${}^r$ & avg std max & $ 73.3$  $ \pm 7.9  $ $ 89.9  $ & $ 78.3$  $ \pm 5.1  $ $ 89.0  $ & $ 81.5$  $ \pm 6.9  $ $ 95.2  $ & $ 43.7$  $ \pm 3.3  $ $ 49.9  $ & $ 58.4$  $ \pm 2.3  $ $ 62.7  $ & $\mathbf{ 59.3} $  $ \pm 4.1  $ $ 67.3  $ & $ 59.9$  $ \pm 3.5  $ $ 67.5  $ & $ 67.7$  $ \pm 2.9  $ $ 73.2  $ & $ 68.4$  $ \pm 4.3  $ $ 79.3  $ & $\mathbf{ 6.0} $  $ \pm 0.2  $ $ 6.5  $ & $\mathbf{ 10.1} $  $ \pm 0.2  $ $ 10.3  $ & $\mathbf{ 24.0} $  $ \pm 0.5  $ $ 24.8  $   \\ \midrule

AE-CM${}^r$ & avg std max & $\mathbf{ 77.9} $  $ \pm 4.0  $ $ 88.6  $ & $ 80.9$  $ \pm 2.4  $ $ 87.2  $ & $\mathbf{ 86.1} $  $ \pm 3.2  $ $ 94.6  $ & $ 43.7$  $ \pm 2.9  $ $ 48.9  $ & $ 55.6$  $ \pm 1.8  $ $ 58.5  $ & $\mathbf{ 59.2} $  $ \pm 3.5  $ $ 65.6  $ & $ 55.1$  $ \pm 4.5  $ $ 60.6  $ & $ 63.4$  $ \pm 3.6  $ $ 67.4  $ & $ 65.8$  $ \pm 4.7  $ $ 72.2  $ & $ 4.1$  $ \pm 0.8  $ $ 5.3  $ & $ 7.5$  $ \pm 1.3  $ $ 9.3  $ & $ 20.4$  $ \pm 1.5  $ $ 22.5  $   \\
AE-CM${}^p$ & avg std max & $\mathbf{ 79.4} $  $ \pm 0.4  $ $ 80.3  $ & $\mathbf{ 82.4} $  $ \pm 0.4  $ $ 83.2  $ & $\mathbf{ 86.5} $  $ \pm 0.4  $ $ 87.3  $ & $ 43.1$  $ \pm 2.6  $ $ 48.4  $ & $ 56.3$  $ \pm 1.7  $ $ 58.5  $ & $\mathbf{ 58.5} $  $ \pm 2.8  $ $ 64.9  $ & $\mathbf{ 69.7} $  $ \pm 4.1  $ $ 80.3  $ & $ 76.7$  $ \pm 2.3  $ $ 80.5  $ & $\mathbf{ 76.8} $  $ \pm 3.8  $ $ 87.5  $ & $ 4.1$  $ \pm 0.6  $ $ 5.6  $ & $ 7.6$  $ \pm 1.1  $ $ 9.7  $ & $ 20.2$  $ \pm 0.8  $ $ 21.4  $   \\

            \bottomrule
            \multirow{2}{*}{Model} & & \multicolumn{3}{c}{R10K} & \multicolumn{3}{c}{20News} & \multicolumn{3}{c}{10x73k} & \multicolumn{3}{c}{Pendigit} \\
            & & ARI & NMI & ACC & ARI & NMI & ACC & ARI & NMI & ACC & ARI & NMI & ACC \\ \midrule
            
AE+KM & avg std max & $ 61.0$  $ \pm 3.5  $ $ 67.3  $ & $ 56.8$  $ \pm 3.1  $ $ 62.2  $ & $ 74.5$  $ \pm 3.3  $ $ 83.3  $ & $ 11.3$  $ \pm 1.6  $ $ 13.8  $ & $ 27.4$  $ \pm 2.5  $ $ 31.0  $ & $ 24.8$  $ \pm 2.5  $ $ 28.6  $ & $ 54.3$  $ \pm 6.6  $ $ 64.5  $ & $ 72.5$  $ \pm 3.0  $ $ 78.3  $ & $ 64.4$  $ \pm 4.8  $ $ 72.4  $ & $ 55.2$  $ \pm 3.6  $ $ 62.9  $ & $ 68.2$  $ \pm 1.7  $ $ 71.5  $ & $ 70.2$  $ \pm 4.3  $ $ 78.5  $   \\ \midrule

DCN${}^r$ & avg std max & $ 18.0$  $ \pm 10.3  $ $ 40.1  $ & $ 19.3$  $ \pm 8.8  $ $ 35.1  $ & $ 49.5$  $ \pm 7.8  $ $ 63.9  $ & $ 0.0$  $ \pm 0.1  $ $ 0.3  $ & $ 0.2$  $ \pm 0.3  $ $ 0.9  $ & $ 5.6$  $ \pm 0.5  $ $ 6.8  $ & $ 5.3$  $ \pm 6.9  $ $ 26.6  $ & $ 17.2$  $ \pm 16.7  $ $ 52.3  $ & $ 23.9$  $ \pm 7.2  $ $ 38.9  $ & $ 0.1$  $ \pm 0.4  $ $ 2.0  $ & $ 0.8$  $ \pm 2.6  $ $ 11.3  $ & $ 10.8$  $ \pm 1.2  $ $ 15.5  $   \\
DCN${}^p$ & avg std max & $\mathbf{ 65.4} $  $ \pm 1.9  $ $ 67.1  $ & $\mathbf{ 61.1} $  $ \pm 1.0  $ $ 62.3  $ & $\mathbf{ 76.6} $  $ \pm 1.1  $ $ 80.2  $ & $ 11.7$  $ \pm 2.0  $ $ 16.1  $ & $ 33.5$  $ \pm 2.8  $ $ 37.4  $ & $ 25.3$  $ \pm 2.8  $ $ 30.4  $ & $ 9.6$  $ \pm 22.0  $ $ 65.8  $ & $ 13.8$  $ \pm 27.3  $ $ 80.4  $ & $ 25.2$  $ \pm 18.6  $ $ 72.4  $ & $ 56.8$  $ \pm 4.4  $ $ 66.1  $ & $ 72.0$  $ \pm 1.7  $ $ 75.9  $ & $ 70.8$  $ \pm 4.7  $ $ 79.2  $   \\ \midrule

DEC${}^r$ & avg std max & $ 12.2$  $ \pm 6.6  $ $ 26.1  $ & $ 13.2$  $ \pm 6.3  $ $ 27.0  $ & $ 43.8$  $ \pm 6.1  $ $ 56.2  $ & $ 3.2$  $ \pm 0.9  $ $ 4.3  $ & $ 7.8$  $ \pm 2.2  $ $ 10.6  $ & $ 10.0$  $ \pm 1.1  $ $ 10.7  $ & $ 31.4$  $ \pm 8.6  $ $ 53.1  $ & $ 43.5$  $ \pm 9.6  $ $ 66.3  $ & $ 46.3$  $ \pm 8.1  $ $ 68.4  $ & $ 36.8$  $ \pm 6.7  $ $ 49.6  $ & $ 52.3$  $ \pm 6.9  $ $ 65.7  $ & $ 49.3$  $ \pm 6.1  $ $ 61.7  $   \\
DEC${}^p$ & avg std max & $ 56.8$  $ \pm 1.9  $ $ 60.7  $ & $ 56.0$  $ \pm 2.0  $ $ 59.6  $ & $ 72.8$  $ \pm 2.0  $ $ 77.6  $ & $ 5.5$  $ \pm 0.6  $ $ 6.5  $ & $ 11.3$  $ \pm 1.3  $ $ 13.7  $ & $ 11.8$  $ \pm 0.3  $ $ 12.2  $ & $ 53.5$  $ \pm 18.6  $ $ 73.3  $ & $ 67.1$  $ \pm 22.8  $ $ 83.4  $ & $ 62.1$  $ \pm 15.9  $ $ 78.5  $ & $ 59.6$  $ \pm 3.5  $ $ 65.8  $ & $ 72.8$  $ \pm 1.6  $ $ 75.7  $ & $ 72.2$  $ \pm 4.0  $ $ 79.3  $   \\ \midrule

IDEC${}^r$ & avg std max & $ 8.6$  $ \pm 5.6  $ $ 21.7  $ & $ 9.5$  $ \pm 4.9  $ $ 21.8  $ & $ 44.1$  $ \pm 5.6  $ $ 55.3  $ & $ 0.0$  $ \pm 0.0  $ $ 0.1  $ & $ 0.1$  $ \pm 0.3  $ $ 1.1  $ & $ 5.5$  $ \pm 0.4  $ $ 6.7  $ & $ 33.7$  $ \pm 11.6  $ $ 67.5  $ & $ 46.5$  $ \pm 12.9  $ $ 82.8  $ & $ 44.4$  $ \pm 8.5  $ $ 69.6  $ & $ 43.3$  $ \pm 6.9  $ $ 57.3  $ & $ 61.2$  $ \pm 5.7  $ $ 70.7  $ & $ 53.9$  $ \pm 7.5  $ $ 69.3  $   \\
IDEC${}^p$ & avg std max & $ 59.7$  $ \pm 1.3  $ $ 62.5  $ & $ 56.3$  $ \pm 0.9  $ $ 57.8  $ & $ 73.9$  $ \pm 1.6  $ $ 78.8  $ & $ 5.9$  $ \pm 0.4  $ $ 6.6  $ & $ 12.6$  $ \pm 1.2  $ $ 15.0  $ & $ 12.0$  $ \pm 0.2  $ $ 12.4  $ & $ 60.1$  $ \pm 4.9  $ $ 73.4  $ & $ 75.9$  $ \pm 4.0  $ $ 83.5  $ & $ 66.5$  $ \pm 3.9  $ $ 78.8  $ & $ 57.9$  $ \pm 3.8  $ $ 65.7  $ & $ 71.6$  $ \pm 1.8  $ $ 75.1  $ & $ 71.0$  $ \pm 4.3  $ $ 79.0  $   \\ \midrule

DKM${}^r$ & avg std max & $ 51.3$  $ \pm 4.9  $ $ 63.3  $ & $ 49.5$  $ \pm 4.4  $ $ 58.9  $ & $ 72.3$  $ \pm 3.3  $ $ 81.0  $ & $ 4.7$  $ \pm 0.9  $ $ 5.6  $ & $ 14.1$  $ \pm 2.5  $ $ 17.8  $ & $ 10.9$  $ \pm 1.1  $ $ 13.0  $ & $ 65.5$  $ \pm 5.1  $ $ 77.0  $ & $ 71.3$  $ \pm 3.7  $ $ 78.5  $ & $ 77.0$  $ \pm 5.0  $ $ 88.1  $ & $ 52.4$  $ \pm 3.1  $ $ 60.3  $ & $ 65.6$  $ \pm 1.6  $ $ 68.8  $ & $ 66.9$  $ \pm 3.3  $ $ 73.6  $   \\
DKM${}^p$ & avg std max & $ 57.7$  $ \pm 1.1  $ $ 59.5  $ & $ 55.5$  $ \pm 1.3  $ $ 58.1  $ & $\mathbf{ 76.5} $  $ \pm 1.8  $ $ 78.5  $ & $ 20.9$  $ \pm 6.5  $ $ 30.7  $ & $ 39.2$  $ \pm 5.3  $ $ 46.7  $ & $ 34.3$  $ \pm 6.9  $ $ 44.0  $ & $ 38.1$  $ \pm 14.6  $ $ 57.2  $ & $ 55.4$  $ \pm 11.4  $ $ 69.2  $ & $ 51.6$  $ \pm 10.2  $ $ 63.6  $ & $ 15.4$  $ \pm 11.1  $ $ 33.1  $ & $ 27.4$  $ \pm 18.0  $ $ 50.2  $ & $ 25.1$  $ \pm 10.7  $ $ 46.1  $   \\ \midrule

Cluster GAN${}^r$ & avg std max & $ 33.7$  $ \pm 11.6  $ $ 47.8  $ & $ 35.5$  $ \pm 11.7  $ $ 50.3  $ & $ 61.4$  $ \pm 8.6  $ $ 71.5  $ & $ 18.6$  $ \pm 2.2  $ $ 22.0  $ & $ 34.1$  $ \pm 2.6  $ $ 38.8  $ & $ 34.1$  $ \pm 2.3  $ $ 39.7  $ & $ 39.5$  $ \pm 2.0  $ $ 43.5  $ & $ 52.1$  $ \pm 2.1  $ $ 55.9  $ & $ 55.5$  $ \pm 2.6  $ $ 61.6  $ & $\mathbf{ 62.9} $  $ \pm 1.7  $ $ 66.8  $ & $ 74.2$  $ \pm 1.3  $ $ 76.5  $ & $\mathbf{ 75.6} $  $ \pm 2.1  $ $ 78.0  $   \\

VIB-GMM${}^r$ & avg std max & $ 27.8$  $ \pm 7.1  $ $ 43.9  $ & $ 28.7$  $ \pm 6.3  $ $ 42.5  $ & $ 56.6$  $ \pm 4.9  $ $ 64.9  $ & $ 0.0$  $ \pm 0.0  $ $ 0.0  $ & $ 0.0$  $ \pm 0.0  $ $ 0.0  $ & $ 0.0$  $ \pm 0.0  $ $ 0.0  $ & $ 51.5$  $ \pm 27.2  $ $ 81.1  $ & $ 60.7$  $ \pm 30.7  $ $ 83.8  $ & $ 60.0$  $ \pm 23.3  $ $ 87.0  $ & $\mathbf{ 64.5} $  $ \pm 4.3  $ $ 72.3  $ & $\mathbf{ 75.7} $  $ \pm 2.8  $ $ 80.1  $ & $\mathbf{ 74.2} $  $ \pm 3.5  $ $ 82.7  $   \\ \midrule

AE-CM${}^r$ & avg std max & $ 42.9$  $ \pm 11.9  $ $ 62.7  $ & $ 45.6$  $ \pm 6.9  $ $ 57.6  $ & $ 67.7$  $ \pm 7.3  $ $ 79.6  $ & $\mathbf{ 31.5} $  $ \pm 4.6  $ $ 38.7  $ & $\mathbf{ 45.3} $  $ \pm 3.3  $ $ 50.8  $ & $\mathbf{ 43.3} $  $ \pm 4.1  $ $ 50.5  $ & $ 73.1$  $ \pm 5.9  $ $ 85.6  $ & $ 79.0$  $ \pm 3.7  $ $ 86.2  $ & $ 80.4$  $ \pm 5.4  $ $ 92.1  $ & $\mathbf{ 64.6} $  $ \pm 4.1  $ $ 75.3  $ & $\mathbf{ 75.0} $  $ \pm 2.8  $ $ 82.1  $ & $\mathbf{ 75.7} $  $ \pm 4.1  $ $ 84.3  $   \\
AE-CM${}^p$ & avg std max & $ 64.1$  $ \pm 2.0  $ $ 66.7  $ & $ 60.0$  $ \pm 1.3  $ $ 62.5  $ & $\mathbf{ 76.3} $  $ \pm 1.8  $ $ 82.3  $ & $ 16.8$  $ \pm 2.5  $ $ 21.2  $ & $ 29.0$  $ \pm 2.7  $ $ 33.5  $ & $ 32.5$  $ \pm 2.9  $ $ 37.8  $ & $\mathbf{ 82.3} $  $ \pm 5.7  $ $ 86.9  $ & $\mathbf{ 83.7} $  $ \pm 3.4  $ $ 86.8  $ & $\mathbf{ 89.4} $  $ \pm 5.0  $ $ 92.9  $ & $\mathbf{ 60.1} $  $ \pm 14.4  $ $ 69.8  $ & $\mathbf{ 70.4} $  $ \pm 16.4  $ $ 78.2  $ & $\mathbf{ 73.0} $  $ \pm 14.8  $ $ 81.4  $   \\

            \bottomrule
        \end{tabular*}
    \end{table*}
    
\end{document}


\twocolumn[
    \icmltitle{Supplementary: Theoretically Grounded Centroid-based Deep Clustering}
    
    
    
    \icmlsetsymbol{equal}{*}
    
    \begin{icmlauthorlist}
        \icmlauthor{Aeiau Zzzz}{equal,to}
        \icmlauthor{Bauiu C.~Yyyy}{equal,to,goo}
        \icmlauthor{Cieua Vvvvv}{goo}
        \icmlauthor{Iaesut Saoeu}{ed}
        \icmlauthor{Fiuea Rrrr}{to}
        \icmlauthor{Tateu H.~Yasehe}{ed,to,goo}
        \icmlauthor{Aaoeu Iasoh}{goo}
        \icmlauthor{Buiui Eueu}{ed}
        \icmlauthor{Aeuia Zzzz}{ed}
        \icmlauthor{Bieea C.~Yyyy}{to,goo}
        \icmlauthor{Teoau Xxxx}{ed}
        \icmlauthor{Eee Pppp}{ed}
    \end{icmlauthorlist}
    
    \icmlaffiliation{to}{Department of Computation, University of Torontoland, Torontoland, Canada}
    \icmlaffiliation{goo}{Googol ShallowMind, New London, Michigan, USA}
    \icmlaffiliation{ed}{School of Computation, University of Edenborrow, Edenborrow, United Kingdom}
    
    \icmlcorrespondingauthor{Cieua Vvvvv}{c.vvvvv@googol.com}
    \icmlcorrespondingauthor{Eee Pppp}{ep@eden.co.uk}
    
    \icmlkeywords{Machine Learning, ICML}
    
    \vskip 0.3in
    ]
    
    
    


    \begin{figure*}[!h]
    \centering
        \def\svgwidth{\linewidth}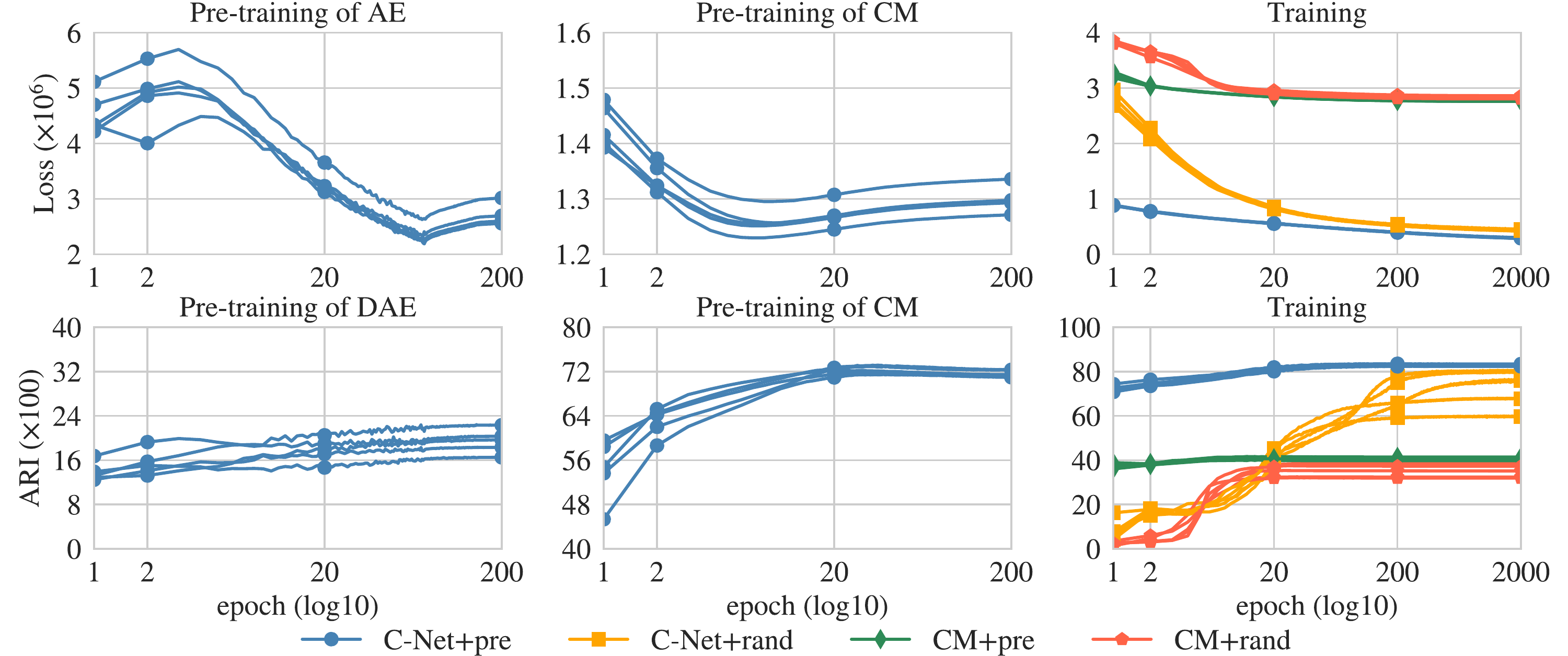
        \caption{Evolution of the loss and ARI during optimization of CM and C-Net with and without pre-training. Five runs of each model are plotted. The epochs (x-axis) are represented in log-scale.}
        \label{fig:exp-loss}
    \end{figure*}
    
    \section{Analysis of the Relaxation}
    
    The loss function of the clustering module nested in the C-Net is relaxed: the regularization term $||\bmu_k||$ is discarded. 
    To empirically justify this modification, the same experiment is run for C-Net with the regularization term in the loss function (with and without pre-training).
    The results in terms of ARI are reported in Table~\ref{tab:exp-relax}.

    \begin{table}[!h]
        \caption
        {Clustering performance in terms of mean ARI~$(\times 100)$ of a regularized C-Net with and without pre-training.
        }\label{tab:exp-relax}
        \centering
        \begin{tabular*}{\linewidth}{@{\extracolsep{\fill}}ccccc}
            \toprule
            & \multicolumn{2}{c}{reg. C-Net+rand} & \multicolumn{2}{c}{reg. C-Net+pre}\\
            & avg. $\pm$ sd.  & best & avg. $\pm$ sd.  & best \\
            \midrule
            MNIST & $35.1 \pm 2.9$ & $39.2$ & $62.4 \pm 2.2$ & $65.1$ \\
            USPS & $26.4 \pm 4.4$ & $32.5$ & $59.9 \pm 4.7$ & $66.6$ \\
            R10K & $37.0 \pm 4.7$ & $45.2$ & $30.0 \pm 5.4$ & $38.8$ \\
            \bottomrule
        \end{tabular*}
    \end{table}
    
    The results confirm the negative impact of the regularization on the clustering performance.
    The regularization yields lower average ARIs in every settings, except for C-Net+rand on R10K.
    For MNIST and USPS, a randomly initialized C-Net returns ARIs more than $50\%$ smaller than without the regularization.
    For these two data-set, the standard deviations of C-Net are larger than those reported in Table~1 of the main paper.
    In case of R10K, the regularized model performs on par with the non-regularized one.

    \section{Analysis of the Optimization}\label{sec:opt}
    
    The evolution of the loss function and of the ARI during the optimization of CM and C-Net, with and without pre-training is shown in Figure~\ref{fig:exp-loss}.
    The benefit of pre-training the CM and C-Net appears clearly in the plots of the last column. 
    The respective curves in terms of loss or ARI start with better values.
    The impact on the stability is better observed in the plots in terms of ARI (bottom).
    Except for C-Net+rand, convergence is reached before $200$ epochs, and even earlier for CM (about $20$ seems enough).
    Regarding C-Net+rand, the runs converging the fastest in terms of ARI do so in about $200$ epochs, which is less than the $200+200$ epochs of the pre-training, but also perform worse. 
    
    The two first columns focus on the two pre-training phases of C-Net: the pre-training of the autoencoder and of the clustering module, respectively.
    Recall that the CM is initialized with $k$-means++ before being pre-trained, hence the gap between plots of the first and second columns.
    The tuning of the AE greatly reduces the loss but has limited impact on the clustering.
    The inverse holds for the tuning of the CM. 
    Note that during both phases, the loss function reaches a minimum and then increases again. 
    This behavior does not correlate with that of the ARI.
    
    As a summary, CM needs less than $50$ iterations to converge, be it for training or pre-training.
    As other studies~\cite{xie2016unsupervised} also pointed out, pre-training the AE does yield more consistent results but not necessarily a faster convergence, if the tuning epochs are included. 
    Further studies are necessary to assess the impact of a shorter tuning of the AE, e.g., stop before convergence.

    \section{Additional experimental results}
    
    In this section we provide tables equivalent to Table 1 of the main paper in terms of normalized mutual information (NMI) and accuracy.
    
    \begin{table*}[!t]
    \caption[Comparison of clustering performance in terms of mean NMI.]
    {Comparison of clustering performance in terms of mean NMI~$(\times 100)$ with standard deviation and the results of the best performing model out of ten runs.
    The highest, statistically significant score for each data-set is marked in bold face. 
    }\label{tab:exp-results-nmi}
    
    \centering
    \begin{tabular*}{\textwidth}{@{\extracolsep{\fill}}ccccccc}
        \toprule
        \multirow{2}{*}{Method} & \multicolumn{2}{c}{MNIST} & \multicolumn{2}{c}{USPS} & \multicolumn{2}{c}{R10K} \\
        & avg. $\pm$ std. & best & avg. $\pm$ std.  & best & avg. $\pm$ std.  & best \\
        \midrule
$k$-means & $48.8 \pm 1.3$ & $50.5$ & $62.3 \pm 1.4$ & $ 65.0 $ & $34.7 \pm 7.5$ & $56.5$\\ 
GMM       & $35.3 \pm 1.3$ & $37.9$ & $51.6 \pm 2.4$ & $ 53.8 $ & $34.9\pm 7.5$ & $56.5$\\ 
        \midrule
DEC+rand & $ 25.9 \pm 13.5 $ & $ 39.2$ & $ 38.9 \pm 4.9 $ & $ 48.4$ & $ 11.9 \pm 5.0 $ & $ 18.8$\\
IDEC+rand & $ 34.7 \pm 5.5 $ & $ 44.5$ & $ 45.2 \pm 13.2 $ & $ 65.7$ & $ 15.5 \pm 3.5 $ & $ 22.1$\\
DCN+rand & $44.9 \pm 1.4$ & $47.4$ & $48.9 \pm 16.6$ & $60.3$ & $11.1 \pm 3.7$ & $16.6$\\ 
        \midrule
DEC+pre & $ 80.4 \pm 1.4 $ & $ 83.2$ & $ 72.9 \pm 2.5 $ & $ 75.7$ & $ \bf 54.7 \pm 3.7 $ & $ \bf 61.6$\\
IDEC+pre & $ 82.3 \pm 0.8 $ & $ 84.1$ & $ 72.7 \pm 2.5 $ & $ 76.6$ & $ 50.7 \pm 2.6 $ & $ 56.5$\\
DCN+pre & $82.0 \pm 0.2$ & $ 82.2 $ & $71.0 \pm 1.3$ & $ 74.7 $ & $38.0 \pm 4.9$ & $ 40.1 $ \\
        \midrule
IIC+rand & $ 52.9 \pm 4.3 $ & $ 62.3$ & $ 59.7 \pm 4.5 $ & $ 66.1$ & $ 20.0 \pm 5.4 $ & $ 28.6$\\
        \midrule
        \midrule
CM+rand & $ 49.2 \pm 3.8 $ & $ 53.9$ & $ 60.4 \pm 3.2 $ & $ 65.1$ & $ 9.2 \pm 1.0 $ & $ 11.6$\\
CM+pre & $ 55.0 \pm 0.5 $ & $ 55.3$ & $ 67.0 \pm 0.5 $ & $ 68.1$ & $ 33.9 \pm 2.3 $ & $ 37.3$\\
        \midrule
C-Net+rand & $ 82.0 \pm 5.2 $ & $ \bf90.9$ & $ 65.4 \pm 2.1 $ & $ 69.8$ & $ 31.6 \pm 7.2 $ & $ 45.5$\\
C-Net+pre & $ \bf 86.3 \pm 0.5 $ & $ 87.7$ & $ \bf 76.0 \pm 1.5 $ & $ \bf 78.7$ & $ 48.2 \pm 3.8 $ & $ 54.7$\\
        \bottomrule
    \end{tabular*}
\end{table*}

\begin{table*}[!t]
    \caption[Comparison of clustering performance in terms of mean accuracy.]
    {Comparison of clustering performance in terms of mean accuracy~$(\times 100)$ with standard deviation and the results of the best performing model out of ten runs.
    The highest, statistically significant score for each data-set is marked in bold face. 
    }\label{tab:exp-results-acc}
    
    \centering
    \begin{tabular*}{\textwidth}{@{\extracolsep{\fill}}ccccccc}
        \toprule
        \multirow{2}{*}{Method} & \multicolumn{2}{c}{MNIST} & \multicolumn{2}{c}{USPS} & \multicolumn{2}{c}{R10K} \\
        & avg. $\pm$ std. & best & avg. $\pm$ std.  & best & avg. $\pm$ std.  & best \\
        \midrule
$k$-means & $52.6 \pm 3.5$ & $58.5$ & $64.6 \pm 3.2$ & $ 68.4 $ & $56.3 \pm 8.2$ & $80.3$\\ 
GMM       & $41.9 \pm 2.2$ & $45.6$ & $53.1 \pm 3.0$ & $ 56.9 $ & $56.4 \pm 8.1$ & $80.3$\\ 
        \midrule
DEC+rand & $ 31.2 \pm 10.7 $ & $ 44.4$ & $ 40.2 \pm 3.3 $ & $ 46.1$ & $ 43.9 \pm 5.9 $ & $ 52.6$\\
IDEC+rand & $ 37.5 \pm 4.1 $ & $ 46.3$ & $ 46.8 \pm 10.4 $ & $ 63.4$ & $ 46.6 \pm 3.0 $ & $ 52.4$\\
DCN+rand & $50.0 \pm 4.3$ & $58.1$ & $50.5 \pm 14.3$ & $64.8$ & $40.3 \pm 4.2$ & $47.9$\\ 
        \midrule
DEC+pre & $ 83.6 \pm 1.6 $ & $ 86.5$ & $ 69.2 \pm 5.2 $ & $ 76.1$ & $ \bf 74.4 \pm 5.2 $ & $ \bf 83.5$\\
IDEC+pre & $ 85.0 \pm 0.8 $ & $ 86.7$ & $ 69.1 \pm 4.5 $ & $ 75.4$ & $ \bf 73.6 \pm 5.0 $ & $ 81.2$\\
DCN+pre & $85.9 \pm 0.1$ & $ 86.1 $ & $71.2 \pm 1.1$ & $ 74.6 $ & $62.9 \pm 5.3$ & $ 65.1 $ \\
        \midrule
IIC+rand & $ 55.9 \pm 6.0 $ & $ 67.7$ & $ 58.6 \pm 5.8 $ & $ 66.7$ & $ 47.2 \pm 6.5 $ & $ 55.2$\\
        \midrule
        \midrule
CM+rand & $ 49.1 \pm 4.6 $ & $ 56.0$ & $ 63.9 \pm 5.4 $ & $ 72.5$ & $ 43.5 \pm 1.2 $ & $ 44.9$\\
CM+pre & $ 56.2 \pm 0.2 $ & $ 56.6$ & $ 68.9 \pm 0.3 $ & $ 69.7$ & $ 52.9 \pm 1.3 $ & $ 55.1$\\
        \midrule
C-Net+rand & $ \bf 86.3 \pm 6.1 $ & $ \bf 96.3$ & $ 67.1 \pm 3.1 $ & $ 73.0$ & $ 60.4 \pm 8.4 $ & $ 72.9$\\
C-Net+pre & $ \bf 88.5 \pm 2.0 $ & $ 94.3$ & $ \bf  75.8 \pm 0.9 $ & $ \bf 77.0$ & $ \bf 72.2 \pm 4.6 $ & $ 81.2$\\
        \bottomrule
    \end{tabular*}
\end{table*}

    \section{Source code}
    The code for the experiments that we performed in this paper can be found in the \emph{Code} folder. A README.md file is included with usage instructions. 
    
    \bibliography{biblio}
    \bibliographystyle{plainnat}